\documentclass{article}

\usepackage[main, preprint]{neurips_2026}

\usepackage[utf8]{inputenc} %
\usepackage[T1]{fontenc}    %
\usepackage{hyperref}       %
\usepackage{url}            %
\usepackage{booktabs}       %
\usepackage{amsfonts}       %
\usepackage{nicefrac}       %
\usepackage{microtype}      %
\usepackage{xcolor}         %

\usepackage{enumitem}
\usepackage{algorithm}
\usepackage{algorithmic} 
\usepackage{amsmath}  
\usepackage{amssymb}
\usepackage{graphicx}
\usepackage{caption}
\usepackage{multirow}
\usepackage{bm}
\usepackage{tablefootnote}
\usepackage[normalem]{ulem}
\useunder{\uline}{\ul}{}

\title{Holistic Scaling Laws for Optimal Mixture-of-Experts Architecture Optimization}

\author{
  Weilin Wan$^{1,2}$\thanks{This work was conducted while Weilin Wan was an intern at Hi Lab, Xiaohongshu Inc.}, Jingtao Han$^{3}$, Debing Zhang$^{2}$, Weizhong Zhang$^{4}$\thanks{Corresponding author: weizhongzhang@fudan.edu.cn.}, Cheng Jin$^{1}$ \\
  $^{1}$College of Computer Science and Artificial Intelligence, Fudan University \\
  $^{2}$Hi Lab, Xiaohongshu Inc.\\
  $^{3}$Project Numina\\
  $^{4}$School of Data Science, Fudan University
}

\renewcommand{\acksection}{\section*{Acknowledgments and disclosure of funding}}

\begin{document}

\maketitle

\begin{abstract}
Scaling laws govern macroscopic resource allocation for LLMs, yet precise architectural configurations for Mixture-of-Experts (MoE) models remain guided by heuristics. Existing MoE scaling studies either incorporate MoE-specific variables into scaling formulas, causing fitted coefficients to grow rapidly without proportional experimental support, or fix all non-MoE factors, implicitly assuming global architecture does not influence local MoE scaling.
We propose a reusable framework for holistic MoE architectural optimization. We first show that MoE architectures decouple FLOPs per token ($M$), active parameters ($N_a$), and total parameters ($N$), which serve as distinct proxies for computation, weight access, and parameter storage. We therefore establish a joint constraint triad of $M$, active parameters ($N_a$), and total parameters ($N$) for rigorous MoE characterization. To tame the resulting $\mathcal{O}(n^{16})$ search space, we employ mathematical decoupling: structural constraints and a rank-preserving property of the hidden dimension factorize the optimization into an $\mathcal{O}(n^3) + \mathcal{O}(n^2)$ two-phase search.
Through extensive validation across 670+ MoE models spanning $10^{18}$ to $3 \times 10^{20}$ FLOPs, we derive globally applicable scaling laws mapping any compute budget to its optimal architecture. A key finding is that the near-optimal configuration band widens with scale, providing a quantitative basis for trade-offs between scaling law recommendations and engineering constraints. Our work delivers actionable blueprints for optimal MoE design under arbitrary compute budgets.
\end{abstract}
\section{Introduction}
\label{sec:intro}

Scaling laws~\citep{hestness2017deep,kaplan2020scaling,henighan2020scaling,achiam2023gpt,jaech2024openai,team2023gemini,bi2024deepseek,clark2022unified,sardana2023beyond,gadre2024language} provide a predictable framework for LLM resource allocation, primarily formulating optimal model size or FLOPs per token under a given compute budget~\citep{kaplan2020scaling, hoffmann2022training}. The reliability of such laws hinges on the ratio of data points to fitted coefficients: \cite{hoffmann2022training} trained over 400 models to fit merely 5 coefficients. Yet translating macro-level predictions into concrete architectural configurations, such as network depth, width, or the Attention-to-FFN parameter ratio, remains underexplored. Although \cite{kaplan2020scaling} largely concluded that structural hyperparameters have minimal impact, their data revealed loss variations exceeding 8\% across configurations. Despite this non-trivial gap, the ``ancestral wisdom'' of dense Transformer design~\citep{brown2020language,vaswani2017attention}, accumulated over years of community practice, has effectively mitigated the need for configuration-level scaling laws in dense models.

The Mixture-of-Experts (MoE) architecture~\citep{shazeer2017outrageously,lepikhin2020gshard,fedus2022switch,muennighoff2024olmoe} fundamentally disrupts this paradigm. MoE introduces additional architectural dimensions, including the number of experts, routing sparsity, and expert granularity, exponentially expanding the design space. Unlike dense hyperparameters with decades of accumulated intuition, these MoE dimensions lack established heuristics. Crucially, the trade-offs among these novel dimensions and traditional dense components remain opaque, rendering heuristics developed for dense LLMs inadequate for this expanded space.

A growing body of work~\citep{clark2022unified,ludziejewski2024scaling,wang2024scaling,ludziejewski2025joint,dai2024deepseekmoe,zoph2022st,jelassi2024mixture,abnar2025parameters} has addressed MoE scaling, generally adopting one of two strategies. The first directly incorporates MoE-specific variables into the scaling formula, but each new variable introduces additional fitted coefficients, making reliable fitting difficult without proportionally more experiments. The second fixes all non-MoE factors (e.g., Attention, dense backbone, global model shape) and studies only MoE internals such as expert count or routing sparsity. While experimentally tractable, this approach implicitly assumes that global architectural parameters do not influence local MoE scaling behavior, leaving cross-component interactions unexplored.

This paper identifies and addresses the gap between these two strategies: studying the interaction among global architectural dimensions, both classical and MoE-specific, under MoE settings. The curse of dimensionality (the joint space of all relevant dimensions is too vast for brute-force exploration) explains why this problem has remained open. A core contribution is an efficient dimension decomposition strategy that makes holistic investigation feasible.
We first show that MoE architectures decouple computation, weight access, and parameter storage, represented by $M$, $N_a$, and $N$, respectively. Unlike dense models, where these costs largely move together, MoE configurations can have similar $M$ but substantially different $N_a$ and $N$. Fixing $M$ alone is appropriate for minimizing loss at that compute level, but does not expose the other resource costs (Sec.~\ref{subsec:fair_evaluation}). To enable fair structural optimization, we constrain the search space by a joint triad: $M$, $N_a$, and $N$.

However, this strictly constrained space exacerbates dimensionality. Incorporating $N_a$, $N$, and eleven structural hyperparameters (Table~\ref{tab:notations}) yields a 16-dimensional optimization with $\mathcal{O}(n^{16})$ cost. Our framework addresses this through systematic reduction: we first apply classical $(C,M,D)$ scaling laws to fix two dimensions, then leverage algebraic constraints and strategic parameter fixation to reduce the space to four degrees of freedom ($M, N_a, N, d$), achieving $\mathcal{O}(n^4)$.
To further reduce cost, we identify a rank-preserving property of the hidden dimension $d$: the relative performance rankings of $(M, N_a, N)$ configurations remain stable across valid choices of $d$. This allows us to use the median of $d$'s feasible range as a proxy, factorizing the search into two sequential phases: $\mathcal{O}(n^3)$ for optimal $(M, N_a, N)$ and $\mathcal{O}(n^2)$ for optimal $d$.
We validate this framework through 670+ MoE models across six compute scales ($10^{18}$ to $3\times10^{20}$ FLOPs). Our main contributions are:
\begin{itemize}
    \item \textbf{Global MoE architectural scaling laws.} We establish scaling laws encompassing global architectural parameters, not just local MoE-specific ones, deriving the complete optimal allocation $(M, D, N_a, N, d)$ and full configuration from a given compute budget $C$.
    \item \textbf{Dimension decomposition methodology.} Through mathematical constraints, the rank-preserving property, and coordinate transformation, we decompose the $\mathcal{O}(n^{16})$ search space into $\mathcal{O}(n^3) + \mathcal{O}(n^2)$. After locking $(M, N_a, N)$, the remaining structural degrees of freedom compress to a single dimension, yielding a reusable experimental framework for MoE architecture research.
    \item \textbf{Near-optimal band widens with scale.} The range of configurations yielding near-optimal loss broadens as compute increases; larger models are more tolerant of architectural deviations. This provides a quantitative basis for trade-offs between scaling law recommendations and engineering constraints.
\end{itemize}

\section{Related work}
\label{sec:related}

\paragraph{Dense scaling laws.}
\cite{kaplan2020scaling} established power-law relationships between loss, parameters, and compute. \cite{hoffmann2022training} refined this with compute-optimal (Chinchilla) laws, showing parameters and tokens must scale proportionally. Subsequent work extended these foundations to data quality~\citep{bi2024deepseek}, inference-aware optimization~\citep{sardana2023beyond}, over-training regimes~\citep{gadre2024language}, hyperparameter scaling~\citep{li2025predictable}, and downstream generalization~\citep{chen2024scaling, ruan2024observational, roberts2025compute, yang2025scaling}.

\paragraph{MoE scaling laws.}
MoE architectures~\citep{shazeer2017outrageously,lepikhin2020gshard,fedus2022switch,muennighoff2024olmoe} decouple total parameters from per-token compute via sparse routing, invalidating dense scaling laws.
Prior work has advanced along three themes.
\emph{Joint scaling:} \cite{clark2022unified} explored model-size--expert-count co-scaling; \cite{wang2024scaling} derived transferable dense-MoE scaling laws; \cite{ludziejewski2025joint} studied joint scaling of MoE size and sparsity for memory efficiency, motivating explicit treatment of parameter storage alongside compute.
\emph{Expert granularity and sparsity:} \cite{ludziejewski2024scaling} showed fine-grained experts with variable data sizes consistently outperform dense baselines; DeepSeekMoE~\citep{dai2024deepseekmoe} demonstrated advantages of shared experts; \cite{zoph2022st} identified diminishing returns beyond 256 experts; \cite{tian2025towards} proposed ``Efficiency Leverage'' to unify activation ratio and granularity effects.
\emph{Task heterogeneity:} \cite{jelassi2024mixture} and \cite{abnar2025parameters} showed MoE benefits vary across memorization vs.\ reasoning tasks.

\paragraph{Parameter allocation.}
Dense models tolerate heuristic shape choices: \cite{kaplan2020scaling} found structural impact ``minimal,'' though their data shows up to 8\% loss variance across shapes. MoE breaks this tolerance: its expanded design space requires systematic optimization, yet existing studies isolate MoE internals by freezing global architecture~\citep{clark2022unified,ludziejewski2024scaling,ludziejewski2025joint,wang2024scaling,abnar2025parameters}, neglecting cross-component interactions. Our work addresses this gap by jointly optimizing global and MoE-specific dimensions. A more detailed discussion of related work is provided in Appendix~\ref{app:related_work_full}.

\section{Preliminaries and general setup}
\label{sec:pre}

\subsection{Notations and definitions}

To maintain clarity across the theoretical derivations and empirical analysis, we summarize the primary architectural variables, macroscopic metrics, and optimization constraints in Table~\ref{tab:notations}. 

\begin{table}[htbp]
\centering
\caption{Summary of core mathematical notations and architectural variables.}
\label{tab:notations}
\renewcommand{\arraystretch}{1.0}
\begin{tabular}{cl}
\toprule
\textbf{Symbol} & \textbf{Description} \\
\midrule
\multicolumn{2}{c}{\textit{Overall Resource Metrics}} \\
$M$ & Total floating-point operations (FLOPs) per token during the forward pass \\
$N_a$ & Active parameter count per token, excluding embeddings \\
$N$ & Total parameter count of the model, excluding embeddings \\
$C$ & Total compute budget allocated for model training \\
\midrule
\multicolumn{2}{c}{\textit{Structural Parameters}} \\
$L, L_d, L_m$ & Total layer count, dense layer count, and MoE layer count \\
$d, d_{qkv}$ & Hidden state dimension and attention head dimension \\
$d_d, d_m$ & Intermediate hidden dimensions for dense FFNs and MoE FFNs \\
$N_q, N_{kv}$ & Number of query heads and key/value heads \\
$S$ & Context sequence length during training \\
\midrule
\multicolumn{2}{c}{\textit{MoE Routing Specifications}} \\
$N_e$ & Total number of experts distributed across the MoE layers \\
$K$ & Number of active experts selected per token during routing \\
\midrule
\multicolumn{2}{c}{\textit{Derived Resource Constraints}} \\
$r_a, r_d, r_m$ & Resource allocation ratios for attention, dense FFNs, and MoE components \\
\bottomrule
\end{tabular}
\end{table}

\subsection{Defining model scale via computation}

We define the total compute budget as $C = MD$, where $M$ denotes the exact forward FLOPs per token and $D$ the total training tokens. The parameter-based approximation $C \approx 6N_aD$~\citep{kaplan2020scaling, hoffmann2022training} systematically underestimates actual cost by ignoring non-parameter operations~\citep{bi2024deepseek}, particularly in MoE architectures. The exact $C = MD$ provides a faithful representation. Layer-wise formulas for computing $M$ are in Appendix~\ref{app:def_model_scale}.

\subsection{Scaling laws for optimal MoE resource allocation}
Following Chinchilla~\citep{hoffmann2022training}, we train models across a grid of configurations and fit power-law scaling laws to the observed losses.
For a given compute budget $C$, we seek the architectural hyperparameters $\theta_{arch}$ and training duration $D$ minimizing loss $L(\theta_{arch}, D)$:
\begin{equation}
    \min_{\theta_{arch}, D}  L(\theta_{arch}, D) \quad \text{s.t.} \quad  C = M(\theta_{arch}) \cdot D.
\end{equation}
Here $M(\theta_{arch})$ is FLOPs per token as a function of architecture. Directly fitting across all entangled dimensions is prohibitive; the following sections detail our multi-stage decoupling approach.

The common experimental infrastructure, including backbone architecture, datasets, and training setup, is detailed in Appendix~\ref{app:exp_infra}.
\section{Decoupling and reducing MoE scaling dimensions}
\label{sec:decoup}
\begin{figure}[!h]
  \centering
  \includegraphics[width=0.9\linewidth]{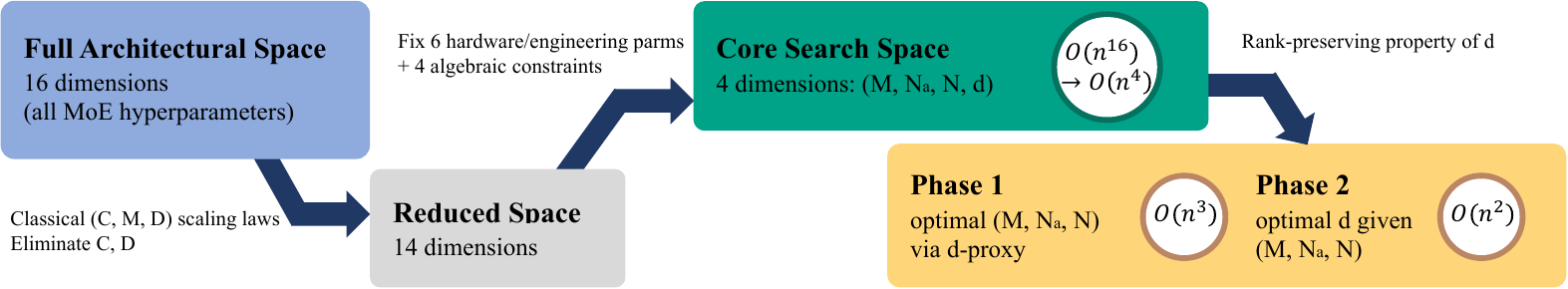} 
  \caption{Overview of the dimension reduction pipeline. Starting from the full 16-dimensional MoE architectural space, we systematically reduce the search complexity to two sequential phases of $\mathcal{O}(n^3)$ and $\mathcal{O}(n^2)$ through classical scaling laws, algebraic constraints, engineering fixations, and the rank-preserving property of $d$.}
  \label{fig:dimension_reduction_overview}
\end{figure}
To bridge the gap in MoE scaling laws for optimal resource allocation, this section presents a framework to systematically reduce the high-dimensional search space. We first address biases in MoE evaluation by emphasizing a fair comparison metric, then detail how mathematical constraints and a novel two-phase strategy transform an intractable problem into a manageable optimization.

\subsection{Why $(M, N_a, N)$ for MoE scaling}
\label{subsec:fair_evaluation}
For dense LLMs, compute, active parameters, and total parameters largely move together. However, MoE models decouple these resources through sparse routing and differing compute-to-parameter ratios of Attention and FFN layers. We use $M$, $N_a$, and $N$ as proxies for computation, weight access, and parameter storage, respectively; actual latency and bandwidth requirements also depend on the implementation and hardware. At nearly identical $M$, the three configurations in Table~\ref{tab:resource_example} span $N=5.28$--$10.02$B. Selecting the lowest-loss configuration at fixed $M$ is valid, but does not account for its parameter-storage cost. Fair architectural comparison therefore requires making the joint triad $M$, $N_a$, and $N$ explicit.

\subsection{Reduction of scaling dimensions}
\label{subsec:dimension_reduction}

Even with the $(M, N_a, N)$ triad, incorporating all eleven additional structural hyperparameters (Table~\ref{tab:notations}) yields a 16-dimensional space with a hypothetical exhaustive-grid cost of $\mathcal{O}(n^{16})$, where $n$ is the number of sampled values per dimension. Classical $(C,M,D)$ scaling laws remove two dimensions ($16\to14$); fixed architectural choices and algebraic constraints leave four free variables, $(M,N_a,N,d)$ ($14\to4$).

\subsubsection{Leveraging classical scaling laws for initial reduction}
\label{subsubsec:classical_scaling_reduction}

Our first step in dimension reduction involves re-fitting the classical $(C, M, D)$ scaling laws within our own experimental infrastructure, following the established methodology by \citep{hoffmann2022training}. This re-fitting allows us to precisely determine the optimal model size ($M$) and dataset size ($D$) for any given compute budget ($C$). Based on our experiments, we derive the following optimal formulas:
\begin{equation}
\label{eq:optimal_cmd_scaling}
\begin{aligned}
D^{\text{opt}}(C) &= 22.8929 \cdot C^{0.4563}, \\
M^{\text{opt}}(C) &= 0.04368 \cdot C^{0.5437}.
\end{aligned}
\end{equation}
By applying these derived relationships, we establish a deterministic mapping from $M$ to $C$ and $D$. This effectively eliminates two dimensions from the overarching optimization problem.
We identify six compute-optimal $(C, M, D)$ combinations as experimental checkpoints across different scales (see Appendix~\ref{app:cmd_table} for specific values). This initial reduction provides a robust and computationally aligned foundation for subsequent structural optimization.

\subsubsection{Algebraic reduction to four degrees of freedom}
\label{subsubsec:algebraic_reduction}
Building upon the initial reduction of $(C, M, D)$, we systematically resolve the remaining structural dependencies. This is achieved by anchoring the core triad of $M$, $N_a$, and $N$, and then applying a combination of exact mathematical constraints governing internal architectural parameters. Furthermore, we fix parameters that are either heavily constrained by hardware/operators or commonly considered irrelevant to conventional shape tuning in MoE models.

We fix $S{=}8192$, attention head dimensions ($N_q$, $N_{kv}$, $d_{qkv}$) to hardware-aligned powers of 2 (Appendix~\ref{app:hyper_model_config}), and adopt $K{=}8$, $N_e{=}288$ following prior work~\citep{clark2022unified,dai2024deepseekmoe,zoph2022st,abnar2025parameters} (rationale in Appendix~\ref{app:fixed_parameter_selection}). The dense FFN intermediate dimension $d_d = 3d$ is derived, not independent.

With these fixed parameters, the core architectural variables influencing the model structure are $M, N_a, N, d, L_d, L_m, d_m$. These variables are governed by the following fundamental equations:
\begin{equation}
\begin{aligned}
M &= 6N_a L + 6 S N_q d_{qkv} L, \\
N_a &= 2(N_q + N_{kv})d_{qkv}dL + 9d^2 L_d + 3dd_m(K+1)L_m, \\
N &= N_a + 3dd_m(N_e - K)L_m.
\end{aligned}
\label{eq:fundamental_equations}
\end{equation}
Here, $L = L_d + L_m$ represents the total number of layers. Given that we have 7 effective variables ($M, N_a, N, d, L_d, L_m, d_m$) and 3 independent mathematical constraints, the optimization problem is effectively transformed into a strictly bounded search space with four degrees of freedom. We choose these four degrees of freedom to be $M$, $N_a$, $N$, and the hidden dimension ($d$). The specific layer parameters ($L_d, L_m, d_m$) are thus determined by these choices and the fundamental equations.

To further simplify the exploration of this architectural space and gain a more intuitive understanding of resource allocation, we introduce compute distribution ratios that represent the proportion of total FLOPs ($M$) allocated to each component: $r_d$ for dense layers, $r_m$ for MoE layers, and $r_a$ for attention layers. The macroscopic metrics $N_a$ and $N$ can then be equivalently expressed in terms of $M, d$ and these resource ratios as:
\begin{equation}
    \begin{gathered}
        \begin{aligned}
            N_a &= \frac{1}{6} M \left(1 - \frac{r_a}{1 + \frac{2d(1 + N_{kv}/N_q)}{S}}\right), \\
            N &= N_a + \frac{1}{6} M r_m \left(\frac{N_e - K}{K + 1}\right),
        \end{aligned} \\
        \text{where } r_d + r_m + r_a = 1 \text{ and } 0 \le r_d, r_m, r_a \le 1.
    \end{gathered}
    \label{eq:architectural_constraints}
\end{equation}
This principled reduction immediately lowers the experimental complexity from $\mathcal{O}(n^{16})$ to a significantly more manageable $\mathcal{O}(n^4)$, making the exploration of MoE architectural scaling laws feasible. 

\paragraph{Generality of the reduction framework.}
It is worth noting that after locking the macroscopic parameters $(M, N_a, N)$, the remaining structural degrees of freedom are compressed to exactly one. In this paper, we choose the hidden dimension $d$ to fill this single degree of freedom. However, the framework is not limited to studying $d$: any structural ratio of interest, such as the MoE FFN intermediate width relative to the hidden size, or the proportion of dense versus MoE layers, can equivalently serve as this free variable. Sweeping over any such ratio, with $(M, N_a, N)$ held fixed, uniquely determines the full model architecture via Eq.~\eqref{eq:fundamental_equations}. This generality makes our reduction framework a reusable experimental paradigm for systematically studying arbitrary MoE structural properties under fair, scale-aligned comparisons.
We provide a concrete demonstration of this generality in Appendix~\ref{app:moe_width_ratio}, where we apply the same framework to study the scaling properties of the MoE FFN width ratio ($(K+1)\cdot d_m/d$).

\subsection{Two-phase search via rank-preserving property of $d$}
\label{subsec:efficient_search_strategy}
While the preceding algebraic reductions effectively condensed the MoE architectural design space from a prohibitive $\mathcal{O}(n^{16})$ to a more manageable $\mathcal{O}(n^4)$ defined by the four degrees of freedom ($M, N_a, N, d$), conducting an exhaustive grid search across these remaining dimensions is still computationally expensive. To further optimize the exploration process and reduce experimental costs, we introduce an efficient search strategy predicated on an empirical property of the hidden dimension $d$: its rank-preserving characteristic.

\subsubsection{Rank-preserving property of the hidden dimension ($d$)}

We observe that the relative performance rankings of $(M, N_a, N)$ configurations remain consistent across valid choices of $d$: variability from $d$ within a fixed $(M, N_a, N)$ is much smaller than differences between distinct configurations. This means $d$ acts as a fine-tuning parameter, not a primary performance determinant. We therefore use the median of $d$'s feasible range as a proxy for comparing $(M, N_a, N)$ configurations, effectively decoupling the search for optimal $d$ from the broader $(M, N_a, N)$ optimization.

We validated the median proxy at $C=10^{18}$ across 10 configurations (144 experiments). The proxy demonstrates exceptionally high correlation with the true minimum loss (Pearson $r=0.9793$, Spearman $\rho=0.9758$, all $p<0.001$), confirming its efficacy in preserving configuration rankings. Full statistical details, visualization, and additional cross-scale validation are provided in Appendix~\ref{app:median_proxy_validation}.

\subsubsection{Two-phase sequential search strategy} Leveraging this property, we decompose the search:

\begin{itemize}
    \item \textbf{Phase 1:} Fix $d$ at its median proxy and search over $(M, N_a, N)$ to identify optimal $N_a^{opt}(M)$ and $N^{opt}(M)$, reducing cost from $\mathcal{O}(n^4)$ to $\mathcal{O}(n^3)$.
    \item \textbf{Phase 2:} For each optimal $(M, N_a, N)$, search for $d^{opt}$ within its valid range at $\mathcal{O}(n^2)$ cost.
\end{itemize}

\section{Scaling laws for $(M,N_a,N)$}
\label{sec:mnna}
\subsection{Mathematical constraints, feasible space, and ratio normalization}
\label{sec:constraints_normalization}
The four degrees of freedom $(M, N_a, N, d)$ are coupled by the constraints in Eq.~\eqref{eq:architectural_constraints}, which delineate a bounded feasible region for $N_a$ and $N$ given fixed $M$ and median-proxy $d$.
This feasible region is irregular and non-rectangular (see Appendix~\ref{app:feasible_vis} for visualization), which complicates straightforward grid sampling due to many infeasible configurations. This motivates transforming the space into a more manageable, normalized coordinate system for efficient exploration.

To facilitate the setup of sampling points and to decouple the search space from specific computational scales, we introduce two scale-invariant ratios: $M/N_a$ and $N/N_a$. As shown in Eq.~\eqref{eq:architectural_constraints}, this transformation re-frames the optimization problem from the raw quantities $(M, N_a, N)$ to a normalized coordinate system $(M, M/N_a, N/N_a)$, where $M$ remains the primary scaling factor. The ratio $M/N_a$ represents the \textbf{active compute density}, quantifying the computation (FLOPs/token) per active parameter. The ratio $N/N_a$ represents the \textbf{parameter expansion ratio}, indicating total parameters per active parameter. Here, we introduce $r_m^{(N_a)}$ to denote the proportion of active parameters specifically attributable to the MoE part.
\begin{equation}
    \label{eq:ratio_expressions}
    \begin{aligned}
        M/N_a &= \frac{6}{1 - \frac{r_a}{1 + \frac{2d(1 + N_{kv}/N_q)}{S}}}, \\
        N/N_a &= 1 + r_m^{(N_a)} \left(\frac{N_e + 1}{K + 1}-1\right).
    \end{aligned}
\end{equation}
From Eq.~\eqref{eq:ratio_expressions}, $M/N_a > 6$ and $N/N_a \in (1, \frac{N_e+1}{K+1}) = (1, 289/9)$ for our configuration. This transformation rectifies the irregular $(N_a, N)$ feasible region into a rectangular $(M/N_a, N/N_a)$ search space amenable to grid sampling.

\subsection{Experimental setup}
\label{sec:mnna_exp_confg}
We systematically sample $6 \times 6 \times 6 = 216$ configurations across six $M$ scales, six $M/N_a$ values $\{7, 8, 9, 11, 14, 17\}$, and six $N/N_a$ values $\{12, 16, 20, 22, 26, 30\}$. The sampling strategy and complete configurations are detailed in Appendix~\ref{app:sampling_strategy} and~\ref{app:config_of_all}.

\subsection{Results and scaling laws derivation}
\label{sec:mnna_scaling_laws_result}

\begin{figure}[!h]
    \centering
    \begin{minipage}[b]{0.29\textwidth}
        \centering
        \includegraphics[width=\textwidth]{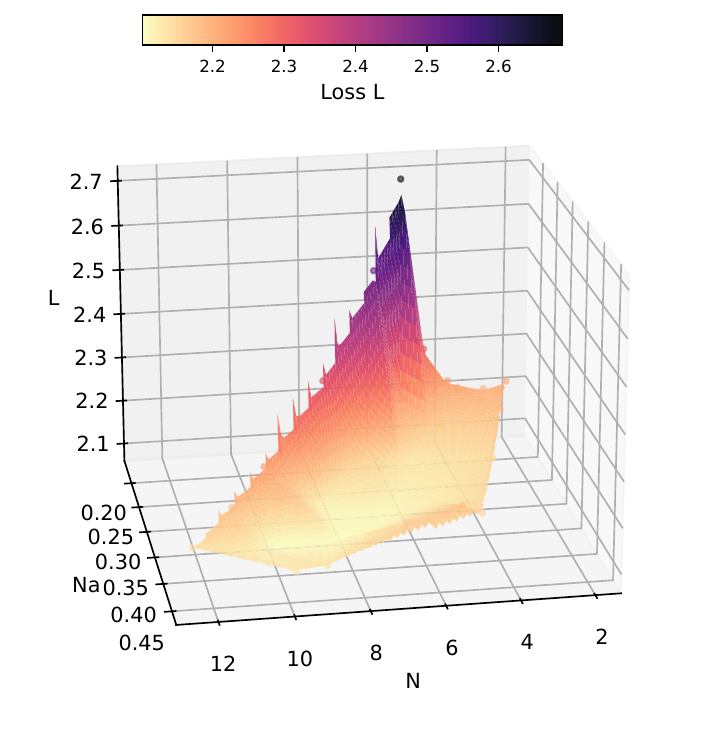} 
        \\(a) In $(N_a, N)$
        \label{fig:performance_na_n}
    \end{minipage}
    \hfill
    \begin{minipage}[b]{0.29\textwidth}
        \centering
        \includegraphics[width=\textwidth]{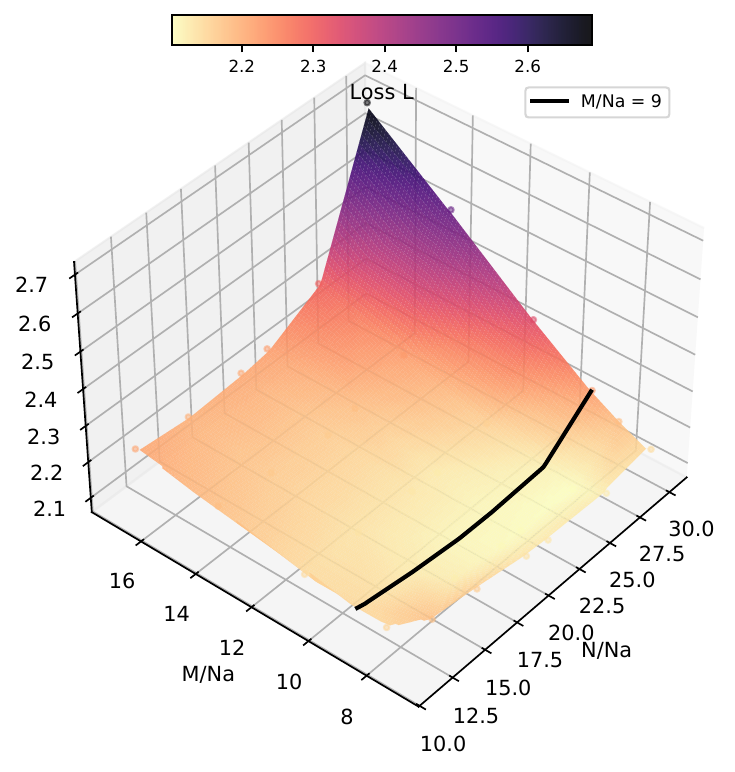} 
        \\(b) In $(M/N_a, N/N_a)$
        \label{fig:performance_mna_nna_3d}
    \end{minipage}
    \hfill
    \begin{minipage}[b]{0.38\textwidth}
        \centering
        \includegraphics[width=\textwidth]{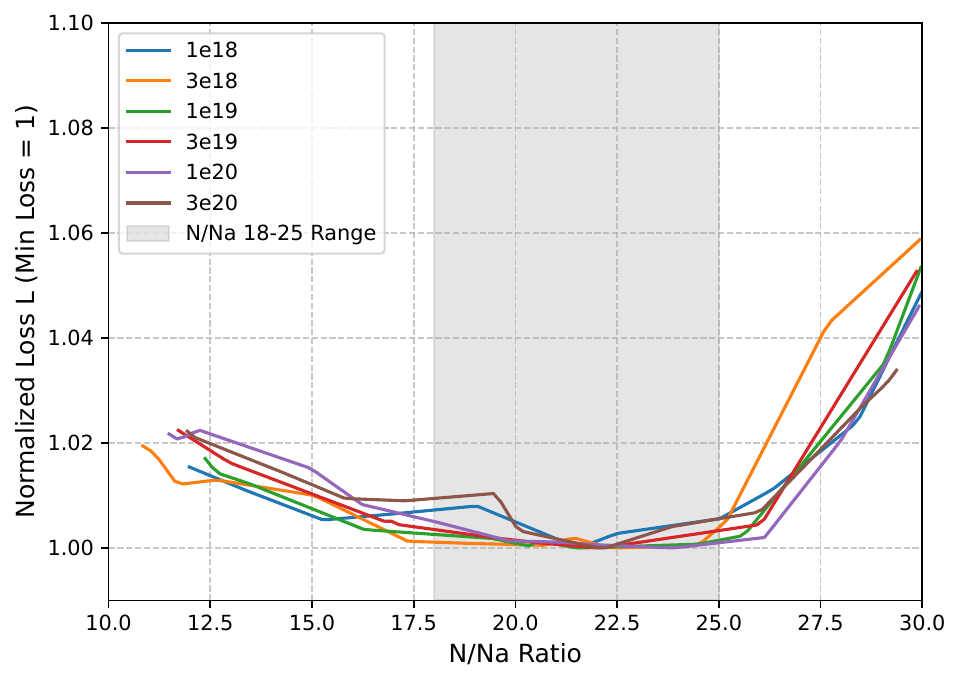} 
        \\(c) Normalized Loss $L$ vs. $N/N_a$
        \label{fig:performance_nna_2d}
    \end{minipage}
    \caption{Results visualized across different parameter spaces. (a) depicts the loss landscape in $(N_a, N)$ for a specific compute budget ($C=10^{20}$, $M=3.2681$ GFLOPs); additional visualizations for different compute scales are available in Appendix~\ref{app:vis_3d}. (b) transforms this into the $(M/N_a, N/N_a)$ space, highlighting a black profile curve within the optimal loss region. (c) presents normalized loss profiles along $N/N_a$ for various compute scales, demonstrating the consistent flatness of the loss landscape near its optimal value.}
    \label{fig:performance_vis}
\end{figure}
Figure~\ref{fig:performance_vis}(a) reveals that the optimal region in $(N_a, N)$ space is centrally located, with loss rising sharply as $N/N_a \to 1$. The transformed $(M/N_a, N/N_a)$ view (b) highlights a clear low-loss basin. Figure~\ref{fig:performance_vis}(c) shows that normalized loss remains near 1 for $N/N_a \in [18, 25]$ across all scales, demonstrating exceptional flatness. We therefore empirically fix $N/N_a$ in subsequent experiments (details in Appendix~\ref{app:n_na_setting}).

\textbf{Additional experiments for optimal $M/N_a$.} Initial experiments showed optimal $M/N_a$ decreases with scale, but sparse sampling below 8 limited fitting precision. We conducted denser sampling in this range, fixing $N/N_a$ to reduce cost (Appendix~\ref{app:config_of_all}).

\begin{figure}[!h]
    \centering
    \begin{minipage}[b]{0.48\textwidth}
        \centering
        \includegraphics[width=\textwidth]{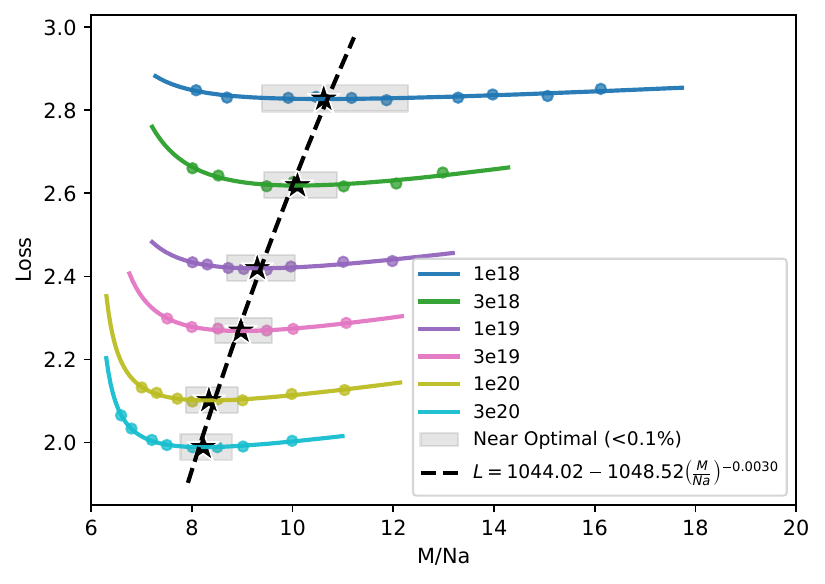}
        \\(a) Loss vs. $M/N_a$
        \label{fig:mna_fit_curves}
    \end{minipage}
    \hfill
    \begin{minipage}[b]{0.48\textwidth}
        \centering
        \includegraphics[width=\textwidth]{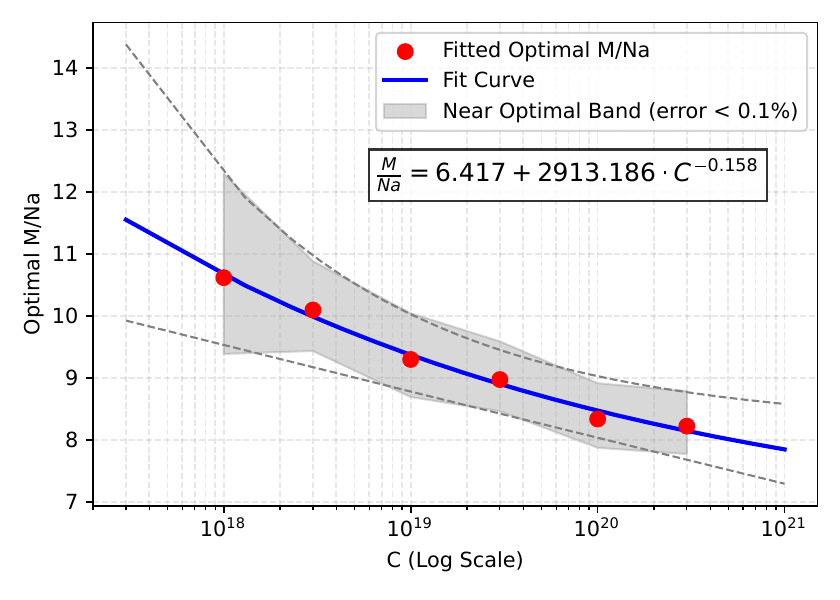}
        \\(b) Scaling of optimal $M/N_a$
        \label{fig:mna_scaling_laws}
    \end{minipage}
    \caption{Analysis of $M/N_a$ optimization. The shaded grey regions in both subfigures represent "near-optimal bands," indicating parameter combinations where the loss is within 0.1\% of the minimum for that scale, thereby implying robustness to slight deviations from the exact optimum. Specifically, (a) illustrates the loss data points and fitted curves across various compute scales, showing the relationship between loss and $M/N_a$. (b) depicts the derived scaling laws for the optimal $M/N_a$ as a function of the compute budget $C$, demonstrating how the preferred $M/N_a$ changes with increasing computational resources while remaining robust within these near-optimal bounds.}
    
    \label{fig:optimal_m_na}
\end{figure}

For each compute scale, we fitted a curve of the form $L = a / (x-6) + b x + c$ (where $x$ represents $M/N_a$) to the data points. The optimal $M/N_a$ was then determined at the minimum point of this fitted curve. Figure~\ref{fig:optimal_m_na}(a) presents these data points and their respective fitted curves. Simultaneously, these data enabled us to derive the scaling laws for the optimal $M/N_a$ as a function of $C$, which are displayed in Figure~\ref{fig:optimal_m_na}(b). The grey band in both subfigures represents the parameter space where $L/L_{\min}\leq1.001$ for that compute scale, a relative tolerance of 0.1\%, highlighting that minor inaccuracies in the fitted optimal $M/N_a$ will have negligible practical impact due to the loss landscape's flatness. As $L_{\min}$ decreases with scale, the absolute tolerance $0.001L_{\min}$ shrinks rather than expands. Having established the scaling laws for optimal $N_a$ and $N$ (indirectly via $M/N_a$) given a specific compute budget $M$, we now proceed to study the scaling laws for the optimal hidden dimension ($d$) in the next section.

\section{Scaling laws for hidden dimension ($d$)}
\label{sec:mnnad}
We now characterize the scaling laws for the hidden dimension $d$, using the six optimal $(M, N_a, N)$ configurations from the preceding section. For each, we explored a range of $d$ values, yielding $(d_i, l_i)$ datasets per compute scale (Appendix~\ref{app:config_of_all}).

\begin{figure}[!h]
    \centering
    \begin{minipage}[b]{0.48\textwidth}
        \centering
        \includegraphics[width=\textwidth]{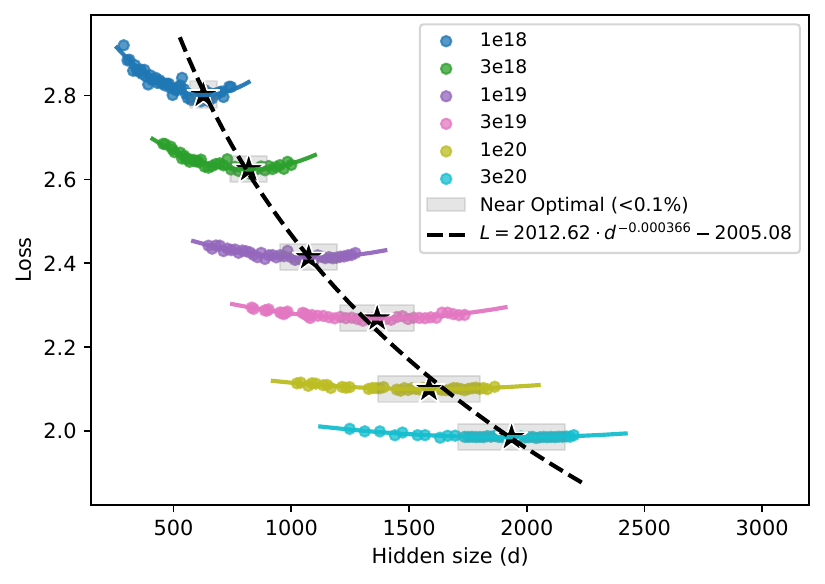}
        \\(a) Loss vs. $d$
        \label{fig:d_fit_curves}
    \end{minipage}
    \hfill
    \begin{minipage}[b]{0.48\textwidth}
        \centering
        \includegraphics[width=\textwidth]{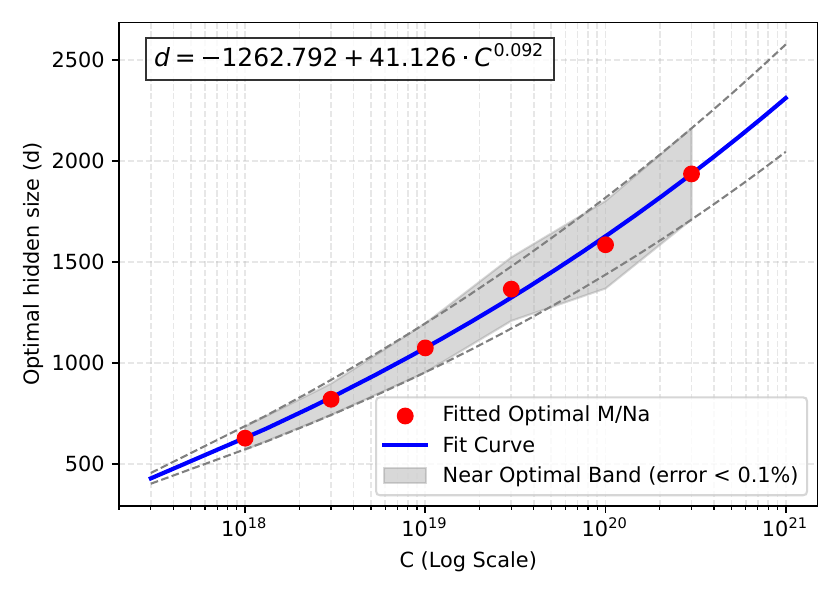}
        \\(b) Scaling of optimal $d$
        \label{fig:d_scaling_laws}
    \end{minipage}
    \caption{Analysis of $d$ optimization. The shaded grey regions in both subfigures represent "near-optimal bands," indicating parameter combinations where the loss is within 0.1\% of the minimum for that scale, thereby implying robustness to slight deviations from the exact optimum. Specifically, (a) illustrates the loss data points and fitted curves across various compute scales, showing the relationship between loss and $d$. (b) depicts the derived scaling laws for the optimal $d$ as a function of the compute budget $C$, demonstrating how the preferred $d$ changes with increasing computational resources while remaining robust within these near-optimal bounds.}
    
    \label{fig:optimal_d}
\end{figure}

A quadratic function fitted to each dataset (Figure~\ref{fig:optimal_d}(a)) reveals an optimal $d$ at the parabola minimum. Plotting $d^{\text{opt}}$ against $C$ yields a power-law relationship (Figure~\ref{fig:optimal_d}(b)). We also identify the near-optimal interval $(d_l, d_r)$ where loss remains within 0.1\% of the minimum; both $d_l$ and $d_r$ follow power laws in $C$, bounding the optimal trend and enabling flexible $d$ selection.

\paragraph{Engineering significance of the near-optimal band.}
A notable trend in Figure~\ref{fig:optimal_d} is that the near-optimal band $(d_l, d_r)$ \emph{widens} as the compute scale increases. At small scales, the interval of $d$ values yielding loss within 0.1\% of the optimum is relatively narrow, meaning that small-scale experiments are more sensitive to the choice of model shape. Conversely, at larger scales, this interval broadens considerably, indicating that large-scale models are increasingly tolerant of architectural deviations from the strict optimum.

This observation carries two practical implications. First, when designing small-scale proxy experiments to fit scaling laws, practitioners should pay particular attention to the model shape, as suboptimal choices of $d$ can introduce non-negligible noise into the loss measurements. Second, and more importantly for large-scale deployment, the widening band provides a quantitative basis for trade-offs between the scaling law recommendations and infrastructure engineering constraints. For instance, when the infrastructure team prefers to deviate from $d^{\text{opt}}$ in favor of hardware-friendly values (e.g., aligning $d$ to powers of 2 for memory efficiency), the near-optimal band directly quantifies the performance cost of such deviations. At sufficiently large scales, this cost is minimal, granting engineers significant flexibility without meaningful performance degradation.

\section{Discussion}
\label{sec:dis}

The scaling laws from Sections~\ref{sec:mnna} and~\ref{sec:mnnad} yield a practical pipeline for MoE design. Given a target compute budget $C$:
\begin{enumerate}
\item Determine optimal $M^{\text{opt}}$ and $D^{\text{opt}}$ via Eq.~\eqref{eq:optimal_cmd_scaling}.
\item Compute optimal $N_a^{\text{opt}}$ from $M/N_a^{\text{opt}}(C)$ (Figure~\ref{fig:optimal_m_na}(b)); 
\item Determine $d^{\text{opt}}$ from the power-law relationship with $C$ (Figure~\ref{fig:optimal_d}(b)).
\item Set $q=N/N_a$ to the fitted per-scale optimum in Appendix~\ref{app:n_na_setting}, then compute $N=qN_a^{\mathrm{opt}}$. For intermediate budgets, choose $q$ within the empirically flat region $[18,25]$; beyond the tested scales this remains an extrapolation.
\item Solve for remaining structural parameters via Eq.~\eqref{eq:fundamental_equations}.
\end{enumerate}
This pipeline provides a $C$-constrained design tool with $N$ as an explicit output (Table~\ref{tab:pipeline_memory}). It does not take a parameter-memory budget or an additional $M$ cap as an input; these constrained extensions and a full Pareto frontier are future work.

\paragraph{Generality of the framework.}Beyond the specific scaling laws presented in this paper, the dimension decomposition methodology itself constitutes a reusable research framework. The key structural insight is that, after locking the macroscopic parameters $(M, N_a, N)$, the remaining architectural degrees of freedom collapse to exactly one. This single degree of freedom can be filled by any structural ratio of interest, such as the hidden dimension $d$ (as studied here), the MoE FFN intermediate width relative to $d$, the proportion of dense versus MoE layers, or other design choices. Each such choice uniquely determines the complete architecture via the constraint equations, enabling controlled, fair comparisons across structural variants.

As a concrete demonstration, Appendix~\ref{app:moe_width_ratio} applies the same pipeline to the MoE FFN width ratio ($(K+1)\cdot d_m/d$), confirming seamless transfer. Future work can apply it to different sparsity configurations $(N_e, K)$, alternative structural ratios, or different modalities, provided the rank-preserving property holds.

\paragraph{Near-optimal band and engineering flexibility.}The near-optimal band (the range of parameter values yielding loss within 0.1\% of the minimum) describes tolerance along the direction being varied. The $d$ band (Figure~\ref{fig:optimal_d}) holds $(M,N_a,N)$ fixed and therefore offers no flexibility in non-embedding parameter storage. In Figure~\ref{fig:optimal_m_na}, $N$ varies inversely with $M/N_a$ at fixed $N/N_a$, exposing a trade-off between parameter storage and loss.

In practice, infrastructure teams often deviate from optimal configurations for hardware reasons (e.g., aligning dimensions to powers of 2, adjusting layer counts for pipeline parallelism). The near-optimal band provides a principled basis for evaluating such deviations: practitioners can consult the band width at their target scale to determine available flexibility. Conversely, narrower bands at small scales caution that proxy experiments for fitting scaling laws are more sensitive to shape choices.

\paragraph{Limitations.} Our experiments cover six compute scales up to $3\times10^{20}$ FLOPs, limiting confidence in extrapolation. With only six scales, more expressive fitting functions risk overfitting; we favor parsimonious forms. The $(C,M,D)$ laws were fitted with heuristic architectures and were not iteratively refitted after shape optimization. Numerical optima are specific to $K=8$ and $N_e=288$; other sparsity settings require reapplying the pipeline. Evaluation is limited to pre-training cross-entropy: whether its landscape aligns with downstream performance in architecture space remains unresolved. The framework can use another scalar objective, provided its empirical conditions, including rank preservation, are validated for that objective. Further details are in Appendix~\ref{app:limitations}.
\section{Conclusion}
\label{sec:conclu}
We presented a systematic framework for holistic MoE architectural optimization that addresses the gap between two prevalent strategies in MoE scaling research: expanding the variable space without sufficient experiments, and isolating MoE-internal parameters while ignoring global interactions. By combining algebraic constraints, a rank-preserving property of the hidden dimension, and coordinate transformations, our approach reduces the $\mathcal{O}(n^{16})$ architectural search space to two sequential phases of $\mathcal{O}(n^3)$ and $\mathcal{O}(n^2)$. The resulting scaling laws, validated across six compute scales ($10^{18}$ to $3 \times 10^{20}$ FLOPs) and over 670 model configurations, provide a deterministic pipeline from compute budget to complete MoE architecture. A key empirical finding is that the near-optimal band widens with compute scale, offering practitioners a quantitative basis for balancing scaling law recommendations against infrastructure constraints.

Beyond the specific scaling laws derived here, the dimension decomposition methodology is designed to be reusable: after locking macroscopic parameters $(M, N_a, N)$, the remaining structural degrees of freedom reduce to one, enabling controlled study of arbitrary MoE structural properties. Future work can extend this framework to alternative sparsity configurations $(N_e, K)$, downstream task evaluation, iterative refinement of $(C, M, D)$ trajectories, and multi-modal architectures.

\begin{ack}
We thank the anonymous reviewers for their constructive comments and suggestions, which helped improve this paper.
This work was supported by the Smart Grid--National Science and Technology Major Project (2026ZD0812403), the National Natural Science Foundation of China (Grant No. 62472097), the Shanghai Municipal Science and Technology Commission (Grant No. 25511102200), and the High-Quality Development Project of the Shanghai Municipal Commission of Economy and Informatization (Grant No. 2024-GZL-RGZN-02010).
\end{ack}

{
	\small
\bibliographystyle{assets/plainnat}
\bibliography{main}	
}
\newpage
\appendix
\newcommand{\mytoptitlebar}{
  \noindent\rule{\textwidth}{4pt}
  \vskip 0.16in
}
\newcommand{\mybottomtitlebar}{
  \noindent\rule{\textwidth}{1.pt}
}

\mytoptitlebar
\begin{center}
	{\Large\bf Appendix: Holistic Scaling Laws for Optimal Mixture-of-Experts Architecture Optimization }
\end{center}
\mybottomtitlebar
\vskip 0.2in

In this appendix, we provide comprehensive additional materials to supplement the main text. The contents include:
\begin{itemize}
\item \textbf{Details on defining model scale via computation (Section~\ref{app:def_model_scale}):} The exhaustive layer-wise analytical formulas used to compute $M$ for a certain MoE configuration.
\item \textbf{Rationale behind the choice of $K$ and $N_e$ (Section~\ref{app:fixed_parameter_selection}):} Explanation of the rationale behind the choice of $K$ and $N_e$.
\item \textbf{Additional visualizations (Section~\ref{app:vis_3d}):} Feasible region visualization, median proxy validation, and loss landscape visualizations for different compute scales.
\item \textbf{Broader impacts (Section~\ref{app:impacts}):} A discussion on the broader implications of our research.
\item \textbf{Case study: MoE FFN width ratio (Section~\ref{app:moe_width_ratio}):} A case study demonstrating the generality of the dimension decomposition framework by studying the scaling properties of the MoE FFN intermediate width relative to the hidden size ($(K+1)\cdot d_m/d$).
\item \textbf{Limitations (Section~\ref{app:limitations}):} A discussion of limitations of this work.
\item \textbf{Extended related work (Section~\ref{app:related_work_full}):} A detailed discussion of related work supplementing the main text.
\item \textbf{Experimental setup (Section~\ref{app:exp_setup}):} Common experimental infrastructure, compute-optimal $(C,M,D)$ combinations, hyperparameter configurations, and specific model configurations.
\end{itemize}

\section{Details on defining model scale via computation}
\label{app:def_model_scale}

This section meticulously details the methodology for quantifying the model's computational cost per token, denoted as $M$ (floating-point operations per token), and the active parameter count per token, $N_a$. These metrics are crucial for establishing the scaling laws that govern MoE architectures. Our calculations are based on layer-wise analytical formulas, ensuring a precise and reproducible accounting of computational resources as implemented in our experimental setup.

\textbf{Active parameter count ($N_a$) calculation}
The active parameter count per token, $N_a$, represents the total number of parameters effectively utilized during a single forward pass, excluding embedding parameters. This metric is composed of parameters from attention mechanisms, dense Feed-Forward Network (FFN) layers, and the active components of MoE layers. The parameter counts for individual layer components are defined as follows:

\begin{itemize}
    \item \textbf{Attention parameters ($P_{\text{attn}}$):} For each attention layer, the number of parameters is calculated as:
    \begin{equation}
        P_{\text{attn}} = 2 \cdot d \cdot d_{qkv} \cdot (N_q + N_{kv}).
        \label{eq:P_attn}
    \end{equation}
    \item \textbf{Dense FFN parameters ($P_{\text{dense\_ffn}}$):} Each dense FFN layer contributes parameters given by:
    \begin{equation}
        P_{\text{dense\_ffn}} = 3 \cdot d \cdot d_d.
        \label{eq:P_dense_ffn}
    \end{equation}
    \item \textbf{Active MoE expert parameters ($P_{\text{act\_expert}}$):} In MoE layers, only $K$ experts are activated per token. The parameters associated with these active experts within a single MoE layer are:
    \begin{equation}
        P_{\text{act\_expert}} = K \cdot (3 \cdot d \cdot d_m).
        \label{eq:P_act_expert}
    \end{equation}
    \item \textbf{Shared expert parameters ($P_{\text{shared\_expert}}$):} For MoE layers that incorporate a shared expert component, the parameters from this shared expert are:
    \begin{equation}
        P_{\text{shared\_expert}} = 3 \cdot d \cdot d_m.
        \label{eq:P_shared_expert}
    \end{equation}
\end{itemize}
Aggregating these components, the total non-embedding active parameter count ($N_a$) for a model comprising $L$ total layers, $L_d$ dense layers, and $L_m$ MoE layers is given by:
\begin{align}
    N_a &= (P_{\text{attn}} \cdot L) + (P_{\text{dense\_ffn}} \cdot L_d) + L_m \cdot (P_{\text{act\_expert}} + P_{\text{shared\_expert}}) \nonumber \\
    &= \left(2 d d_{qkv} (N_q + N_{kv})\right) L + (3 d d_d) L_d + L_m \left( K (3 d d_m) + 3 d d_m \right).
    \label{eq:Na_raw}
\end{align}

\textbf{FLOPs per token ($M$) calculation}
The total floating-point operations per token ($M$) during the forward pass encompasses the computational burden from both attention mechanisms and the feed-forward components of the model.

\textbf{Attention FLOPs ($M_{\text{attn}}$):} The GFLOPs consumed by the attention mechanisms across all layers for processing a single token are approximated as:
\begin{equation}
    M_{\text{attn}} = 6 \cdot S \cdot N_q \cdot d_{qkv} \cdot L.
    \label{eq:M_attn_raw}
\end{equation}

\textbf{Total FLOPs per token ($M$):} The overall FLOPs per token, $M$, is derived by summing the attention FLOPs with the FLOPs originating from the parameter-based linear operations (e.g., within FFNs and expert networks). A widely adopted heuristic approximates the FLOPs from these parameter-heavy operations as 6 times the active parameter count. Therefore, the total $M$ in GFLOPs is calculated as:
\begin{equation}
    M = 6 \cdot N_a + M_{\text{attn}}.
    \label{eq:M_total}
\end{equation}

\textbf{Discussion.} It is worth noting that in our default formulation, we assume the dense FFN intermediate size is a multiple of the hidden dimension, such as $d_d = 3d$. In practical engineering, however, various design choices exist. For instance, one can set $d_d = \gamma d$ where $\gamma$ is any positive constant; under this generalization, the overall derivation framework remains entirely unaffected. Alternatively, in certain architectures, it is desirable to align the computational width of the dense FFN layers with the active width of the MoE layers, imposing the constraint $d_d = d_m \cdot (K + 1)$ (where $K$ corresponds to the top-$k$ routing strategy). While such specific constraints would slightly alter the exact mathematical forms of Eq~\eqref{eq:fundamental_equations} and~\eqref{eq:architectural_constraints} in the main text, they do not fundamentally impact the final solving process. Essentially, these diverse configurations merely serve to eliminate the degree of freedom associated with $d_d$ by fixing it according to empirical architectural heuristics, ensuring the core methodology for deriving the scaling laws remains intact.

\section{Rationale behind choices of $K$ and $N_e$}
\label{app:fixed_parameter_selection}

Our choice of $K=8$ and $N_e=288$ does not fully constrain the structure of the MoE block but rather fixes its sparsity. The overall size of the MoE part remains flexible through the intermediate dimension of the MoE FFNs. Furthermore, while expert granularity ($d/d_m$) is often a subject of investigation in many MoE scaling law studies, it is also left as a free parameter in our work.
The primary rationale for fixing the sparsity ($K$ and $N_e$) is rooted in the extensive research conducted on this aspect within prior MoE scaling law literature. We directly leverage existing studies to select a high degree of sparsity. 
It is crucial to note that these specific choices for $K$ and $N_e$ are not optimized for the scale of our current experiments (ranging from $1 \times 10^{18}$ to $3 \times 10^{20}$). Instead, they are selected with an eye toward the much larger scales (e.g., $1 \times 10^{24}$) that our derived scaling laws ultimately aim to predict. This approach ensures that the scaling laws we derive are better adapted to these target compute scales. Therefore, we refer to the publicly available configurations of current advanced works operating at comparable (future) scales when making these parameter selections.
\section{Additional visualizations}
\label{app:vis_3d}

\subsection{Feasible region visualization}
\label{app:feasible_vis}

\begin{figure}[htbp]
    \centering
    \begin{minipage}[b]{0.32\textwidth}
        \centering
        \includegraphics[width=\textwidth]{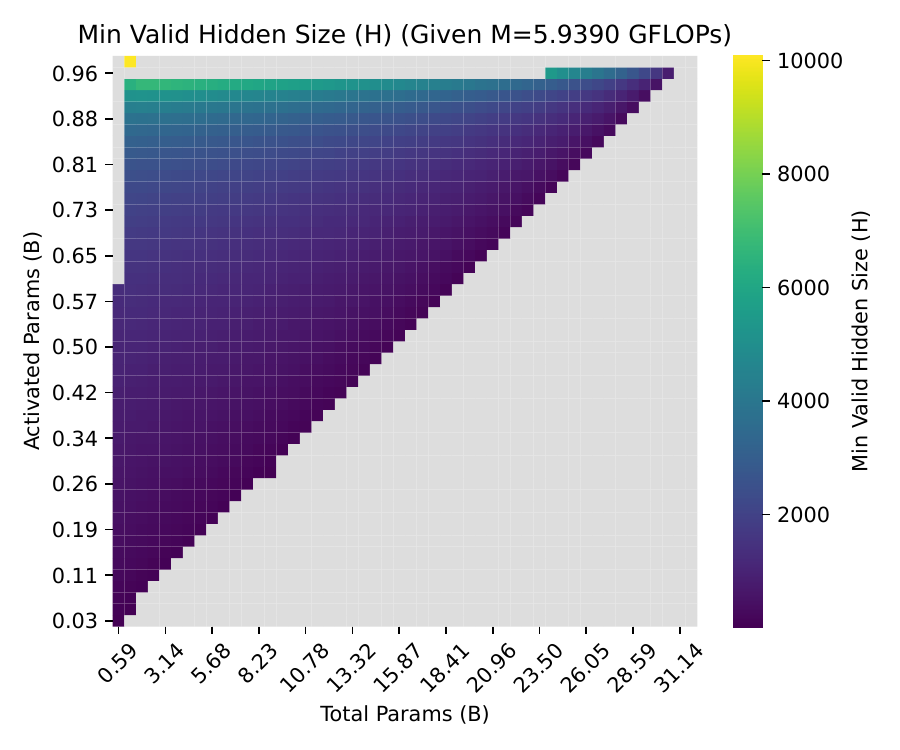}
        \\(a) minimum feasible $d$
    \end{minipage}
    \hfill
    \begin{minipage}[b]{0.32\textwidth}
        \centering
        \includegraphics[width=\textwidth]{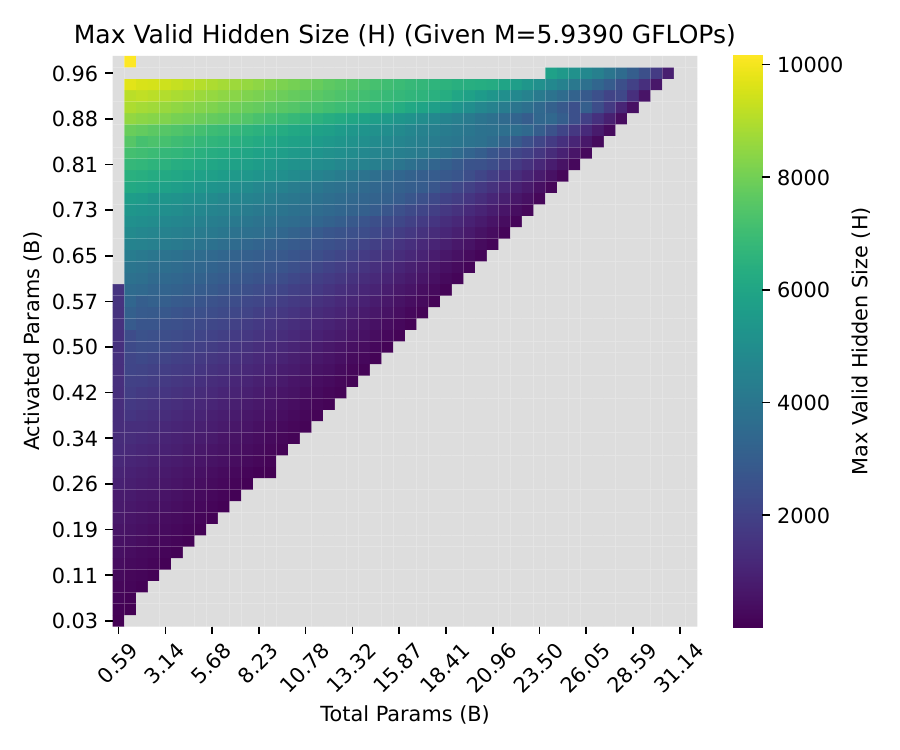}
        \\(b) maximum feasible $d$
    \end{minipage}
    \hfill
    \begin{minipage}[b]{0.32\textwidth}
        \centering
        \includegraphics[width=\textwidth]{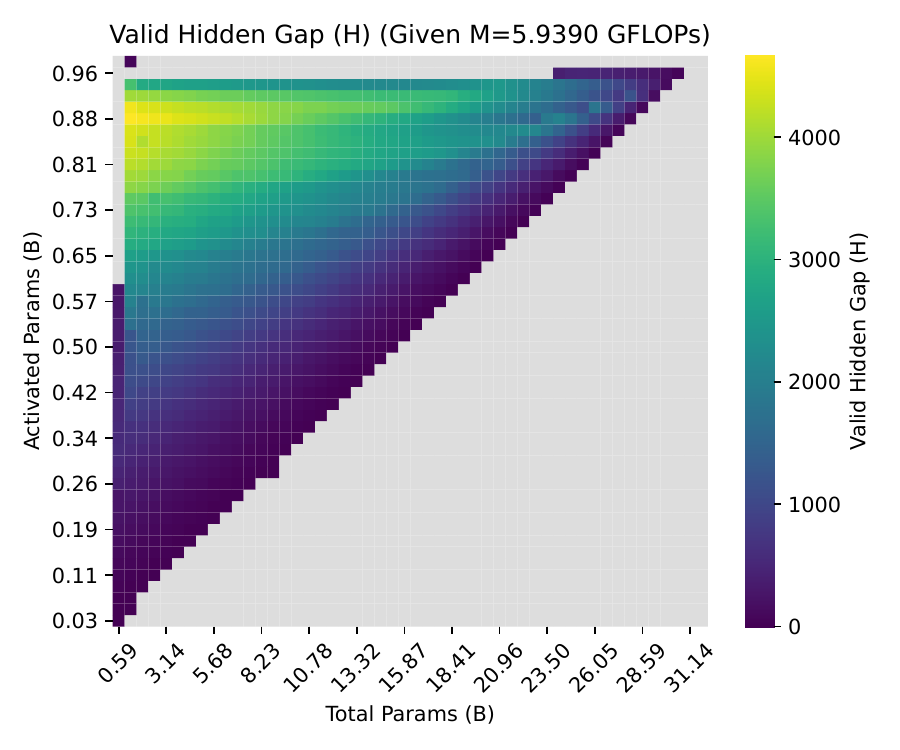}
        \\(c) the range size of $d$
    \end{minipage}
    \caption{Visualization of the mathematical feasible region for the model design space on the $(N_a, N)$ plane with $M=5.9390$ GFLOPs. (a) shows the minimum available hidden dimension $d$ for each $(N_a, N)$ point, (b) shows the maximum hidden dimension $d$, and (c) illustrates the range width of feasible $d$ values.}
    \label{fig:feasible_region_visual}
\end{figure}

\subsection{Validation of the median proxy}
\label{app:median_proxy_validation}

\begin{figure}[htbp]
  \centering
  \includegraphics[width=0.6\linewidth]{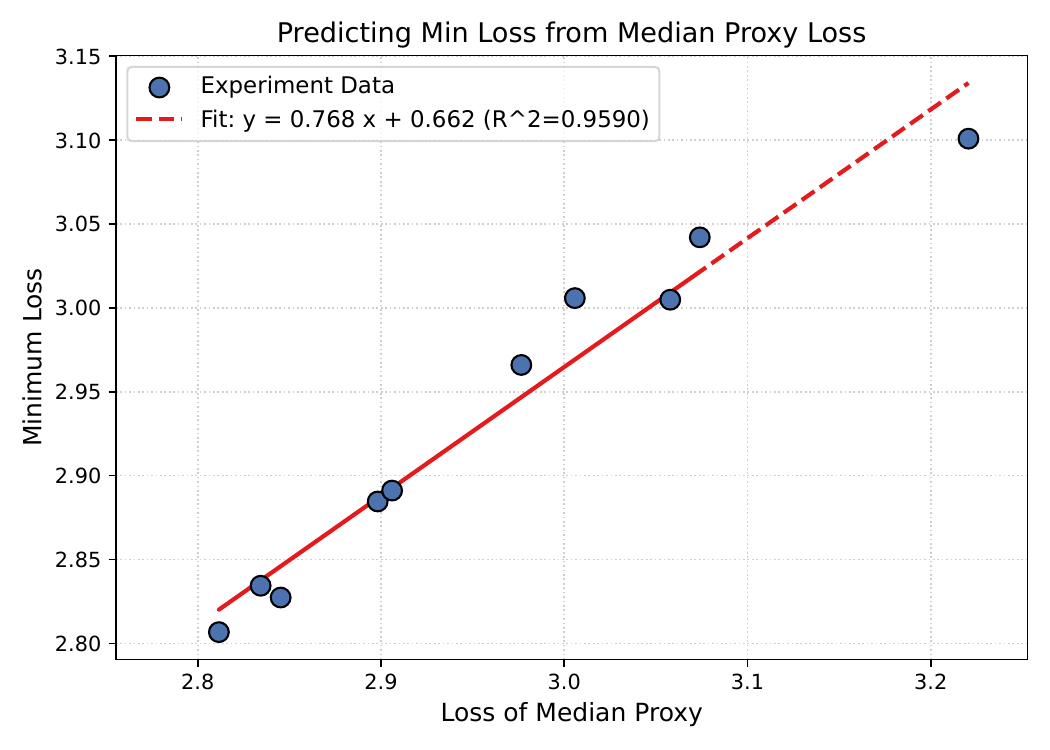}
  \caption{
    Relationship between the median proxy (median feasible $d$ proxy) and the true minimum loss under a fixed budget. The figure displays a scatter plot of the median proxy against the true minimum loss, along with a least-squares linear fit.
  }
  \label{fig:median_proxy_vs_min_loss}
\end{figure}

To validate whether the median proxy preserves the ordering and provides high predictive power for the true minimum loss, we evaluated their relationship at $C=10^{18}$ across 10 distinct configurations (144 experiments in total). Detailed experimental setup and specific model configurations can be found in Appendix~\ref{app:config_of_all}. The statistical summary of this relationship is presented in Table~\ref{tab:stat_results}, with the visual representation provided in Figure~\ref{fig:median_proxy_vs_min_loss}.

The median proxy demonstrates a robust and highly significant relationship with the true minimum loss, as detailed in Table~\ref{tab:stat_results}. All computed correlation measures (Pearson $r$, Spearman $\rho$, and Kendall $\tau$) are exceptionally high and statistically significant, with $p$-values well below conventional thresholds. This statistical evidence confirms the median proxy's efficacy in both predicting the magnitude of the minimum loss and reliably preserving the relative ranking of $(M, N_a, N)$ configurations.

\begin{table}[htbp]
\centering
\caption{Statistical summary of the relationship between the median proxy and true minimum loss.}
\label{tab:stat_results}
\begin{tabular}{lccccc}
\toprule
Metric & Pearson $r$ & Spearman $\rho$ & Kendall $\tau$ & $R^2$ \\
\midrule
Value & 0.9793 & 0.9758 & 0.9111 & 0.9590 \\
$p$-value & $7.89\times 10^{-7}$ & $1.47\times 10^{-6}$ & $2.98\times 10^{-5}$ & -- \\
\bottomrule
\end{tabular}
\end{table}

An additional analysis of the existing Phase 1 grid data compares median-$d$ proxy rankings with rankings based on the minimum loss across evaluated $d$ values, using six $M/N_a$ groups at each compute scale. Table~\ref{tab:cross_scale_proxy} reports the Pearson coefficients from this cross-scale analysis. These results supplement the separate single-scale experiment above and support the use of the median proxy across the six tested scales.

\begin{table}[htbp]
\centering
\caption{Cross-scale median-proxy validation from the additional analysis.}
\label{tab:cross_scale_proxy}
\begingroup{}
\begin{tabular}{lrrrrrr}
\toprule
$C$ (FLOPs) & $10^{18}$ & $3\times10^{18}$ & $10^{19}$ & $3\times10^{19}$ & $10^{20}$ & $3\times10^{20}$ \\
\midrule
Pearson $r$ & 0.9603 & 0.9908 & 0.9533 & 0.9845 & 0.9928 & 0.9883 \\
\bottomrule
\end{tabular}
\endgroup{}
\end{table}

\subsection{Loss landscape visualizations}
\begin{figure}[!h]
    \centering
    \begin{minipage}[b]{0.31\textwidth}
        \centering
        \includegraphics[width=\textwidth]{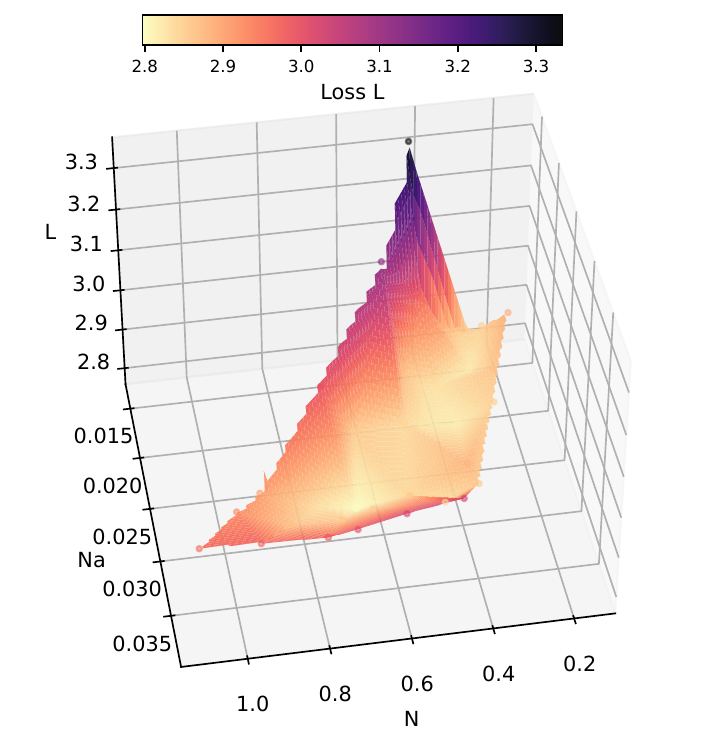} 
        \\(a) $C=1e18$
    \end{minipage}
    \hfill
    \begin{minipage}[b]{0.31\textwidth}
        \centering
        \includegraphics[width=\textwidth]{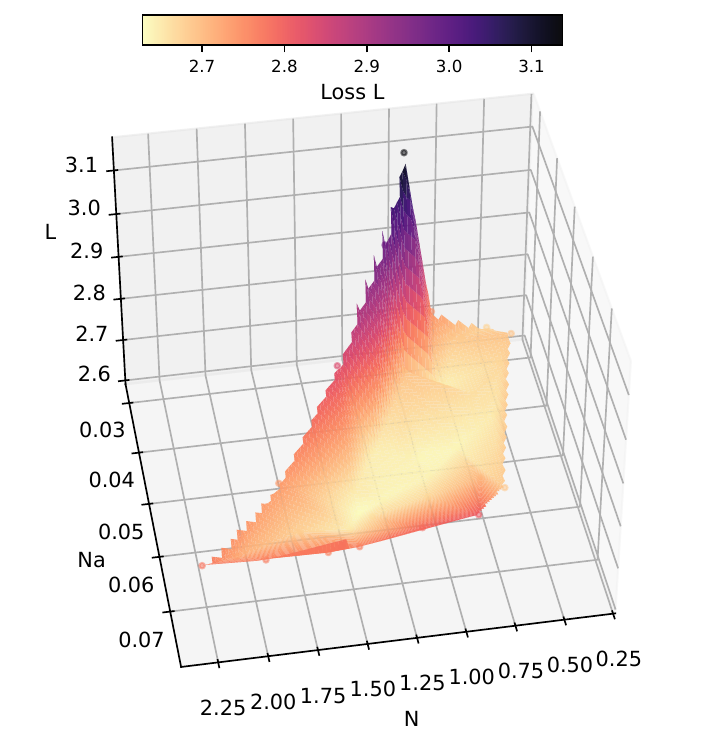}
        \\(b) $C=3e18$
    \end{minipage}
    \hfill
    \begin{minipage}[b]{0.31\textwidth}
        \centering
        \includegraphics[width=\textwidth]{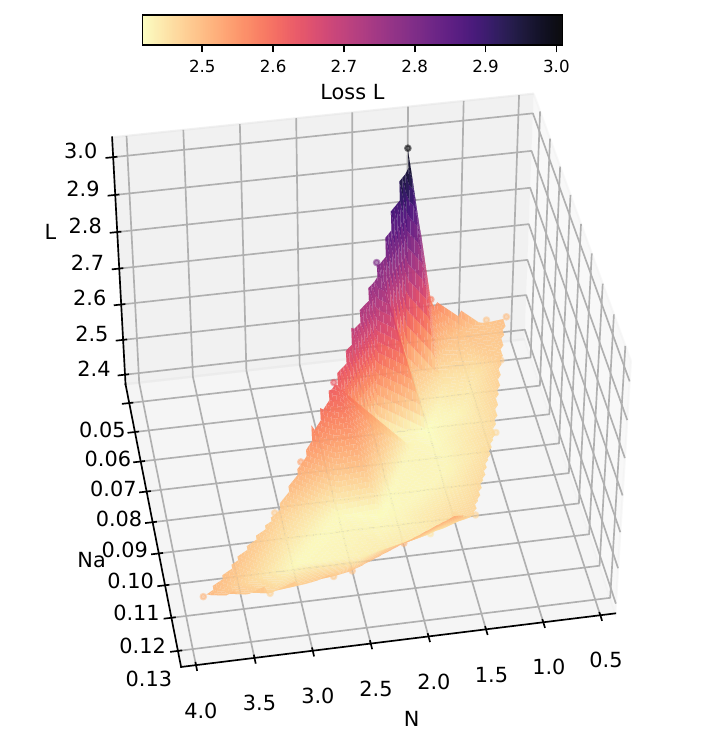}  
        \\(c) $C=1e19$
    \end{minipage}

    \begin{minipage}[b]{0.31\textwidth}
        \centering
        \includegraphics[width=\textwidth]{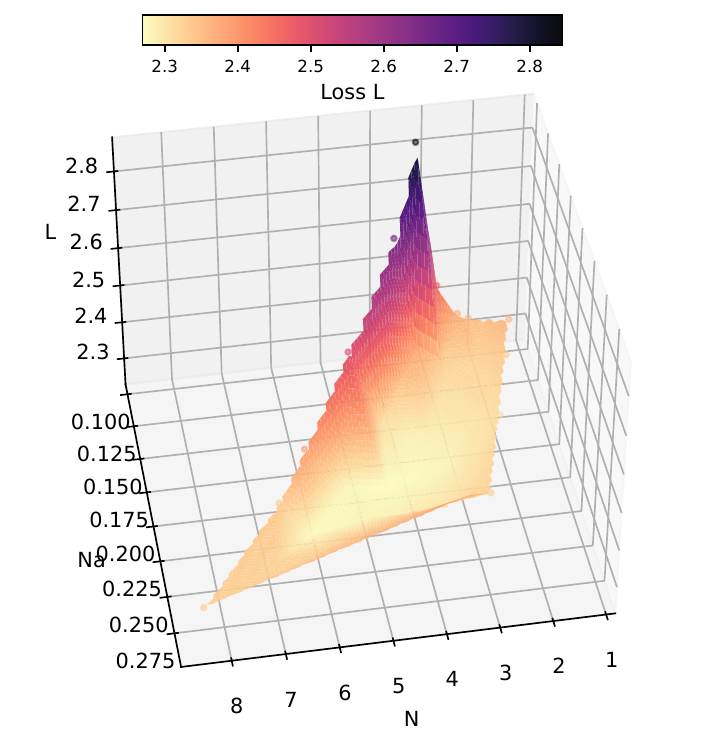} 
        \\(d) $C=3e19$
    \end{minipage}
    \hfill
    \begin{minipage}[b]{0.31\textwidth}
        \centering
        \includegraphics[width=\textwidth]{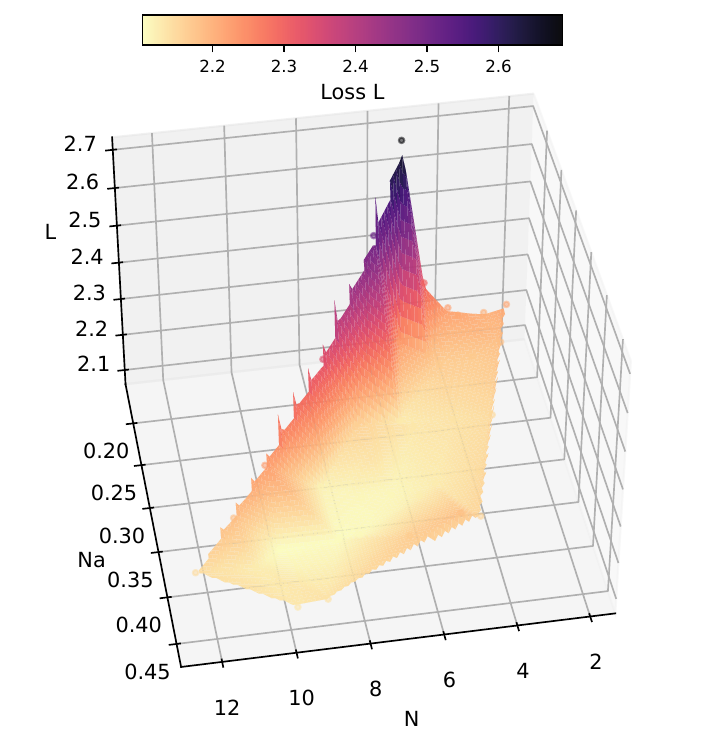}  
        \\(e) $C=1e20$
    \end{minipage}
    \hfill
    \begin{minipage}[b]{0.31\textwidth}
        \centering
        \includegraphics[width=\textwidth]{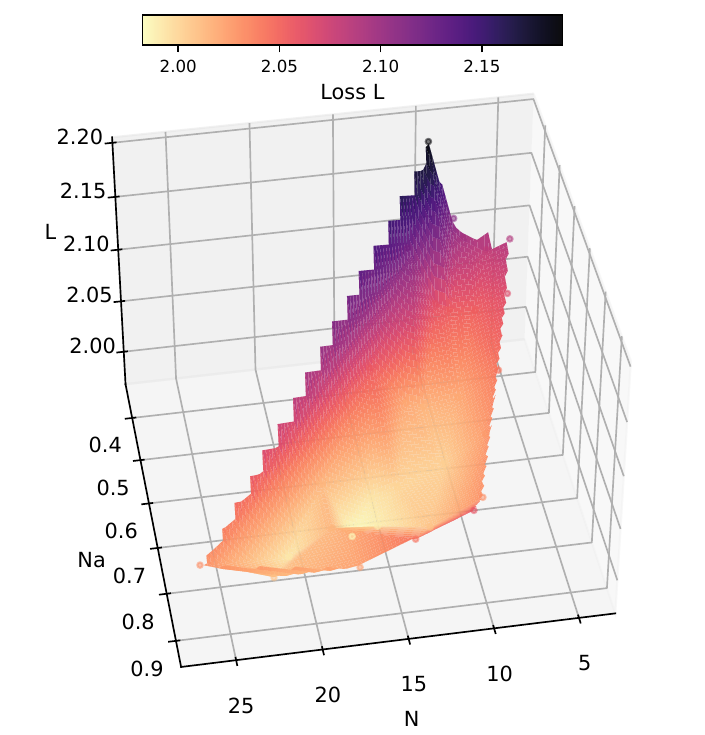}  
        \\(f) $C=3e20$
    \end{minipage}
    \caption{Additional results visualized in $(N_a, N)$ space for different compute scales.}
    \label{fig:vis_add_nna}
\end{figure}

\begin{figure}[!h]
    \centering
    \begin{minipage}[b]{0.31\textwidth}
        \centering
        \includegraphics[width=\textwidth]{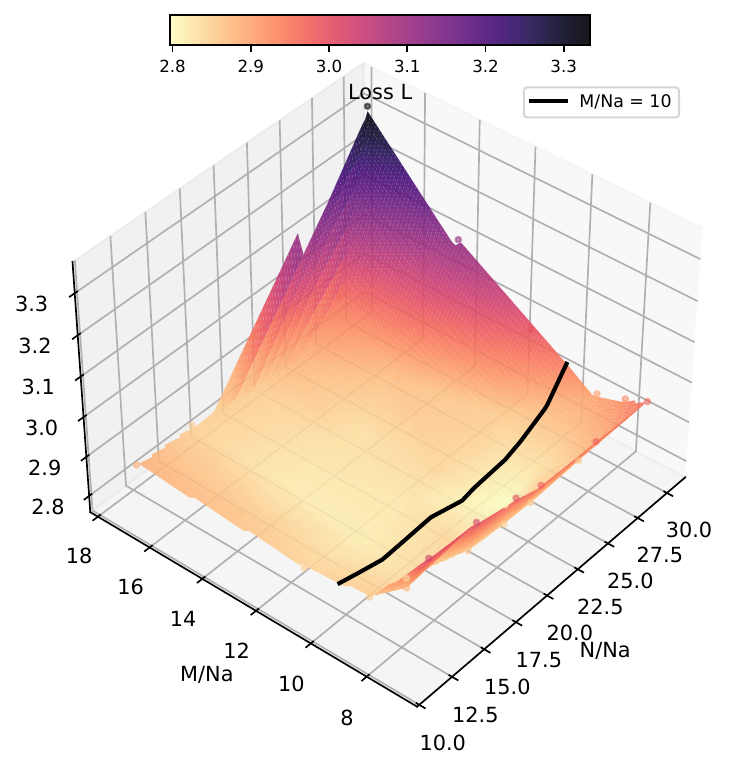}
        \\(a) $C=1e18$
    \end{minipage}
    \hfill
    \begin{minipage}[b]{0.31\textwidth}
        \centering
        \includegraphics[width=\textwidth]{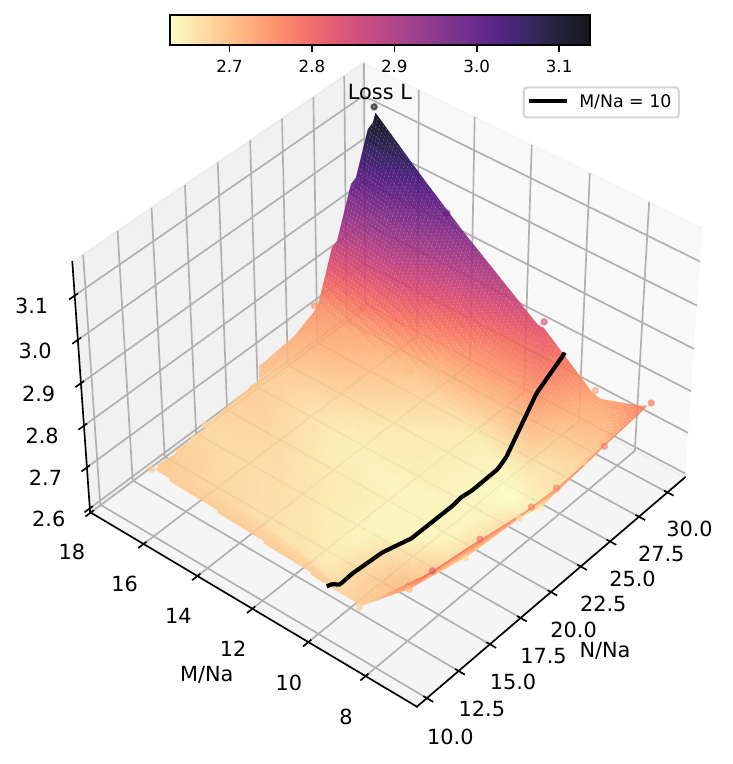}
        \\(b) $C=3e18$
    \end{minipage}
    \hfill
    \begin{minipage}[b]{0.31\textwidth}
        \centering
        \includegraphics[width=\textwidth]{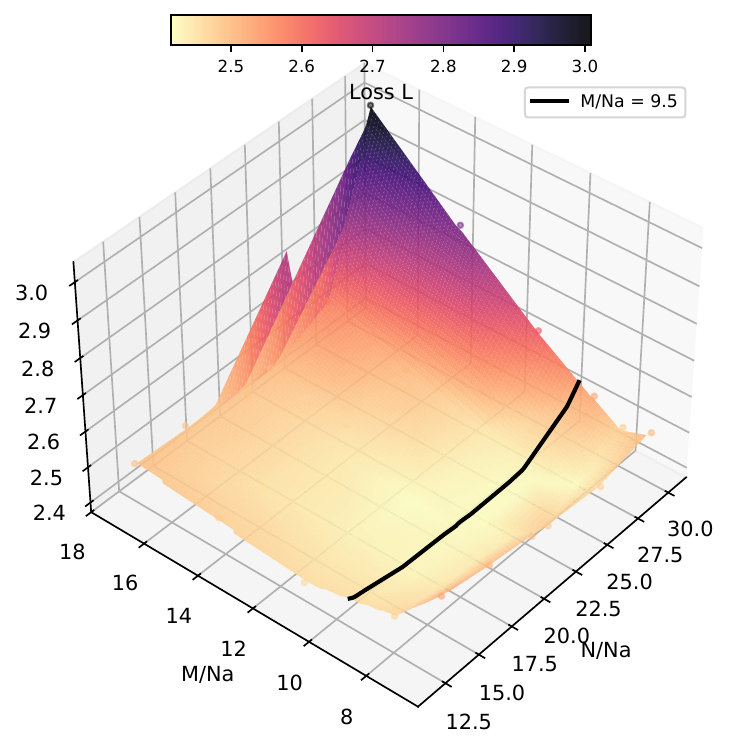}  
        \\(c) $C=1e19$
    \end{minipage}

    \begin{minipage}[b]{0.31\textwidth}
        \centering
        \includegraphics[width=\textwidth]{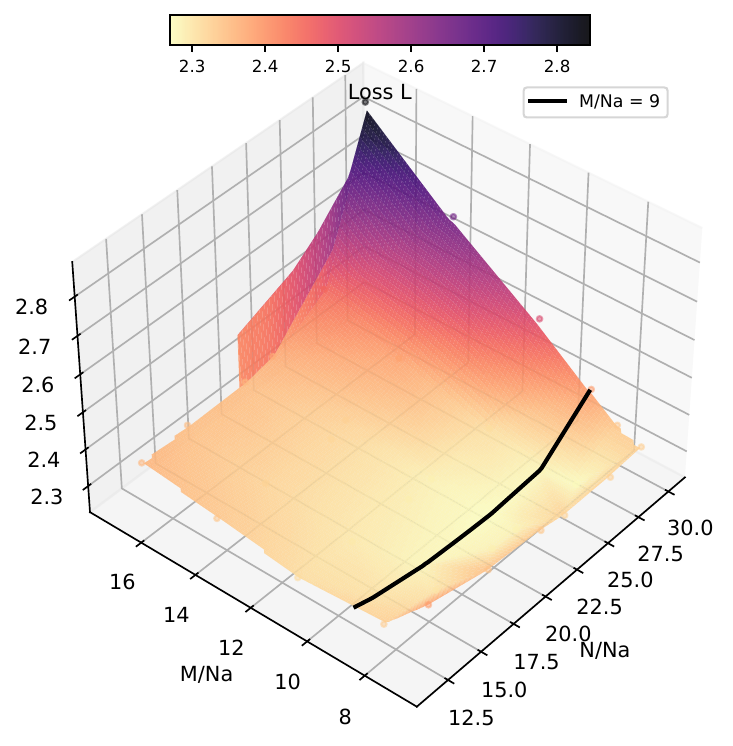}
        \\(d) $C=3e19$
    \end{minipage}
    \hfill
    \begin{minipage}[b]{0.31\textwidth}
        \centering
        \includegraphics[width=\textwidth]{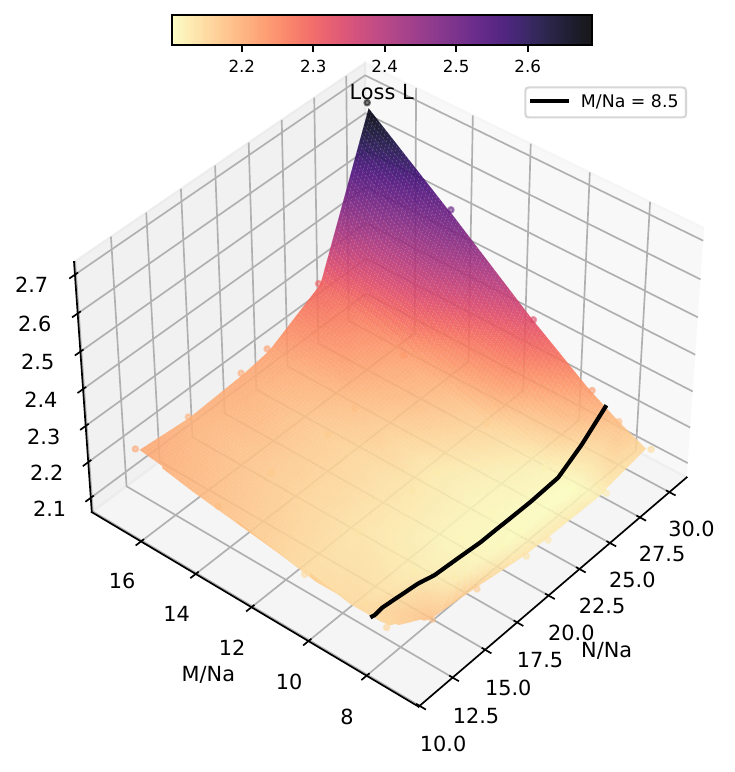}
        \\(e) $C=1e20$
    \end{minipage}
    \hfill
    \begin{minipage}[b]{0.31\textwidth}
        \centering
        \includegraphics[width=\textwidth]{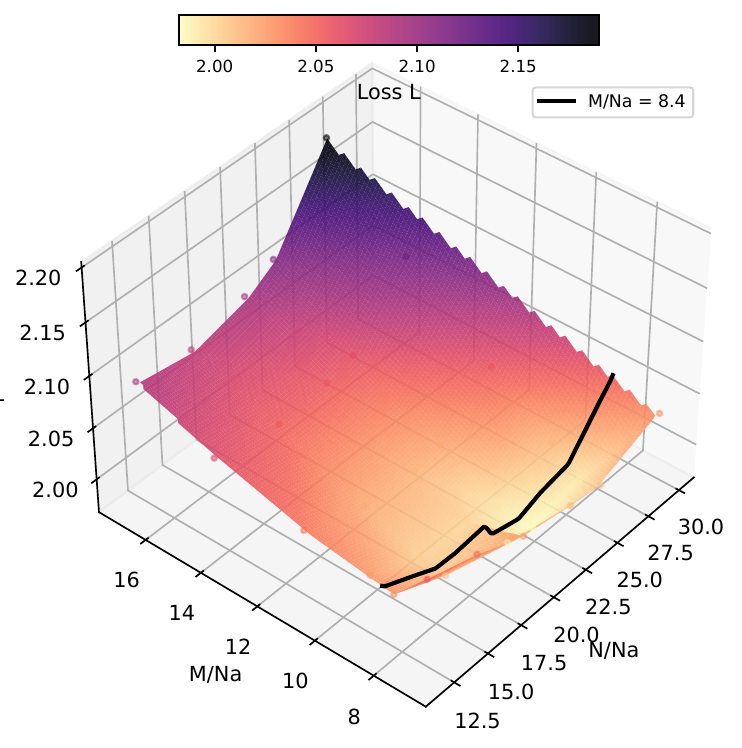}  
        \\(f) $C=3e20$
    \end{minipage}
    \caption{Additional results visualized in $(M/N_a, N/N_a)$ space for different compute scales.}
    \label{fig:vis_add_mnanna}
\end{figure}

We provide additional visualizations for different compute scales in Figure~\ref{fig:vis_add_nna} and~\ref{fig:vis_add_mnanna}.

\section{Broader impacts}
\label{app:impacts}
This work provides a framework for deriving optimal scaling laws for MoE architectures, with significant implications for large language model development.
Our research primarily contributes to enhanced efficiency in training and deploying large-scale AI models by optimizing MoE architectural design. This minimizes computational waste and accelerates development, which is crucial for mitigating the energy consumption associated with large models. By providing clear design principles, this framework also helps to democratize access to advanced AI, allowing researchers with more limited computational resources to explore MoE models effectively. Furthermore, it deepens the fundamental understanding of scaling behaviors in complex neural architectures, advancing theoretical and empirical research in model design.
While our work promotes efficiency and accessibility, the ability to design more powerful MoE models could exacerbate existing inequalities if the benefits are concentrated among a few well-resourced entities. Even with optimized guidelines, substantial computational infrastructure might still be required. Additionally, the enhanced capabilities of these optimized MoE models could potentially be misused, for instance, in generating highly convincing misinformation. These ethical implications necessitate ongoing vigilance and responsible development practices from the AI community.
In summary, this research offers valuable tools for building more efficient and capable MoE models, positively contributing to AI advancement by reducing computational costs and clarifying design principles. We emphasize the importance of continued ethical consideration and equitable access to ensure these advancements benefit society broadly.

\section{Case study: scaling properties of MoE FFN width ratio}
\label{app:moe_width_ratio}

In Section~\ref{sec:decoup}, we established that after locking the macroscopic parameters $(M, N_a, N)$, the remaining structural degrees of freedom collapse to exactly one. The main text fills this degree of freedom with the hidden dimension $d$ and derives its scaling law. To demonstrate the generality of this framework, we present an additional case study using a different structural ratio: the MoE FFN intermediate width relative to the hidden size, $(K+1)\cdot d_m / d$. This ratio governs the capacity of individual expert FFN layers and is a natural design knob for MoE architectures.

\paragraph{Experimental setup.}
We fix the six optimal $(M, N_a, N)$ configurations obtained in Section~\ref{sec:mnna_scaling_laws_result} (corresponding to compute scales $C \in \{10^{18}, 3\times10^{18}, 10^{19}, 3\times10^{19}, 10^{20}, 3\times10^{20}\}$ FLOPs) and vary $(K+1)\cdot d_m/d$ as the sole free variable, sampling five values: $\{3.0, 3.5, 4.0, 4.5, 5.0\}$. This yields $6 \times 5 = 30$ model configurations in total. For each configuration, we train to convergence and record the final loss. We then fit each $(K+1)\cdot d_m/d$ group's loss-versus-$C$ trajectory using two functional forms:
\begin{itemize}
    \item \textbf{Model 1 (E=0):} $\log L = a + b \log C$, equivalent to $L = 10^a \cdot C^b$, the standard power-law without irreducible loss.
    \item \textbf{Model 2 (E$\neq$0):} $L = E + A \cdot C^B$, a three-parameter power law that accounts for an irreducible loss floor $E$.
\end{itemize}

\begin{figure}[htbp]
    \centering
    \begin{minipage}[b]{0.48\textwidth}
        \centering
        \includegraphics[width=\textwidth]{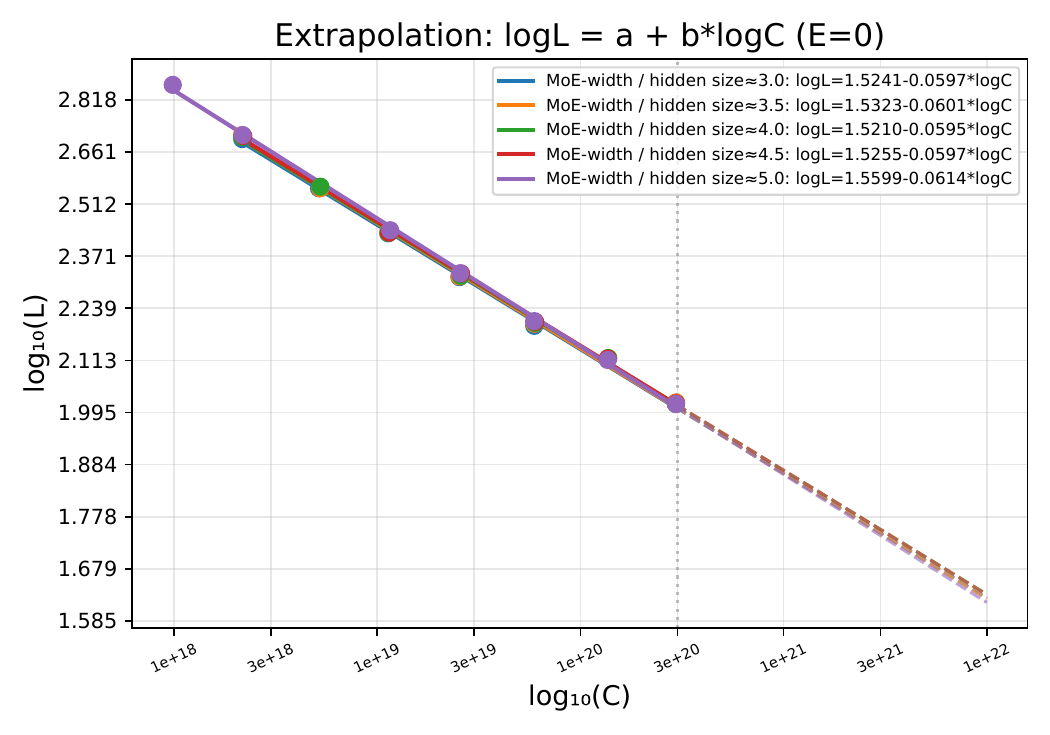}
        \\(a) Model 1: $\log L = a + b\log C$ (E=0)
    \end{minipage}
    \hfill
    \begin{minipage}[b]{0.48\textwidth}
        \centering
        \includegraphics[width=\textwidth]{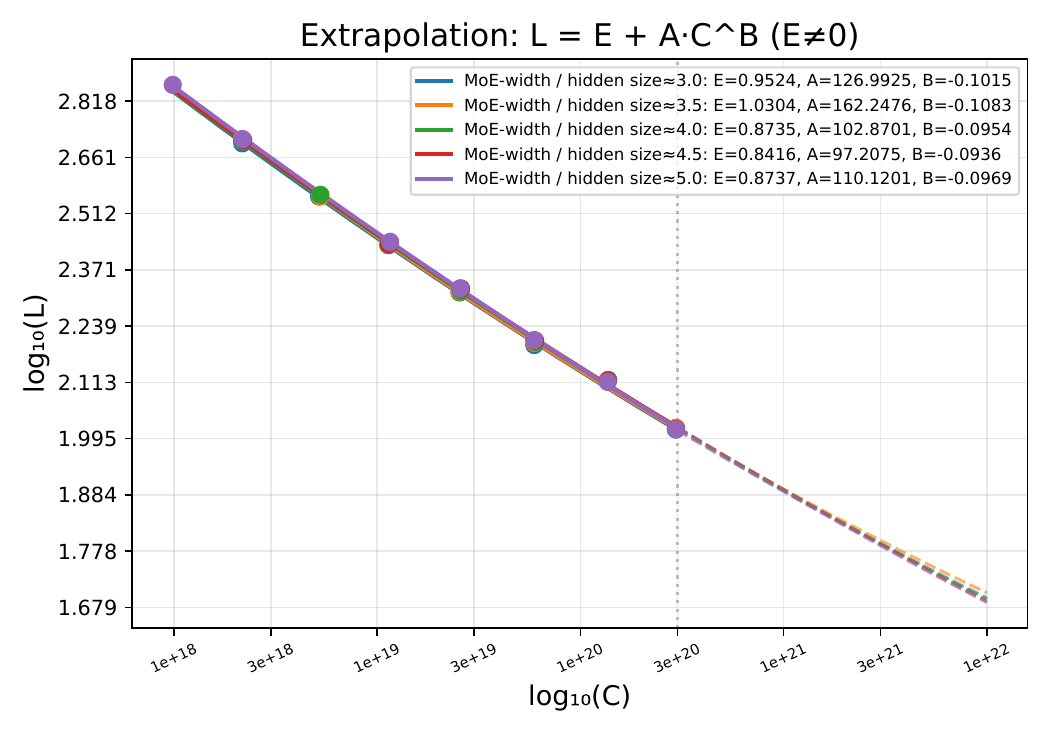}
        \\(b) Model 2: $L = E + A \cdot C^B$ (E$\neq$0)
    \end{minipage}
    \caption{Scaling curves and extrapolation for different MoE FFN width ratios. Solid lines represent fits within the experimental range ($10^{18}$ to $3\times10^{20}$ FLOPs); dashed lines show extrapolation to $10^{22}$ FLOPs. The vertical dotted line marks the boundary of the experimental range.}
    \label{fig:moe_width_extrapolation}
\end{figure}

\paragraph{Scaling curves and extrapolation.}
Figure~\ref{fig:moe_width_extrapolation} presents the fitted scaling curves for all five $(K+1)\cdot d_m/d$ groups under both models, along with their extrapolation beyond the experimental range (dashed lines, extending to $C = 10^{22}$). Within the experimental range ($10^{18}$ to $3 \times 10^{20}$), both models produce nearly indistinguishable fits, and the scaling trajectories of different $(K+1)\cdot d_m/d$ groups are closely clustered. Beyond $3\times10^{20}$, the two models diverge: the E=0 model continues declining without bound, while the E$\neq$0 model asymptotically approaches the irreducible loss $E$, yielding more conservative extrapolations.

\begin{figure}[htbp]
    \centering
    \begin{minipage}[b]{0.32\textwidth}
        \centering
        \includegraphics[width=\textwidth]{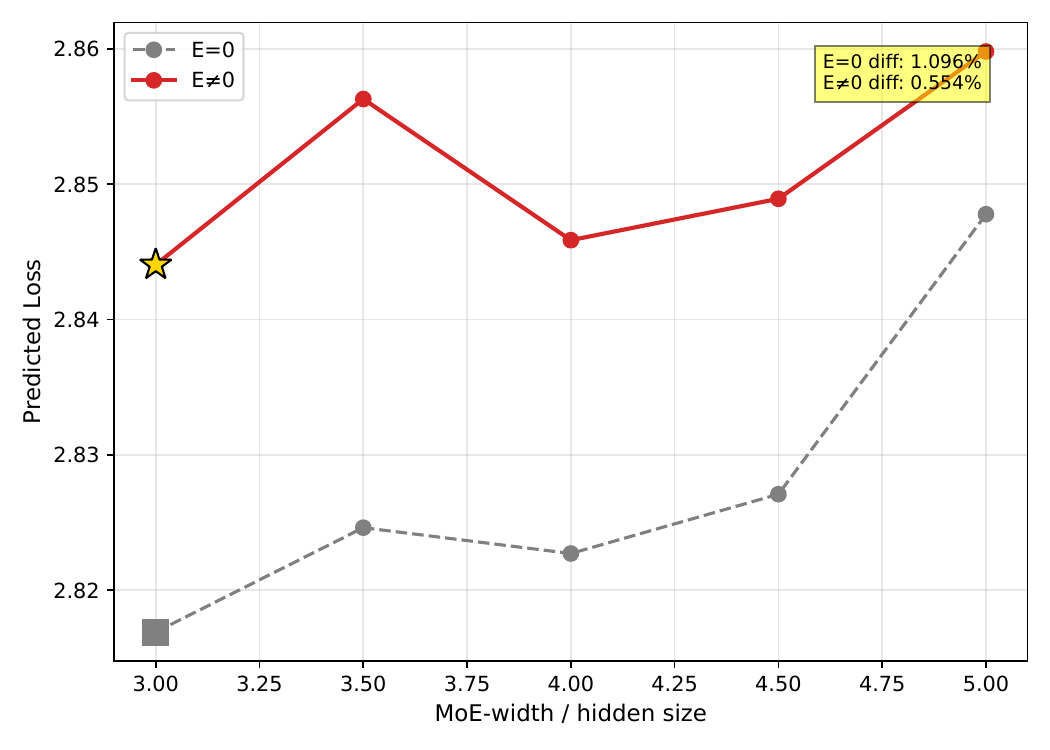}
        \\(a) $C = 10^{18}$
    \end{minipage}
    \hfill
    \begin{minipage}[b]{0.32\textwidth}
        \centering
        \includegraphics[width=\textwidth]{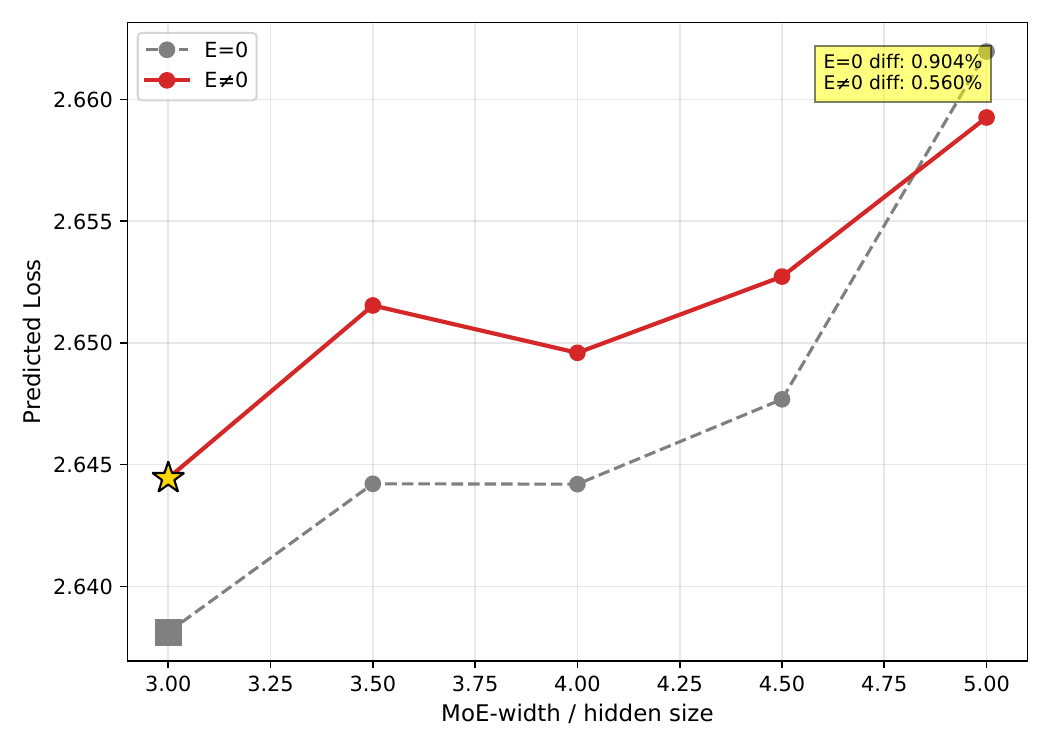}
        \\(b) $C = 3\times10^{18}$
    \end{minipage}
    \hfill
    \begin{minipage}[b]{0.32\textwidth}
        \centering
        \includegraphics[width=\textwidth]{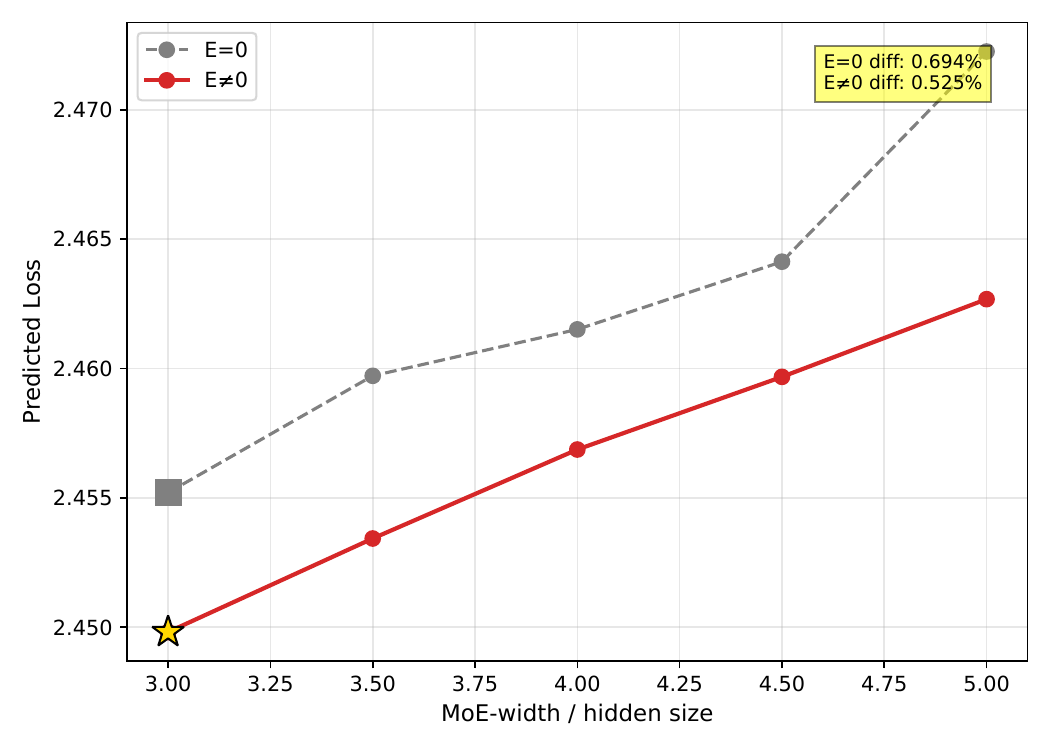}
        \\(c) $C = 10^{19}$
    \end{minipage}
    \\[6pt]
    \begin{minipage}[b]{0.32\textwidth}
        \centering
        \includegraphics[width=\textwidth]{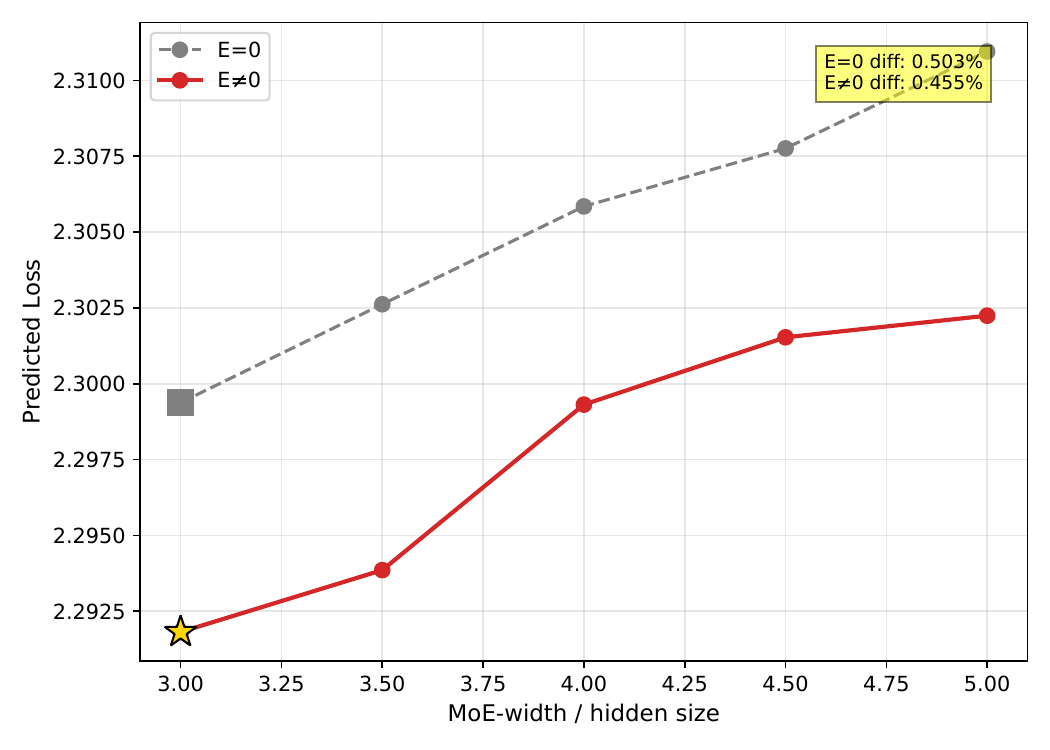}
        \\(d) $C = 3\times10^{19}$
    \end{minipage}
    \hfill
    \begin{minipage}[b]{0.32\textwidth}
        \centering
        \includegraphics[width=\textwidth]{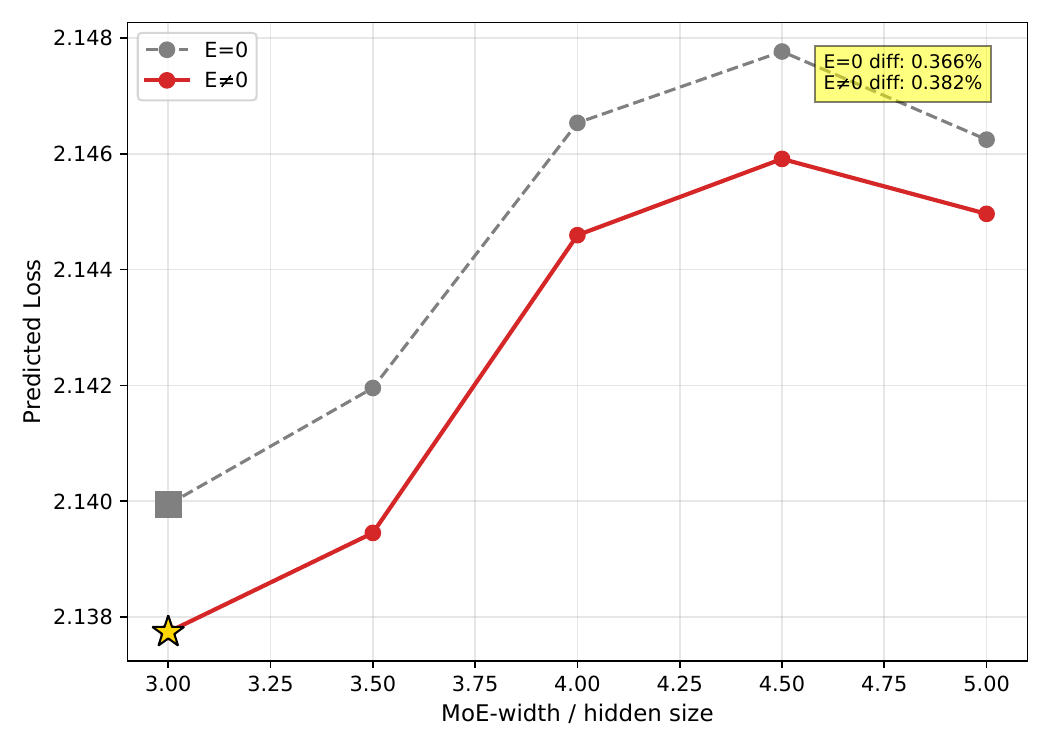}
        \\(e) $C = 10^{20}$
    \end{minipage}
    \hfill
    \begin{minipage}[b]{0.32\textwidth}
        \centering
        \includegraphics[width=\textwidth]{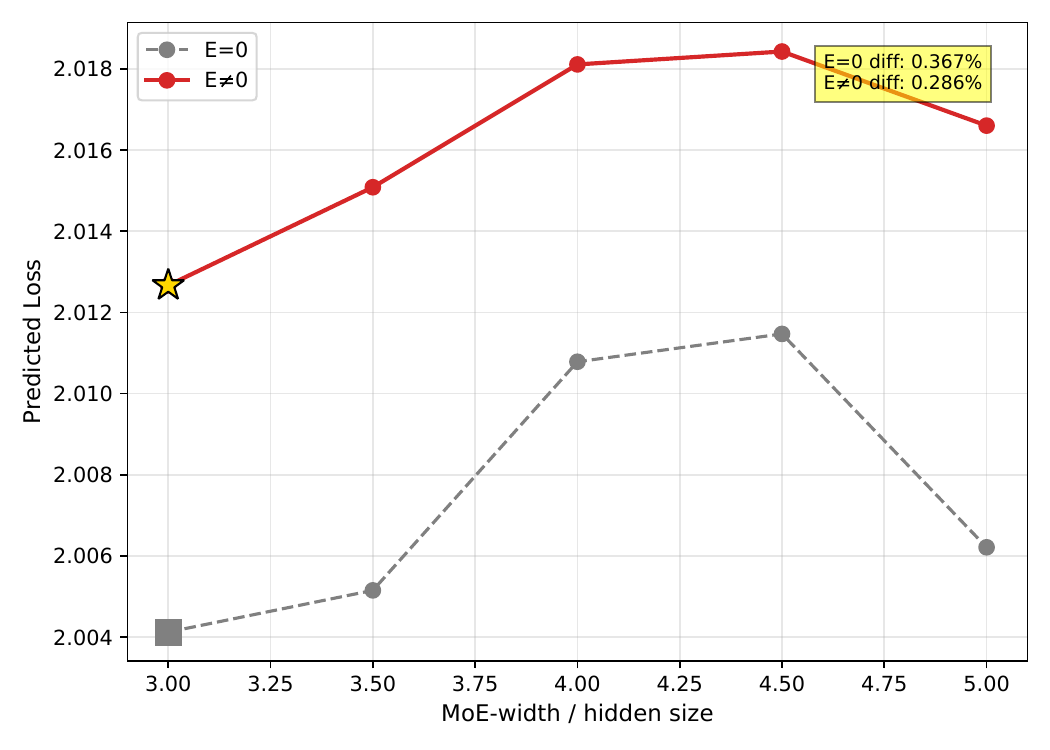}
        \\(f) $C = 3\times10^{20}$
    \end{minipage}
    \caption{Predicted loss vs.\ MoE FFN width ratio at each experimental compute scale, comparing E=0 (gray dashed) and E$\neq$0 (red solid) model predictions. The loss variation across different $(K+1)\cdot d_m/d$ values is consistently below $0.5\%$.}
    \label{fig:moe_width_by_C_exp}
\end{figure}

\paragraph{Optimal ratio within the experimental range.}
Figure~\ref{fig:moe_width_by_C_exp} displays the predicted loss as a function of $(K+1)\cdot d_m/d$ at each of the six experimental compute scales, comparing E=0 and E$\neq$0 predictions. Across all scales, the loss variation among different $(K+1)\cdot d_m/d$ values is remarkably small (less than $0.5\%$), with the optimum consistently near $d_m/d \approx 3.0$--$3.5$. The two models agree closely within this range, and both identify similar optimal ratios.

\begin{figure}[htbp]
    \centering
    \begin{minipage}[b]{0.32\textwidth}
        \centering
        \includegraphics[width=\textwidth]{figs/C_3e20.pdf}
        \\(a) $C = 3\times10^{20}$
    \end{minipage}
    \hfill
    \begin{minipage}[b]{0.32\textwidth}
        \centering
        \includegraphics[width=\textwidth]{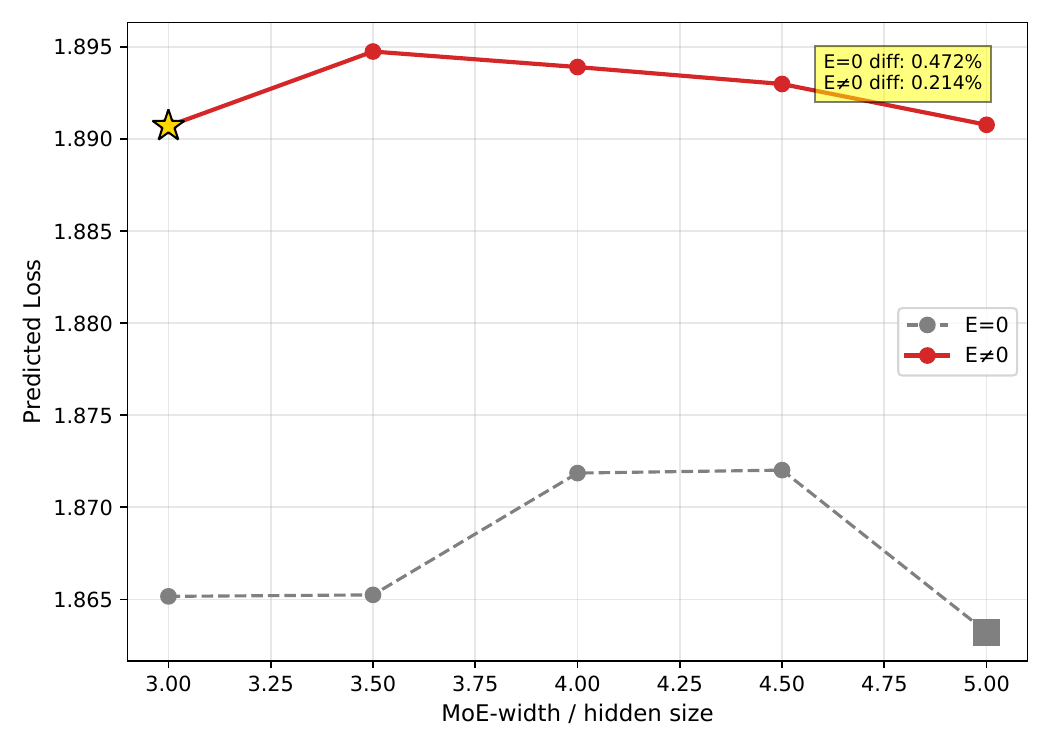}
        \\(b) $C = 10^{21}$
    \end{minipage}
    \hfill
    \begin{minipage}[b]{0.32\textwidth}
        \centering
        \includegraphics[width=\textwidth]{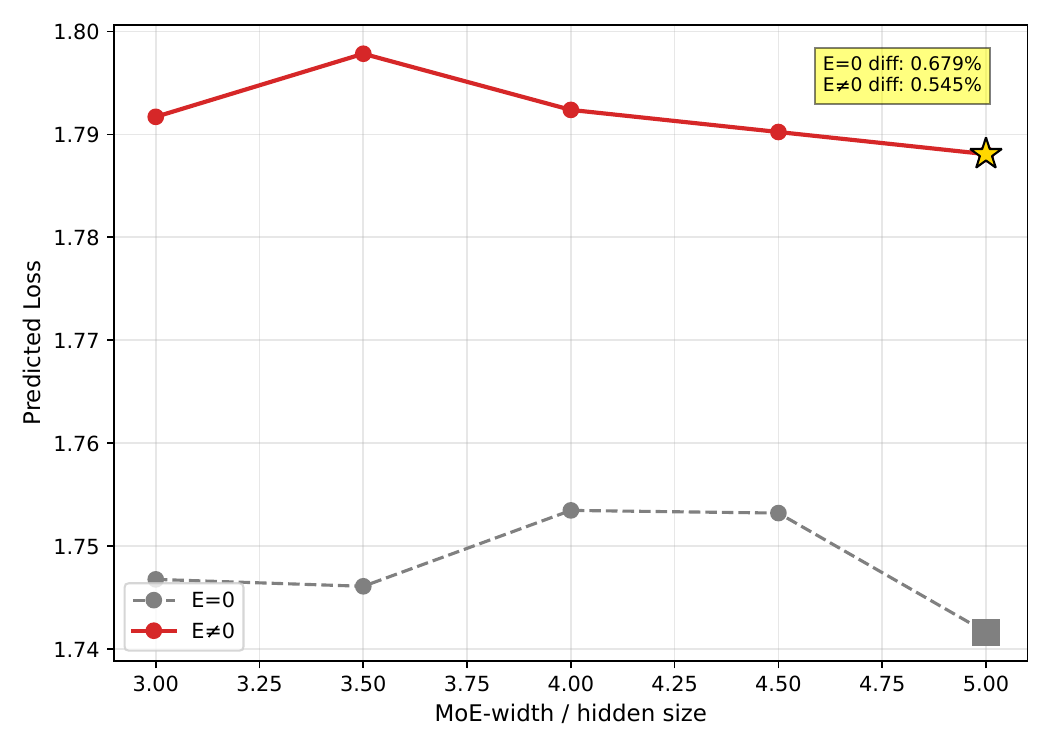}
        \\(c) $C = 3\times10^{21}$
    \end{minipage}
    \\[6pt]
    \begin{minipage}[b]{0.32\textwidth}
        \centering
        \includegraphics[width=\textwidth]{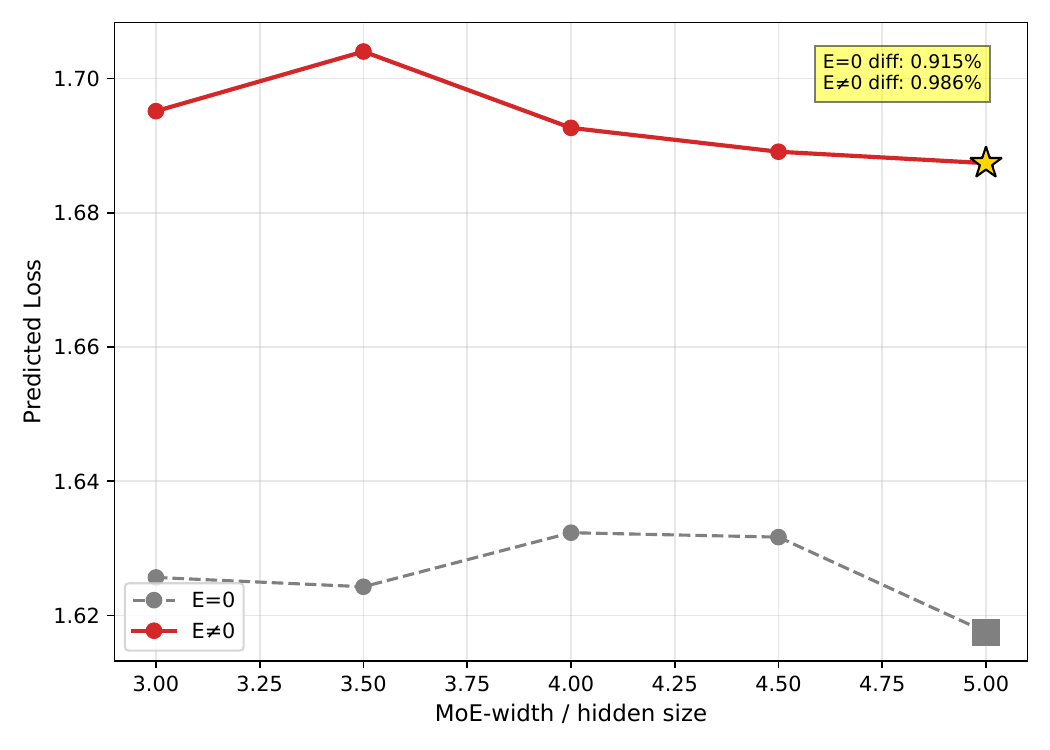}
        \\(d) $C = 10^{22}$
    \end{minipage}
    \hfill
    \begin{minipage}[b]{0.32\textwidth}
        \centering
        \includegraphics[width=\textwidth]{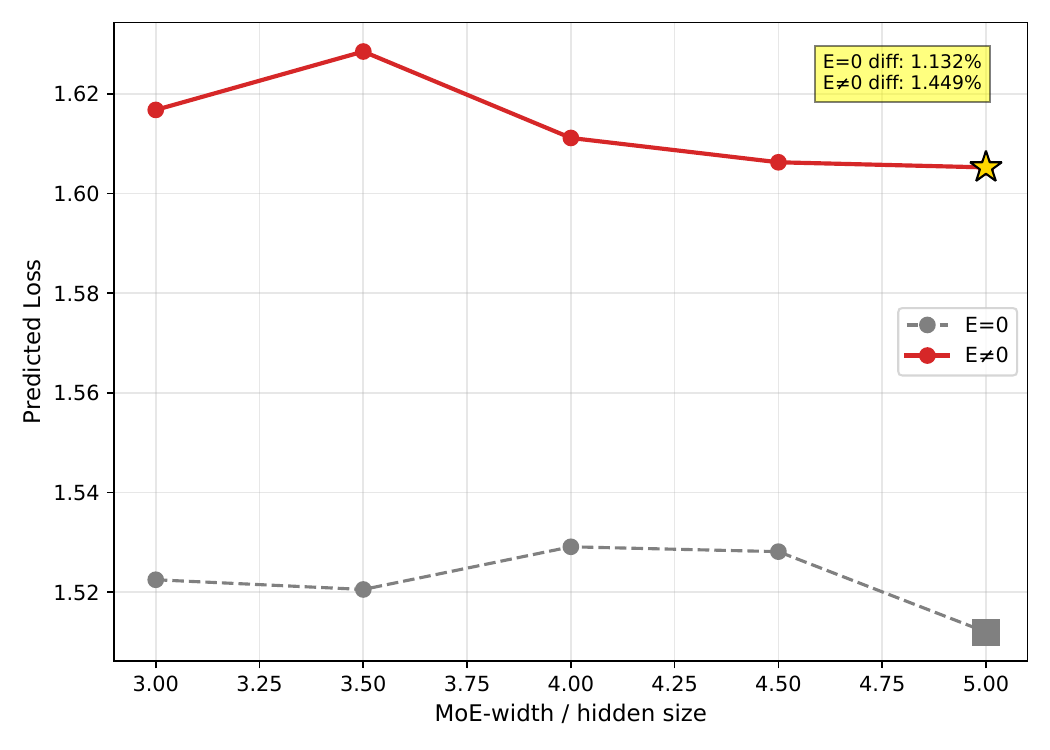}
        \\(e) $C = 3\times10^{22}$
    \end{minipage}
    \hfill
    \begin{minipage}[b]{0.32\textwidth}
        \centering
        \includegraphics[width=\textwidth]{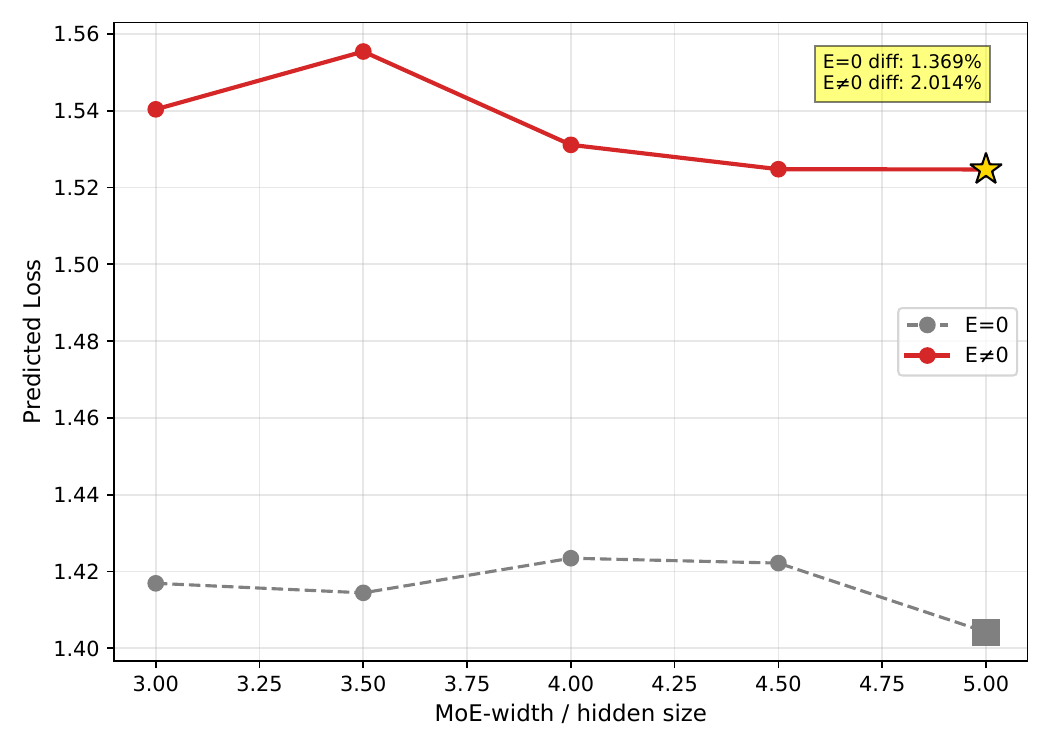}
        \\(f) $C = 10^{23}$
    \end{minipage}
    \caption{Predicted loss vs.\ MoE FFN width ratio at extrapolated compute scales, comparing E=0 (gray dashed) and E$\neq$0 (red solid) model predictions. $C = 3\times10^{20}$ (panel a) overlaps with the experimental range to bridge the two regimes.}
    \label{fig:moe_width_by_C_ext}
\end{figure}

\paragraph{Extrapolation to larger scales.}
Figure~\ref{fig:moe_width_by_C_ext} extends the analysis to extrapolated compute scales ($3\times10^{20}$ to $10^{23}$), where $C = 3\times10^{20}$ is shared with Figure~\ref{fig:moe_width_by_C_exp} to provide continuity. At larger extrapolated scales, the two models begin to diverge in their predictions of the optimal $(K+1)\cdot d_m/d$: the E=0 model tends to favor lower ratios, while the E$\neq$0 model shows a somewhat different ranking. However, in both cases, the absolute loss differences among $(K+1)\cdot d_m/d$ groups remain small, suggesting that this ratio has limited influence on overall model performance compared to the macroscopic parameters $(M, N_a, N)$ and the hidden dimension $d$.

\paragraph{Summary.}
This case study validates the generality of our dimension decomposition framework. By simply replacing the hidden dimension $d$ with the MoE FFN width ratio $(K+1)\cdot d_m/d$ as the single free structural variable, while keeping the macroscopic parameters $(M, N_a, N)$ locked at their optimal values, we can systematically study the scaling properties of this new ratio with minimal additional experimental effort. The results indicate that $(K+1)\cdot d_m/d$ has a comparatively weak influence on model performance (less than $0.5\%$ loss variation across the tested range), suggesting that practitioners have considerable flexibility in choosing this ratio based on hardware or engineering considerations without incurring significant performance penalties.

\section{Limitations}
\label{app:limitations}
We acknowledge several limitations of this work:
\begin{enumerate}
    \item \textbf{Experimental scale under resource constraints.} While our dimension decomposition strategy reduces the search space to a feasible range, the total number of experiments ($\sim$670 models) may not fully guarantee the robustness of all fitted coefficients, particularly for the power-law relationships that govern extrapolation to larger scales. The framework itself is designed to be strengthened with additional experiments: denser sampling within the same pipeline would directly improve coefficient reliability without requiring methodological changes.

    \item \textbf{$(C, M, D)$ scaling laws fitted with heuristic architectures.} The classical $(C, M, D)$ relationships (Eq.~\eqref{eq:optimal_cmd_scaling}) were fitted using models with heuristic architectural configurations, whereas the optimal architectures discovered in Sections~\ref{sec:mnna}--\ref{sec:mnnad} may deviate from these heuristics. In principle, one could iterate: fit $(C, M, D)$ $\to$ find optimal $(N_a, N, d)$ $\to$ re-fit $(C, M, D)$ with the new architectures $\to$ repeat until convergence. We did not perform this iteration due to resource constraints, though preliminary checks suggest the perturbation to the $(C, M, D)$ trajectory is small.

    \item \textbf{Fixed sparsity configuration.} All experiments use $N_e = 288$ and $K = 8$. While this fixes the sparsity level rather than the full MoE structure (expert granularity and layer allocation remain free), the derived scaling laws may not directly transfer to substantially different $(N_e, K)$ configurations. The framework can be reapplied to alternative sparsity settings, but doing so requires repeating the experimental pipeline.

    \item \textbf{Evaluation limited to pre-training loss.} Our scaling laws are fitted and validated exclusively on pre-training cross-entropy loss. Prior work~\citep{jelassi2024mixture,abnar2025parameters} has shown that MoE scaling behaviors can differ significantly across downstream tasks, particularly between memorization-intensive and reasoning-intensive tasks. The extent to which our architectural recommendations transfer to downstream performance remains an open question.
\end{enumerate}

\section{Extended related work}
\label{app:related_work_full}
The following provides a more detailed discussion of related work, supplementing the condensed treatment in the main text.

\subsection{Scaling laws for language models}

Scaling laws provide the fundamental theoretical framework for predicting language model performance relative to computational resources. \cite{kaplan2020scaling} formalized this paradigm by demonstrating robust power-law relationships between cross-entropy loss, parameter count, and total compute budget. Shifting the focus to training efficiency, \cite{hoffmann2022training} established the ``compute-optimal'' criterion (the Chinchilla laws), proving that model parameters and training tokens must scale equiproportionally to mitigate diminishing returns under fixed compute budgets.
Recent advancements extend these foundational frameworks to more granular operational dimensions. \cite{bi2024deepseek} demonstrated that high-quality datasets shift the scaling frontier, yielding higher training efficiency. Addressing practical deployment constraints, \cite{sardana2023beyond} integrated inference costs into the optimal scaling formulations, while \cite{gadre2024language} delineated the predictive boundaries of severe over-training. Furthermore, contemporary studies have successfully generalized scaling laws to global hyperparameter optimization~\cite{li2025predictable}, and validated their robust predictive power across complex downstream tasks, multimodal architectures, and emergent capabilities~\cite{chen2024scaling, ruan2024observational, roberts2025compute, yang2025scaling}.

\subsection{Scaling laws for mixture-of-experts (MoE)}
The Mixture-of-Experts (MoE) architecture~\citep{shazeer2017outrageously, lepikhin2020gshard,fedus2022switch,muennighoff2024olmoe} successfully decouples the total parameter count from the forward computation cost (FLOPs per token) via sparse routing mechanisms. This decoupling of active and total parameters renders traditional dense scaling laws inapplicable, thereby spurring extensive research into scaling behaviors specific to MoE architectures.
To investigate MoE scaling, researchers have analyzed the combined impact of parameter scale, expert count, and structural configuration on model performance.~\cite{clark2022unified} pioneered the exploration of joint scaling between model size and expert count on fixed datasets, observing diminishing returns in routing efficiency as models scale up. However, \cite{ludziejewski2024scaling} revised this view by introducing variable dataset sizes and the concept of "expert granularity," demonstrating that MoE models can consistently outperform dense counterparts given adaptive configurations and appropriate computational budgets. Building on this,~\cite{wang2024scaling} and~\cite{ludziejewski2025joint} refined the theoretical framework, deriving transferable and joint scaling laws applicable to both dense and MoE architectures. Concurrently, the specific formulation of expert configuration has proven critical for scaling efficiency: DeepSeekMoE~\citep{dai2024deepseekmoe} revealed the significant advantages of mixing fine-grained and shared experts; conversely,~\cite{zoph2022st} identified a sharp decline in sparsity benefits when the expert count exceeds 256, highlighting the scaling limits of highly sparse models. \cite{tian2025towards} introduced "Efficiency Leverage" (EL) to quantify MoE's computational advantage, proposing a unified scaling law based on expert activation ratio and expert granularity. Furthermore, MoE scaling benefits exhibit pronounced heterogeneity across downstream tasks. Studies by~\cite{jelassi2024mixture} and~\cite{abnar2025parameters} indicate that increasing expert count substantially accelerates scaling for memorization-intensive tasks, whereas improvements tend to saturate on complex reasoning tasks.
Despite progress in MoE scaling, most studies fix non-MoE architectural variables (e.g., attention, dense layers), confining scaling analysis to internal MoE mechanisms. This limits comprehensive, macroscopic model-level exploration. A critical gap thus remains in establishing a generalized MoE scaling framework that holistically encompasses all architectural shape configurations.

\subsection{Parameter allocation in scaling laws}
Early studies on classic scaling laws established a consensus: model performance is relatively insensitive to specific shape configurations (e.g., aspect ratio, attention heads) provided they remain within reasonable bounds. Consequently, dense model architectures predominantly rely on heuristics or legacy configurations to circumvent performance degradation. For instance, ~\cite{hoffmann2022training} and other existing works merely list their structural choices in the appendices, without providing further explanation or discussion. While \cite{kaplan2020scaling} explicitly investigated these configurations, they concluded their impact is minimal. Yet, their empirical data reveals up to an 8\% performance variance across differing shapes under identical compute budgets, indicating that parameter allocation is decidedly non-trivial.
For standard dense models, this lack of systematic guidance remains tolerable, as practitioners can safely fall back on established heuristics. However, the recent paradigm shift toward MoE architectures breaks this empirical reliance. By introducing additional structural degrees of freedom, MoE exponentially expands the hyperparameter search space. To manage this escalating complexity and mitigate experimental overhead, current MoE scaling studies typically isolate specific components, freezing attention and dense layer dimensions to focus exclusively on scaling expert count or routing sparsity~\citep{clark2022unified,ludziejewski2024scaling,ludziejewski2025joint,wang2024scaling,abnar2025parameters}. Yet, this isolated approach fundamentally neglects the intricate computational trade-offs and resource competition among MoE modules, dense layers, and attention mechanisms. Consequently, the local optima derived from these prior works serve merely as directional heuristics, lacking the rigorous reliability required for globally optimal architecture design.

In summary, existing approaches to MoE scaling laws fall into two broad categories: (1)~directly incorporating MoE-specific variables into the scaling law formulation, which increases the number of fitted coefficients and demands a proportional increase in experimental scale to maintain fitting reliability; and (2)~fixing all non-MoE architectural factors and studying MoE-internal properties in isolation, which implicitly assumes that global architectural parameters do not influence local MoE scaling behavior. Both strategies leave a critical gap: the interaction between MoE-specific and traditional architectural dimensions under a holistic optimization objective remains unexplored. Our work addresses precisely this gap.

\section{Experimental setup}
\label{app:exp_setup}

\subsection{Common experimental infrastructure}
\label{app:exp_infra}

In our early independent reruns of identical configurations, the absolute loss fluctuation across runs was approximately 0.005.

We detail the common experimental setup and infrastructure adopted across all MoE scaling experiments, ensuring reproducibility and comparability.

\textbf{Backbone models}
A unified set of architectural characteristics is maintained, largely consistent with~\cite{huo2025dots} and~\cite{liu2024deepseek}:  each MoE layer includes 1 shared expert and 288 routed experts, with 8 experts activated per token. We employ Rotary Position Embedding (RoPE) \citep{su2024roformer} and support a fixed context length of 8192 tokens. For the attention mechanism, Grouped Query Attention (GQA)~\citep{ainslie2023gqa} is integrated, setting the ratio of query heads to key-value heads ($N_q/N_{kv}$) to 2. The intermediate dimension of the feed-forward network is consistently set to 3 times the hidden dimension, and the SwiGLU activation function\citep{shazeer2020glu} is employed. Details on hyperparameters varying with model scale are provided in Appendix~\ref{app:hyper_model_config}.

\textbf{Datasets and training}

For pretraining, all models utilize a high-quality curated text corpus from online sources, encompassing English and multiple other languages across web text, code, mathematics, books, question-answering, and reasoning domains, all subjected to rigorous filtering. Tokenization is performed using a tokenizer with a fixed vocabulary size of 152064. Our training regimen employs the \texttt{AdamW} optimizer~\citep{loshchilov2017decoupled}. The learning rate schedule follows a \texttt{WSD (Warmup-Stable-Decay)} strategy~\citep{hu2024minicpm}. Following~\cite{bi2024deepseek} and~\cite{team2025every}, we extensively explored scaling laws for training hyperparameters, such as learning rate and global batch size, applying these findings directly to our experiments.  As this is not the primary focus of this paper, detailed experiments and discussions are provided in Appendix~\ref{app:hyper_lr_gbs}. All models are trained using the \texttt{Megatron-LM framework}~\citep{megatron-lm} for distributed training. Further training details for each model scale are available in Appendix~\ref{app:hyper_training_details}.

\subsection{Compute-optimal $(C, M, D)$ combinations}
\label{app:cmd_table}

\begin{table}[htbp]
\centering
\caption{Configurations with nearly identical per-token compute at the $C=10^{20}$ experimental scale. Similar $M$ does not imply similar parameter storage.}
\label{tab:resource_example}
\begingroup{}
\begin{tabular}{rrrr}
\toprule
$M$ (GFLOPs) & Attention share & $N$ (B) & Loss \\
\midrule
3.02 & 1.9\% & 5.28 & 2.125 \\
3.03 & 0.5\% & 9.90 & 2.102 \\
3.01 & 0.2\% & 10.02 & 2.226 \\
\bottomrule
\end{tabular}
\endgroup{}
\end{table}

\begin{table}[htbp]
    \centering
    \caption{Six compute-optimal $(C, M, D)$ combinations}
    \label{tab:cmd_combinations}
    \begin{tabular}{ccc}
        \toprule
        $C$ (EFLOPs) & $M$ (GFLOPs) & $D$ (Billion Tokens) \\
        \midrule
        1 & $0.2672$ & $3.7420$ \\
        3 & $0.4856$ & $6.1775$ \\
        10 & $0.9345$ & $10.7004$ \\
        30 & $1.6983$ & $17.6649$ \\
        100 & $3.2681$ & $30.5985$ \\
        300 & $5.9390$ & $50.5138$ \\
        \bottomrule
    \end{tabular}
\end{table}

\subsection{Common hyperparameters varying with model scale}
\label{app:hyper_model_config}
\label{app:hyper_training_details}
We provide a comprehensive overview of the specific architectural and training hyperparameters that were adjusted across different model scales (compute budgets). Table~\ref{tab:hyper_scales} summarizes these configurations, detailing how key parameters were set for each defined computational budget $C$.

\begin{table}[htbp]
\centering
\caption{Common hyperparameters and architectural configurations across different compute scales. \textbf{Note:} The actual $M$ values in our experiments might slightly deviate from theoretically derived optimal values, as model dimensions were constrained to be multiples of $2^{3}$ to $2^{5}$ for hardware efficiency. We ensured this deviation remained within 5\% to maintain experimental reliability.}
\label{tab:hyper_scales}
\renewcommand{\arraystretch}{1.2}
\begin{tabular}{l c c c c c c}
\toprule
\textbf{Parameter} & \multicolumn{6}{c}{\textbf{Different Compute Scales}} \\
\midrule
\multicolumn{7}{c}{\textit{Overall Resource Metrics}} \\
$C$ & $1e18$ & $3e18$ & $1e19$ & $3e19$ & $1e20$ & $3e20$ \\
$M$ (GFLOPs) & 0.2672 & 0.4856 & 0.9345 & 1.6983 & 3.2681 & 5.9390 \\
\midrule
\multicolumn{7}{c}{\textit{Structural Parameters}} \\
$d_{qkv}$ & 64 & 64 & 64 & 128 & 128 & 128 \\
$N_q$ & 4 & 8 & 8 & 8 & 8 & 16 \\
$N_{kv}$ & 2 & 4 & 4 & 4 & 4 & 8 \\
$S$ & \multicolumn{6}{c}{8192} \\
$K$ & \multicolumn{6}{c}{8} \\
$N_e$ & \multicolumn{6}{c}{288} \\
$L, L_d, L_m, d, d_d, d_m$ & \multicolumn{6}{c}{Not fixed; varies within each compute scale} \\
\midrule
\multicolumn{7}{c}{\textit{Training Hyperparameters}} \\
Learning Rate & 1.20e-3 & 1.00e-3 & 8.28e-4 & 6.95e-4 & 5.74e-4 & 4.82e-4 \\
Batch Size & 64 & 80 & 128 & 160 & 208 & 288 \\
Optimizer & \multicolumn{6}{c}{\textit{AdamW}} \\
$\beta_1$ & \multicolumn{6}{c}{0.9} \\
$\beta_2$ & \multicolumn{6}{c}{0.95} \\
Weight Decay & \multicolumn{6}{c}{0.1} \\
Clip Gradient & \multicolumn{6}{c}{1.0} \\
Vocabulary Size & \multicolumn{6}{c}{152064} \\
\bottomrule
\end{tabular}
\end{table}

\subsection{Scaling laws for training hyperparameters}
\label{app:hyper_lr_gbs}
\begin{figure}[!h]
    \centering
    \begin{minipage}[b]{0.48\textwidth}
        \centering
        \includegraphics[width=\textwidth]{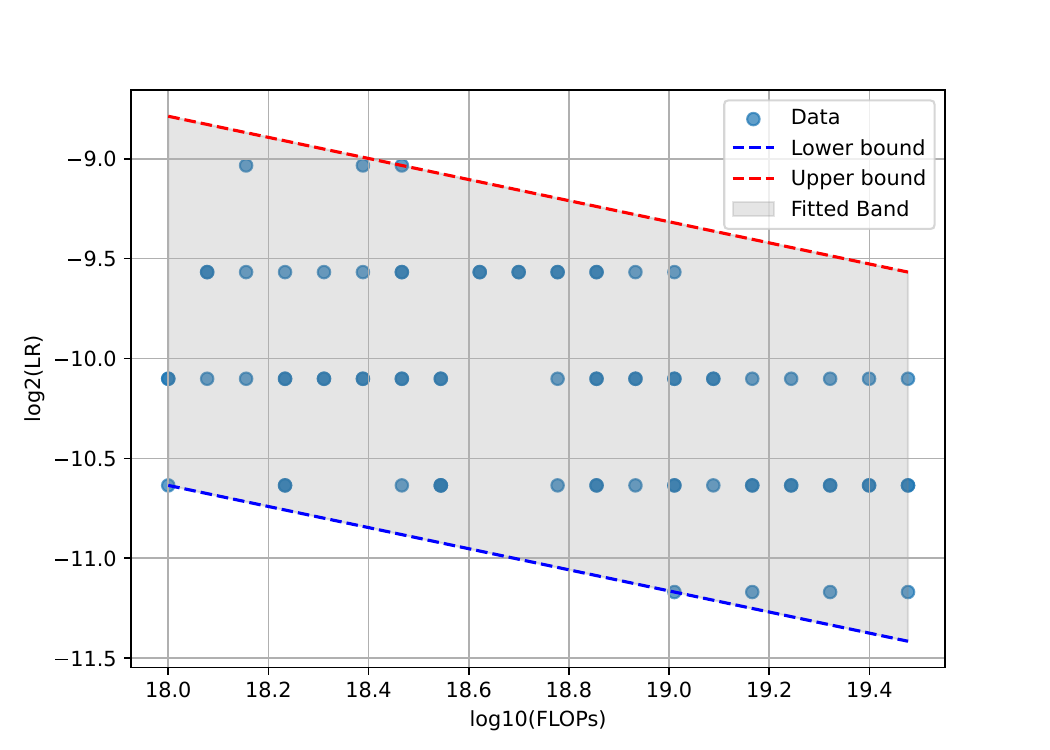}
        \\(a) Scaling of learning rate
        \label{fig:hyper_lr}
    \end{minipage}
    \hfill
    \begin{minipage}[b]{0.48\textwidth}
        \centering
        \includegraphics[width=\textwidth]{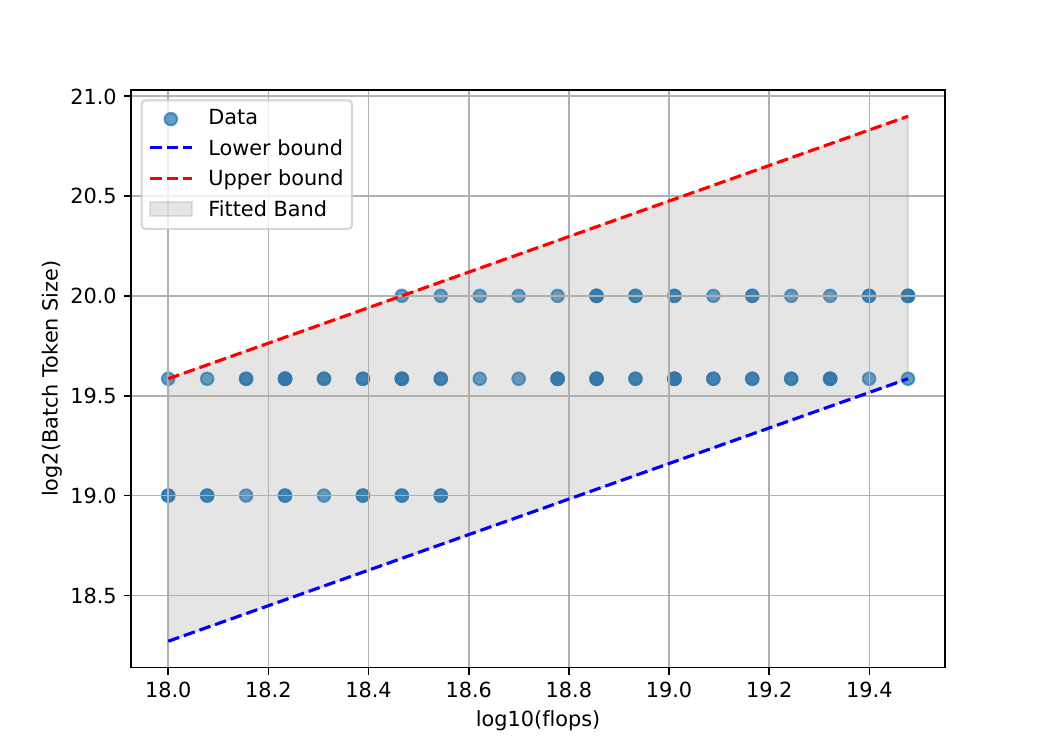}
        \\(b) Scaling of batch size
        \label{fig:hyper_bs}
    \end{minipage}
    \caption{\textbf{Scaling laws for learning rate and batch size.} This figure illustrates the scaling behavior of optimal hyperparameters with respect to computational budget (FLOPs). Both plots share a common x-axis, $\log_{10}(\text{FLOPs})$. The blue dots represent experimental optimal settings, while the dashed blue and red lines define the lower and upper bounds of the robust linear fitted band (shaded gray area), respectively.
        \textbf{(a) Scaling of learning rate:} Y-axis shows $\log_{10}(\text{learning rate})$.
        \textbf{(b) Scaling of batch size:} Y-axis shows $\log_{2}(\text{Batch Token Size})$.
        These derived scaling laws inform the hyperparameter choices in our experiments.}
    
    \label{fig:hyper_lr_gbs}
\end{figure}

Following~\cite{bi2024deepseek} and~\cite{team2025every}, we provide a detailed account of our methodology for exploring and applying scaling laws for training hyperparameters, specifically the learning rate (LR) and global batch size (GBS). While not the primary focus of this paper, these investigations were crucial for optimizing the training efficiency and performance of our models.

For each distinct model configuration (denoted as $C$), we conducted a series of experiments to identify optimal hyperparameter settings. Our definition of "optimal" for a given $C$ is inspired by the practices in~\cite{bi2024deepseek} and~\cite{team2025every}. Specifically, we consider an LR-GBS combination to be optimal if its achieved training loss falls within a 0.2\% tolerance band of the minimum observed loss for that configuration. 

As illustrated in Figure~\ref{fig:hyper_lr_gbs}, after collecting the set of optimal LR-GBS pairs for various model configurations $C$, we applied a robust linear band fitting procedure. This process involves fitting two parallel linear lines, an upper bound and a lower bound, to the collected optimal points, ensuring both lines share the same slope. The central line, equidistant from these upper and lower bounds, is then determined as the mean fit line. This approach allows us to establish a robust scaling relationship between model configurations and their optimal learning rate and global batch size settings, which were subsequently applied to the main experiments presented in this paper.

\subsection{Sampling strategy for $(M/N_a, N/N_a)$ exploration}
\label{app:sampling_strategy}

Our experimental scope encompasses six distinct scales for the computational budget $M$, as detailed in Table~\ref{tab:cmd_combinations}. For each chosen $M$ value, we sample six points along both the $M/N_a$ ratio axis and the $N/N_a$ ratio axis. This Cartesian product approach yields a total of $6 \times 6 \times 6 = 216$ unique experimental configurations.

The specific sampling values for these normalized ratios are as follows:
\begin{itemize}
    \item $M/N_a$: $\{7, 8, 9, 11, 14, 17\}$. This range is selected because, as derived in Eq.~\eqref{eq:ratio_expressions}, $M/N_a$ is inherently always greater than 6. While theoretically its upper bound can be very large, excessively high $M/N_a$ values would imply a very small number of active parameters ($N_a$) for a given computational budget ($M$), leading to models with typically poor performance and thus not within our primary region of interest for efficient exploration.
    \item $N/N_a$: $\{12, 16, 20, 22, 26, 30\}$. The lower bound for $N/N_a$ is always greater than 1 (as $N \ge N_a$). Similarly, very small $N/N_a$ values also tend to correlate with suboptimal model performance. The upper limit of $N/N_a$ is approximately $\frac{N_e+1}{K+1}$, which, for our specific configurations, is around 32. Therefore, this chosen range effectively covers the practically relevant and high-performing region of the parameter expansion ratio without exploring values that are either theoretically unachievable or practically uninteresting.
\end{itemize}
These values are chosen to cover a meaningful range within the feasible region identified. For a comprehensive list of all model configurations tested, please refer to the tables below.

\subsection{Specific model configurations in each experiment}
\label{app:config_of_all}
We detail all the specific configurations dynamically adjusted for specific models across different compute scales in each experiment.
For the experiments presented in Section~\ref{subsec:efficient_search_strategy}, corresponding configurations can be found in Tables~\ref{tab:setup_proxy1},~\ref{tab:setup_proxy2},~\ref{tab:setup_proxy3}, and~\ref{tab:setup_proxy4}. The experimental setups for Section~\ref{sec:mnna_exp_confg} are documented in Tables~\ref{tab:setup_mnna1} through~\ref{tab:setup_mnna6}. Furthermore, the configurations utilized for Section~\ref{sec:mnna_scaling_laws_result} are summarized in Table~\ref{tab:setup_mnna_v2}, and those for Section~\ref{sec:mnnad} are delineated in Tables~\ref{tab:setup_mnnad1} through~\ref{tab:setup_mnnad6}.

\begin{table}[p]
\centering
\begin{minipage}[t]{0.49\textwidth}
\captionsetup{font=small,justification=raggedright,singlelinecheck=false}
\centering
\caption{Specific architectural parameters ($d, L_d, d_d, L_m, d_m, L$) in Sec~\ref{subsec:efficient_search_strategy}.}
\label{tab:setup_proxy1}
\small
\setlength{\tabcolsep}{4pt}
\renewcommand{\arraystretch}{1.2}
\begin{tabular}{l c c c c c c} %
\toprule
\textbf{$C$} & \textbf{$d$} & \textbf{$L_d$} & \textbf{$d_d$} & \textbf{$L_m$} & \textbf{$d_m$} & \textbf{$L$} \\
\midrule
\multirow{36}{*}{$1e18$} & 1208 & 2 & 3624 & 1 & 264 & 3 \\
& 1224 & 2 & 3672 & 1 & 248 & 3 \\
& 1240 & 2 & 3720 & 1 & 232 & 3 \\
& 1256 & 2 & 3768 & 1 & 216 & 3 \\
& 1576 & 1 & 4728 & 2 & 104 & 3 \\
& 1592 & 1 & 4776 & 2 & 104 & 3 \\
& 1608 & 1 & 4824 & 2 & 104 & 3 \\
& 1640 & 1 & 4920 & 2 & 96 & 3 \\
& 1656 & 1 & 4968 & 2 & 96 & 3 \\
& 1672 & 1 & 5016 & 2 & 96 & 3 \\
& 1688 & 1 & 5064 & 2 & 96 & 3 \\
& 1704 & 1 & 5112 & 2 & 88 & 3 \\
& 1720 & 1 & 5160 & 2 & 88 & 3 \\
& 1736 & 1 & 5208 & 2 & 88 & 3 \\
& 1752 & 1 & 5256 & 2 & 88 & 3 \\
& 968 & 2 & 2904 & 1 & 736 & 3 \\
& 984 & 2 & 2952 & 1 & 672 & 3 \\
& 1000 & 2 & 3000 & 1 & 624 & 3 \\
& 1016 & 2 & 3048 & 1 & 576 & 3 \\
& 1304 & 1 & 3912 & 2 & 272 & 3 \\
& 1320 & 1 & 3960 & 2 & 264 & 3 \\
& 1336 & 1 & 4008 & 2 & 256 & 3 \\
& 1352 & 1 & 4056 & 2 & 248 & 3 \\
& 1368 & 1 & 4104 & 2 & 240 & 3 \\
& 1384 & 1 & 4152 & 2 & 240 & 3 \\
& 1400 & 1 & 4200 & 2 & 232 & 3 \\
& 1416 & 1 & 4248 & 2 & 224 & 3 \\
& 1432 & 1 & 4296 & 2 & 224 & 3 \\
& 1448 & 1 & 4344 & 2 & 216 & 3 \\
& 1464 & 1 & 4392 & 2 & 216 & 3 \\
& 1480 & 1 & 4440 & 2 & 208 & 3 \\
& 680 & 2 & 2040 & 1 & 1504 & 3 \\
& 696 & 2 & 2088 & 1 & 1336 & 3 \\
& 888 & 1 & 2664 & 2 & 616 & 3 \\
& 904 & 1 & 2712 & 2 & 584 & 3 \\
& 920 & 1 & 2760 & 2 & 568 & 3 \\
\bottomrule
\end{tabular}
\end{minipage}
\hfill
\begin{minipage}[t]{0.49\textwidth}
\captionsetup{font=small,justification=raggedright,singlelinecheck=false}
\centering
\caption{Specific architectural parameters ($d, L_d, d_d, L_m, d_m, L$) in Sec~\ref{subsec:efficient_search_strategy}.}
\label{tab:setup_proxy2}
\small
\setlength{\tabcolsep}{4pt}
\renewcommand{\arraystretch}{1.2}
\begin{tabular}{l c c c c c c} %
\toprule
\textbf{$C$} & \textbf{$d$} & \textbf{$L_d$} & \textbf{$d_d$} & \textbf{$L_m$} & \textbf{$d_m$} & \textbf{$L$} \\
\midrule
\multirow{36}{*}{$1e18$} & 936 & 1 & 2808 & 2 & 544 & 3 \\
& 952 & 1 & 2856 & 2 & 528 & 3 \\
& 968 & 1 & 2904 & 2 & 512 & 3 \\
& 984 & 1 & 2952 & 2 & 496 & 3 \\
& 1000 & 1 & 3000 & 2 & 480 & 3 \\
& 1016 & 1 & 3048 & 2 & 464 & 3 \\
& 1032 & 1 & 3096 & 2 & 448 & 3 \\
& 1048 & 1 & 3144 & 2 & 440 & 3 \\
& 1064 & 1 & 3192 & 2 & 424 & 3 \\
& 1080 & 1 & 3240 & 2 & 416 & 3 \\
& 1096 & 1 & 3288 & 2 & 408 & 3 \\
& 664 & 5 & 1992 & 2 & 168 & 7 \\
& 712 & 4 & 2136 & 3 & 120 & 7 \\
& 736 & 4 & 2208 & 3 & 104 & 7 \\
& 808 & 3 & 2424 & 4 & 80 & 7 \\
& 832 & 3 & 2496 & 4 & 72 & 7 \\
& 856 & 3 & 2568 & 4 & 64 & 7 \\
& 928 & 2 & 2784 & 5 & 56 & 7 \\
& 976 & 2 & 2928 & 5 & 48 & 7 \\
& 1000 & 2 & 3000 & 5 & 48 & 7 \\
& 1024 & 2 & 3072 & 5 & 48 & 7 \\
& 1216 & 1 & 3648 & 6 & 32 & 7 \\
& 1240 & 1 & 3720 & 6 & 32 & 7 \\
& 1264 & 1 & 3792 & 6 & 32 & 7 \\
& 1288 & 1 & 3864 & 6 & 32 & 7 \\
& 1312 & 1 & 3936 & 6 & 32 & 7 \\
& 1336 & 1 & 4008 & 6 & 32 & 7 \\
& 520 & 5 & 1560 & 2 & 528 & 7 \\
& 592 & 4 & 1776 & 3 & 280 & 7 \\
& 640 & 3 & 1920 & 4 & 216 & 7 \\
& 664 & 3 & 1992 & 4 & 192 & 7 \\
& 688 & 3 & 2064 & 4 & 176 & 7 \\
& 760 & 2 & 2280 & 5 & 144 & 7 \\
& 784 & 2 & 2352 & 5 & 136 & 7 \\
& 808 & 2 & 2424 & 5 & 128 & 7 \\
& 832 & 2 & 2496 & 5 & 120 & 7 \\
\bottomrule
\end{tabular}
\end{minipage}
\end{table}
\clearpage

\begin{table}[p]
\centering
\begin{minipage}[t]{0.49\textwidth}
\captionsetup{font=small,justification=raggedright,singlelinecheck=false}
\centering
\caption{Specific architectural parameters ($d, L_d, d_d, L_m, d_m, L$) in Sec~\ref{subsec:efficient_search_strategy}.}
\label{tab:setup_proxy3}
\small
\setlength{\tabcolsep}{4pt}
\renewcommand{\arraystretch}{1.2}
\begin{tabular}{l c c c c c c} %
\toprule
\textbf{$C$} & \textbf{$d$} & \textbf{$L_d$} & \textbf{$d_d$} & \textbf{$L_m$} & \textbf{$d_m$} & \textbf{$L$} \\
\midrule
\multirow{36}{*}{$1e18$} & 856 & 2 & 2568 & 5 & 112 & 7 \\
& 952 & 1 & 2856 & 6 & 96 & 7 \\
& 1000 & 1 & 3000 & 6 & 88 & 7 \\
& 1024 & 1 & 3072 & 6 & 88 & 7 \\
& 1048 & 1 & 3144 & 6 & 80 & 7 \\
& 1072 & 1 & 3216 & 6 & 80 & 7 \\
& 1096 & 1 & 3288 & 6 & 80 & 7 \\
& 1120 & 1 & 3360 & 6 & 72 & 7 \\
& 400 & 4 & 1200 & 3 & 632 & 7 \\
& 448 & 3 & 1344 & 4 & 440 & 7 \\
& 472 & 2 & 1416 & 4 & 384 & 6 \\
& 496 & 2 & 1488 & 5 & 344 & 7 \\
& 520 & 2 & 1560 & 5 & 312 & 7 \\
& 544 & 2 & 1632 & 5 & 280 & 7 \\
& 568 & 2 & 1704 & 5 & 264 & 7 \\
& 616 & 1 & 1848 & 6 & 224 & 7 \\
& 640 & 1 & 1920 & 6 & 216 & 7 \\
& 664 & 1 & 1992 & 6 & 200 & 7 \\
& 688 & 1 & 2064 & 6 & 192 & 7 \\
& 712 & 1 & 2136 & 6 & 184 & 7 \\
& 736 & 1 & 2208 & 6 & 176 & 7 \\
& 760 & 1 & 2280 & 6 & 168 & 7 \\
& 784 & 1 & 2352 & 6 & 160 & 7 \\
& 808 & 1 & 2424 & 6 & 152 & 7 \\
& 880 & 1 & 2640 & 6 & 136 & 7 \\
& 104 & 5 & 312 & 2 & 5112 & 7 \\
& 112 & 4 & 336 & 3 & 3072 & 7 \\
& 120 & 3 & 360 & 4 & 2248 & 7 \\
& 128 & 2 & 384 & 4 & 1800 & 6 \\
& 136 & 2 & 408 & 5 & 1512 & 7 \\
& 144 & 1 & 432 & 5 & 1320 & 6 \\
& 152 & 1 & 456 & 6 & 1168 & 7 \\
& 160 & 1 & 480 & 6 & 1064 & 7 \\
& 168 & 1 & 504 & 6 & 968 & 7 \\
& 176 & 1 & 528 & 6 & 896 & 7 \\
& 352 & 9 & 1056 & 3 & 120 & 12 \\
\bottomrule
\end{tabular}
\end{minipage}
\hfill
\begin{minipage}[t]{0.49\textwidth}
\captionsetup{font=small,justification=raggedright,singlelinecheck=false}
\centering
\caption{Specific architectural parameters ($d, L_d, d_d, L_m, d_m, L$) in Sec~\ref{subsec:efficient_search_strategy}.}
\label{tab:setup_proxy4}
\small
\setlength{\tabcolsep}{4pt}
\renewcommand{\arraystretch}{1.2}
\begin{tabular}{l c c c c c c} %
\toprule
\textbf{$C$} & \textbf{$d$} & \textbf{$L_d$} & \textbf{$d_d$} & \textbf{$L_m$} & \textbf{$d_m$} & \textbf{$L$} \\
\midrule
\multirow{36}{*}{$1e18$} & 376 & 8 & 1128 & 5 & 80 & 13 \\
& 400 & 7 & 1200 & 6 & 64 & 13 \\
& 448 & 5 & 1344 & 7 & 48 & 12 \\
& 472 & 4 & 1416 & 8 & 40 & 12 \\
& 496 & 4 & 1488 & 8 & 32 & 12 \\
& 520 & 3 & 1560 & 9 & 32 & 12 \\
& 544 & 3 & 1632 & 9 & 32 & 12 \\
& 592 & 2 & 1776 & 10 & 24 & 12 \\
& 616 & 2 & 1848 & 10 & 24 & 12 \\
& 640 & 2 & 1920 & 10 & 24 & 12 \\
& 760 & 1 & 2280 & 11 & 16 & 12 \\
& 784 & 1 & 2352 & 11 & 16 & 12 \\
& 808 & 1 & 2424 & 11 & 16 & 12 \\
& 832 & 1 & 2496 & 11 & 16 & 12 \\
& 856 & 1 & 2568 & 11 & 16 & 12 \\
& 304 & 7 & 912 & 5 & 200 & 12 \\
& 328 & 6 & 984 & 6 & 152 & 12 \\
& 352 & 5 & 1056 & 7 & 120 & 12 \\
& 376 & 4 & 1128 & 8 & 104 & 12 \\
& 400 & 4 & 1200 & 9 & 88 & 13 \\
& 424 & 3 & 1272 & 9 & 80 & 12 \\
& 448 & 3 & 1344 & 10 & 72 & 13 \\
& 472 & 2 & 1416 & 10 & 64 & 12 \\
& 496 & 2 & 1488 & 10 & 56 & 12 \\
& 544 & 1 & 1632 & 11 & 48 & 12 \\
& 568 & 1 & 1704 & 11 & 48 & 12 \\
& 592 & 1 & 1776 & 11 & 48 & 12 \\
& 640 & 1 & 1920 & 11 & 40 & 12 \\
& 664 & 1 & 1992 & 11 & 40 & 12 \\
& 72 & 10 & 216 & 2 & 4416 & 12 \\
& 80 & 7 & 240 & 5 & 1440 & 12 \\
& 88 & 5 & 264 & 7 & 904 & 12 \\
& 96 & 3 & 288 & 9 & 672 & 12 \\
& 104 & 2 & 312 & 10 & 552 & 12 \\
& 112 & 1 & 336 & 11 & 464 & 12 \\
& 120 & 1 & 360 & 12 & 408 & 13 \\
\bottomrule
\end{tabular}
\end{minipage}
\end{table}
\clearpage

\begin{table}[p]
\centering
\begin{minipage}[t]{0.49\textwidth}
\captionsetup{font=small,justification=raggedright,singlelinecheck=false}
\centering
\caption{Specific architectural parameters ($d, L_d, d_d, L_m, d_m, L$) in Sec~\ref{sec:mnna_exp_confg}.}
\label{tab:setup_mnna1}
\small
\setlength{\tabcolsep}{4pt}
\renewcommand{\arraystretch}{1.2}
\begin{tabular}{l c c c c c c} %
\toprule
\textbf{$C$} & \textbf{$d$} & \textbf{$L_d$} & \textbf{$d_d$} & \textbf{$L_m$} & \textbf{$d_m$} & \textbf{$L$} \\
\midrule
\multirow{36}{*}{$1e18$} & 1496 & 1 & 4488 & 2 & 168 & 3 \\
& 1352 & 1 & 4056 & 2 & 248 & 3 \\
& 1208 & 1 & 3624 & 2 & 336 & 3 \\
& 1096 & 1 & 3288 & 2 & 408 & 3 \\
& 816 & 1 & 2448 & 2 & 664 & 3 \\
& 432 & 1 & 1296 & 2 & 1456 & 3 \\
& 1192 & 1 & 3576 & 4 & 88 & 5 \\
& 1048 & 1 & 3144 & 4 & 144 & 5 \\
& 936 & 1 & 2808 & 4 & 192 & 5 \\
& 848 & 1 & 2544 & 4 & 240 & 5 \\
& 616 & 1 & 1848 & 4 & 392 & 5 \\
& 304 & 1 & 912 & 4 & 928 & 5 \\
& 896 & 2 & 2688 & 5 & 88 & 7 \\
& 832 & 2 & 2496 & 5 & 120 & 7 \\
& 720 & 2 & 2160 & 5 & 168 & 7 \\
& 712 & 1 & 2136 & 6 & 176 & 7 \\
& 576 & 1 & 1728 & 6 & 248 & 7 \\
& 248 & 1 & 744 & 6 & 704 & 7 \\
& 720 & 2 & 2160 & 7 & 56 & 9 \\
& 640 & 2 & 1920 & 7 & 88 & 9 \\
& 536 & 2 & 1608 & 7 & 136 & 9 \\
& 480 & 2 & 1440 & 7 & 168 & 9 \\
& 376 & 1 & 1128 & 8 & 240 & 9 \\
& 168 & 2 & 504 & 8 & 616 & 10 \\
& 520 & 3 & 1560 & 9 & 56 & 12 \\
& 496 & 2 & 1488 & 9 & 72 & 11 \\
& 424 & 2 & 1272 & 10 & 104 & 12 \\
& 392 & 2 & 1176 & 10 & 120 & 12 \\
& 280 & 2 & 840 & 10 & 192 & 12 \\
& 112 & 2 & 336 & 9 & 592 & 11 \\
& 448 & 3 & 1344 & 10 & 40 & 13 \\
& 416 & 2 & 1248 & 11 & 56 & 13 \\
& 344 & 2 & 1032 & 11 & 88 & 13 \\
& 296 & 2 & 888 & 11 & 120 & 13 \\
& 208 & 2 & 624 & 11 & 200 & 13 \\
& 80 & 4 & 240 & 9 & 688 & 13 \\
\bottomrule
\end{tabular}
\end{minipage}
\hfill
\begin{minipage}[t]{0.49\textwidth}
\captionsetup{font=small,justification=raggedright,singlelinecheck=false}
\centering
\caption{Specific architectural parameters ($d, L_d, d_d, L_m, d_m, L$) in Sec~\ref{sec:mnna_exp_confg}.}
\label{tab:setup_mnna2}
\small
\setlength{\tabcolsep}{4pt}
\renewcommand{\arraystretch}{1.2}
\begin{tabular}{l c c c c c c} %
\toprule
\textbf{$C$} & \textbf{$d$} & \textbf{$L_d$} & \textbf{$d_d$} & \textbf{$L_m$} & \textbf{$d_m$} & \textbf{$L$} \\
\midrule
\multirow{36}{*}{$3e18$} & 2080 & 1 & 6240 & 2 & 240 & 3 \\
& 1848 & 1 & 5544 & 2 & 360 & 3 \\
& 1568 & 1 & 4704 & 2 & 544 & 3 \\
& 1408 & 1 & 4224 & 2 & 672 & 3 \\
& 1048 & 1 & 3144 & 2 & 1080 & 3 \\
& 536 & 1 & 1608 & 2 & 2464 & 3 \\
& 1672 & 1 & 5016 & 4 & 128 & 5 \\
& 1416 & 1 & 4248 & 4 & 208 & 5 \\
& 1168 & 1 & 3504 & 4 & 328 & 5 \\
& 1048 & 1 & 3144 & 4 & 400 & 5 \\
& 760 & 1 & 2280 & 4 & 656 & 5 \\
& 440 & 1 & 1320 & 5 & 1096 & 6 \\
& 1272 & 2 & 3816 & 5 & 112 & 7 \\
& 1160 & 1 & 3480 & 6 & 160 & 7 \\
& 992 & 1 & 2976 & 6 & 240 & 7 \\
& 872 & 1 & 2616 & 6 & 304 & 7 \\
& 600 & 2 & 1800 & 5 & 536 & 7 \\
& 288 & 1 & 864 & 6 & 1208 & 7 \\
& 928 & 2 & 2784 & 7 & 96 & 9 \\
& 832 & 2 & 2496 & 8 & 136 & 10 \\
& 712 & 2 & 2136 & 8 & 200 & 10 \\
& 656 & 2 & 1968 & 8 & 232 & 10 \\
& 448 & 2 & 1344 & 8 & 400 & 10 \\
& 176 & 2 & 528 & 7 & 1328 & 9 \\
& 720 & 2 & 2160 & 10 & 72 & 12 \\
& 608 & 3 & 1824 & 10 & 120 & 13 \\
& 504 & 2 & 1512 & 10 & 176 & 12 \\
& 456 & 2 & 1368 & 10 & 208 & 12 \\
& 304 & 2 & 912 & 10 & 376 & 12 \\
& 120 & 3 & 360 & 9 & 1168 & 12 \\
& 584 & 3 & 1752 & 11 & 64 & 14 \\
& 512 & 2 & 1536 & 11 & 96 & 13 \\
& 392 & 3 & 1176 & 11 & 168 & 14 \\
& 344 & 3 & 1032 & 11 & 208 & 14 \\
& 232 & 3 & 696 & 11 & 360 & 14 \\
& 88 & 4 & 264 & 10 & 1264 & 14 \\ 
\bottomrule
\end{tabular}
\end{minipage}
\end{table}
\clearpage

\begin{table}[p]
\centering
\begin{minipage}[t]{0.49\textwidth}
\captionsetup{font=small,justification=raggedright,singlelinecheck=false}
\centering
\caption{Specific architectural parameters ($d, L_d, d_d, L_m, d_m, L$) in Sec~\ref{sec:mnna_exp_confg}.}
\label{tab:setup_mnna3}
\small
\setlength{\tabcolsep}{4pt}
\renewcommand{\arraystretch}{1.2}
\begin{tabular}{l c c c c c c} %
\toprule
\textbf{$C$} & \textbf{$d$} & \textbf{$L_d$} & \textbf{$d_d$} & \textbf{$L_m$} & \textbf{$d_m$} & \textbf{$L$} \\
\midrule
\multirow{36}{*}{$1e19$} & 2344 & 1 & 7032 & 4 & 192 & 5 \\
& 2112 & 1 & 6336 & 4 & 280 & 5 \\
& 1872 & 1 & 5616 & 4 & 392 & 5 \\
& 1744 & 1 & 5232 & 4 & 448 & 5 \\
& 1272 & 1 & 3816 & 4 & 744 & 5 \\
& 616 & 1 & 1848 & 4 & 1816 & 5 \\
& 1608 & 2 & 4824 & 7 & 136 & 9 \\
& 1440 & 2 & 4320 & 7 & 200 & 9 \\
& 1256 & 2 & 3768 & 7 & 288 & 9 \\
& 1184 & 2 & 3552 & 7 & 328 & 9 \\
& 928 & 1 & 2784 & 8 & 464 & 9 \\
& 408 & 1 & 1224 & 7 & 1256 & 8 \\
& 1336 & 2 & 4008 & 9 & 104 & 11 \\
& 1200 & 2 & 3600 & 9 & 152 & 11 \\
& 1040 & 2 & 3120 & 10 & 224 & 12 \\
& 960 & 2 & 2880 & 10 & 256 & 12 \\
& 696 & 2 & 2088 & 10 & 416 & 12 \\
& 280 & 2 & 840 & 9 & 1288 & 11 \\
& 1040 & 3 & 3120 & 13 & 80 & 16 \\
& 920 & 2 & 2760 & 13 & 112 & 15 \\
& 760 & 2 & 2280 & 13 & 176 & 15 \\
& 656 & 3 & 1968 & 13 & 232 & 16 \\
& 456 & 2 & 1368 & 13 & 392 & 15 \\
& 176 & 4 & 528 & 12 & 1328 & 16 \\
& 768 & 3 & 2304 & 17 & 64 & 20 \\
& 664 & 3 & 1992 & 17 & 96 & 20 \\
& 536 & 3 & 1608 & 17 & 160 & 20 \\
& 472 & 3 & 1416 & 17 & 192 & 20 \\
& 296 & 4 & 888 & 16 & 400 & 20 \\
& 112 & 7 & 336 & 13 & 1528 & 20 \\
& 624 & 3 & 1872 & 19 & 56 & 22 \\
& 528 & 3 & 1584 & 19 & 88 & 22 \\
& 416 & 4 & 1248 & 19 & 152 & 23 \\
& 360 & 4 & 1080 & 19 & 192 & 23 \\
& 224 & 5 & 672 & 17 & 400 & 22 \\
& 80 & 13 & 240 & 10 & 2256 & 23 \\
\bottomrule
\end{tabular}
\end{minipage}
\hfill
\begin{minipage}[t]{0.49\textwidth}
\captionsetup{font=small,justification=raggedright,singlelinecheck=false}
\centering
\caption{Specific architectural parameters ($d, L_d, d_d, L_m, d_m, L$) in Sec~\ref{sec:mnna_exp_confg}.}
\label{tab:setup_mnna4}
\small
\setlength{\tabcolsep}{4pt}
\renewcommand{\arraystretch}{1.2}
\begin{tabular}{l c c c c c c} %
\toprule
\textbf{$C$} & \textbf{$d$} & \textbf{$L_d$} & \textbf{$d_d$} & \textbf{$L_m$} & \textbf{$d_m$} & \textbf{$L$} \\
\midrule
\multirow{36}{*}{$3e19$} & 3344 & 1 & 10032 & 4 & 256 & 5 \\
& 2832 & 1 & 8496 & 4 & 416 & 5 \\
& 2368 & 1 & 7104 & 4 & 640 & 5 \\
& 2112 & 1 & 6336 & 4 & 792 & 5 \\
& 1520 & 1 & 4560 & 4 & 1312 & 5 \\
& 880 & 1 & 2640 & 5 & 2200 & 6 \\
& 2240 & 2 & 6720 & 7 & 184 & 9 \\
& 1944 & 2 & 5832 & 7 & 288 & 9 \\
& 1592 & 2 & 4776 & 7 & 456 & 9 \\
& 1424 & 2 & 4272 & 7 & 560 & 9 \\
& 1112 & 1 & 3336 & 8 & 784 & 9 \\
& 440 & 2 & 1320 & 7 & 2480 & 9 \\
& 1744 & 2 & 5232 & 10 & 160 & 12 \\
& 1496 & 2 & 4488 & 10 & 248 & 12 \\
& 1248 & 2 & 3744 & 10 & 376 & 12 \\
& 1128 & 2 & 3384 & 10 & 448 & 12 \\
& 776 & 2 & 2328 & 10 & 776 & 12 \\
& 304 & 3 & 912 & 9 & 2528 & 12 \\
& 1288 & 3 & 3864 & 14 & 120 & 17 \\
& 1112 & 3 & 3336 & 14 & 192 & 17 \\
& 888 & 3 & 2664 & 14 & 312 & 17 \\
& 768 & 3 & 2304 & 14 & 392 & 17 \\
& 496 & 3 & 1488 & 13 & 760 & 16 \\
& 184 & 6 & 552 & 10 & 3080 & 16 \\
& 896 & 4 & 2688 & 17 & 112 & 21 \\
& 760 & 4 & 2280 & 17 & 176 & 21 \\
& 592 & 4 & 1776 & 17 & 304 & 21 \\
& 504 & 5 & 1512 & 16 & 400 & 21 \\
& 320 & 6 & 960 & 15 & 808 & 21 \\
& 120 & 9 & 360 & 11 & 3288 & 20 \\
& 688 & 5 & 2064 & 19 & 112 & 24 \\
& 576 & 5 & 1728 & 19 & 176 & 24 \\
& 448 & 5 & 1344 & 18 & 296 & 23 \\
& 384 & 5 & 1152 & 18 & 384 & 23 \\
& 240 & 7 & 720 & 16 & 824 & 23 \\
& 88 & 14 & 264 & 10 & 4232 & 24 \\
\bottomrule
\end{tabular}
\end{minipage}
\end{table}
\clearpage

\begin{table}[p]
\centering
\begin{minipage}[t]{0.49\textwidth}
\captionsetup{font=small,justification=raggedright,singlelinecheck=false}
\centering
\caption{Specific architectural parameters ($d, L_d, d_d, L_m, d_m, L$) in Sec~\ref{sec:mnna_exp_confg}.}
\label{tab:setup_mnna5}
\small
\setlength{\tabcolsep}{4pt}
\renewcommand{\arraystretch}{1.2}
\begin{tabular}{l c c c c c c} %
\toprule
\textbf{$C$} & \textbf{$d$} & \textbf{$L_d$} & \textbf{$d_d$} & \textbf{$L_m$} & \textbf{$d_m$} & \textbf{$L$} \\
\midrule
\multirow{36}{*}{$1e20$} & 3072 & 2 & 9216 & 6 & 296 & 8 \\
& 2792 & 2 & 8376 & 6 & 432 & 8 \\
& 2456 & 2 & 7368 & 7 & 592 & 9 \\
& 2240 & 2 & 6720 & 7 & 712 & 9 \\
& 1808 & 1 & 5424 & 7 & 968 & 8 \\
& 824 & 1 & 2472 & 7 & 2464 & 8 \\
& 2368 & 3 & 7104 & 12 & 168 & 15 \\
& 2112 & 2 & 6336 & 12 & 256 & 14 \\
& 1808 & 2 & 5424 & 13 & 368 & 15 \\
& 1640 & 2 & 4920 & 13 & 440 & 15 \\
& 1112 & 2 & 3336 & 13 & 784 & 15 \\
& 424 & 4 & 1272 & 11 & 2736 & 15 \\
& 1912 & 3 & 5736 & 17 & 136 & 20 \\
& 1664 & 3 & 4992 & 17 & 208 & 20 \\
& 1392 & 3 & 4176 & 17 & 312 & 20 \\
& 1184 & 3 & 3552 & 17 & 416 & 20 \\
& 776 & 3 & 2328 & 16 & 776 & 19 \\
& 304 & 5 & 912 & 15 & 2528 & 20 \\
& 1360 & 4 & 4080 & 23 & 112 & 27 \\
& 1136 & 4 & 3408 & 23 & 184 & 27 \\
& 904 & 4 & 2712 & 23 & 296 & 27 \\
& 776 & 4 & 2328 & 22 & 384 & 26 \\
& 488 & 6 & 1464 & 21 & 792 & 27 \\
& 192 & 8 & 576 & 19 & 2520 & 27 \\
& 920 & 5 & 2760 & 28 & 104 & 33 \\
& 760 & 6 & 2280 & 28 & 176 & 34 \\
& 592 & 7 & 1776 & 27 & 304 & 34 \\
& 504 & 7 & 1512 & 26 & 400 & 33 \\
& 320 & 9 & 960 & 24 & 808 & 33 \\
& 120 & 15 & 360 & 18 & 3288 & 33 \\
& 696 & 7 & 2088 & 31 & 104 & 38 \\
& 576 & 8 & 1728 & 30 & 176 & 38 \\
& 448 & 9 & 1344 & 30 & 296 & 39 \\
& 376 & 10 & 1128 & 28 & 408 & 38 \\
& 240 & 12 & 720 & 26 & 824 & 38 \\
& 88 & 22 & 264 & 16 & 4232 & 38 \\
\bottomrule
\end{tabular}
\end{minipage}
\hfill
\begin{minipage}[t]{0.49\textwidth}
\captionsetup{font=small,justification=raggedright,singlelinecheck=false}
\centering
\caption{Specific architectural parameters ($d, L_d, d_d, L_m, d_m, L$) in Sec~\ref{sec:mnna_exp_confg}.}
\label{tab:setup_mnna6}
\small
\setlength{\tabcolsep}{4pt}
\renewcommand{\arraystretch}{1.2}
\begin{tabular}{l c c c c c c} %
\toprule
\textbf{$C$} & \textbf{$d$} & \textbf{$L_d$} & \textbf{$d_d$} & \textbf{$L_m$} & \textbf{$d_m$} & \textbf{$L$} \\
\midrule
\multirow{36}{*}{$3e20$} & 4336 & 2 & 13008 & 7 & 392 & 9 \\
& 3856 & 2 & 11568 & 7 & 592 & 9 \\
& 3304 & 2 & 9912 & 7 & 856 & 9 \\
& 3096 & 1 & 9288 & 7 & 968 & 8 \\
& 2080 & 2 & 6240 & 7 & 1752 & 9 \\
& 880 & 2 & 2640 & 7 & 4960 & 9 \\
& 2968 & 3 & 8904 & 13 & 272 & 16 \\
& 2568 & 3 & 7704 & 13 & 424 & 16 \\
& 2176 & 2 & 6528 & 13 & 616 & 15 \\
& 1888 & 2 & 5664 & 13 & 784 & 15 \\
& 1208 & 3 & 3624 & 13 & 1528 & 16 \\
& 464 & 4 & 1392 & 11 & 5288 & 15 \\
& 2288 & 3 & 6864 & 17 & 232 & 20 \\
& 1960 & 3 & 5880 & 17 & 360 & 20 \\
& 1528 & 4 & 4584 & 17 & 600 & 21 \\
& 1312 & 4 & 3936 & 17 & 784 & 21 \\
& 832 & 5 & 2496 & 16 & 1576 & 21 \\
& 320 & 7 & 960 & 14 & 5464 & 21 \\
& 1520 & 5 & 4560 & 23 & 208 & 28 \\
& 1256 & 5 & 3768 & 23 & 352 & 28 \\
& 976 & 6 & 2928 & 22 & 592 & 28 \\
& 824 & 7 & 2472 & 21 & 808 & 28 \\
& 520 & 9 & 1560 & 19 & 1672 & 28 \\
& 200 & 11 & 600 & 17 & 5616 & 28 \\
& 1000 & 8 & 3000 & 27 & 208 & 35 \\
& 824 & 8 & 2472 & 27 & 360 & 35 \\
& 632 & 10 & 1896 & 25 & 624 & 35 \\
& 544 & 10 & 1632 & 25 & 800 & 35 \\
& 344 & 12 & 1032 & 23 & 1640 & 35 \\
& 128 & 16 & 384 & 19 & 6248 & 35 \\
& 752 & 10 & 2256 & 30 & 216 & 40 \\
& 616 & 11 & 1848 & 29 & 368 & 40 \\
& 480 & 12 & 1440 & 28 & 616 & 40 \\
& 408 & 12 & 1224 & 27 & 816 & 39 \\
& 256 & 15 & 768 & 25 & 1720 & 40 \\
& 96 & 15 & 288 & 25 & 5272 & 40 \\
\bottomrule
\end{tabular}
\end{minipage}
\end{table}
\clearpage

\begin{table}[p]
\centering
\begin{minipage}[t]{0.49\textwidth}
\captionsetup{font=small,justification=raggedright,singlelinecheck=false}
\centering
\caption{Specific architectural parameters ($d, L_d, d_d, L_m, d_m, L$) in Sec~\ref{sec:mnna_scaling_laws_result}.}
\label{tab:setup_mnna_v2}
\small
\setlength{\tabcolsep}{4pt}
\renewcommand{\arraystretch}{1.05}
\begin{tabular}{l c c c c c c} %
\toprule
\textbf{$C$} & \textbf{$d$} & \textbf{$L_d$} & \textbf{$d_d$} & \textbf{$L_m$} & \textbf{$d_m$} & \textbf{$L$} \\
\midrule
\multirow{10}{*}{$1e18$} & 1024 & 1 & 3072 & 4 & 152 & 5 \\
& 792 & 2 & 2376 & 5 & 160 & 7 \\
& 680 & 2 & 2040 & 6 & 120 & 8 \\
& 680 & 2 & 2040 & 7 & 96 & 9 \\
& 616 & 2 & 1848 & 8 & 96 & 10 \\
& 568 & 2 & 1704 & 8 & 88 & 10 \\
& 512 & 2 & 1536 & 9 & 80 & 11 \\
& 480 & 2 & 1440 & 10 & 80 & 12 \\
& 456 & 2 & 1368 & 11 & 72 & 13 \\
& 416 & 2 & 1248 & 11 & 72 & 13 \\ \midrule
\multirow{8}{*}{$3e18$} & 1136 & 1 & 3408 & 4 & 344 & 5 \\
& 1024 & 1 & 3072 & 5 & 296 & 6 \\
& 808 & 2 & 2424 & 5 & 344 & 7 \\
& 784 & 1 & 2352 & 6 & 288 & 7 \\
& 744 & 1 & 2232 & 6 & 256 & 7 \\
& 648 & 1 & 1944 & 7 & 232 & 8 \\
& 584 & 1 & 1752 & 8 & 208 & 9 \\
& 496 & 2 & 1488 & 8 & 224 & 10 \\ \midrule
\multirow{8}{*}{$1e19$} & 1120 & 2 & 3360 & 7 & 352 & 9 \\
& 1040 & 2 & 3120 & 8 & 328 & 10 \\
& 1008 & 2 & 3024 & 10 & 272 & 12 \\
& 936 & 2 & 2808 & 10 & 264 & 12 \\
& 896 & 2 & 2688 & 12 & 232 & 14 \\
& 784 & 2 & 2352 & 13 & 240 & 15 \\
& 664 & 3 & 1992 & 14 & 224 & 17 \\
& 592 & 3 & 1776 & 16 & 208 & 19 \\ \midrule
\multirow{7}{*}{$3e19$} & 1864 & 1 & 5592 & 6 & 536 & 7 \\
& 1448 & 2 & 4344 & 7 & 544 & 9 \\
& 1272 & 2 & 3816 & 8 & 480 & 10 \\
& 1120 & 2 & 3360 & 9 & 456 & 11 \\
& 1056 & 2 & 3168 & 11 & 400 & 13 \\
& 952 & 2 & 2856 & 12 & 384 & 14 \\
& 776 & 3 & 2328 & 13 & 384 & 16 \\ \midrule
\multirow{8}{*}{$1e20$} & 2232 & 2 & 6696 & 7 & 712 & 9 \\
& 2080 & 2 & 6240 & 10 & 552 & 12 \\
& 1768 & 2 & 5304 & 12 & 488 & 14 \\
& 1600 & 2 & 4800 & 14 & 456 & 16 \\
& 1384 & 3 & 4152 & 17 & 416 & 20 \\
& 1192 & 3 & 3576 & 19 & 408 & 22 \\
& 936 & 4 & 2808 & 22 & 400 & 26 \\
& 768 & 5 & 2304 & 24 & 392 & 29 \\ \midrule
\multirow{8}{*}{$3e20$} & 3712 & 1 & 11136 & 4 & 1520 & 5 \\
& 3464 & 1 & 10392 & 6 & 1112 & 7 \\
& 2680 & 2 & 8040 & 8 & 936 & 10 \\
& 2280 & 2 & 6840 & 10 & 880 & 12 \\
& 1816 & 3 & 5448 & 12 & 840 & 15 \\
& 1544 & 3 & 4632 & 14 & 784 & 17 \\
& 1304 & 4 & 3912 & 16 & 792 & 20 \\
& 1008 & 5 & 3024 & 18 & 800 & 23 \\
\bottomrule
\end{tabular}
\end{minipage}
\hfill
\begin{minipage}[t]{0.49\textwidth}
\captionsetup{font=small,justification=raggedright,singlelinecheck=false}
\centering
\caption{Specific architectural parameters ($d, L_d, d_d, L_m, d_m, L$) in Sec~\ref{sec:mnnad}.}
\label{tab:setup_mnnad1}
\small
\setlength{\tabcolsep}{4pt}
\renewcommand{\arraystretch}{0.95}
\begin{tabular}{l c c c c c c} %
\toprule
\textbf{$C$} & \textbf{$d$} & \textbf{$L_d$} & \textbf{$d_d$} & \textbf{$L_m$} & \textbf{$d_m$} & \textbf{$L$} \\
\midrule
\multirow{54}{*}{$1e18$} & 288 & 8 & 864 & 2 & 1136 & 10 \\
& 304 & 7 & 912 & 3 & 712 & 10 \\
& 312 & 6 & 936 & 3 & 608 & 9 \\
& 328 & 6 & 984 & 4 & 472 & 10 \\
& 336 & 5 & 1008 & 4 & 424 & 9 \\
& 344 & 5 & 1032 & 5 & 384 & 10 \\
& 352 & 5 & 1056 & 5 & 360 & 10 \\
& 360 & 5 & 1080 & 5 & 328 & 10 \\
& 368 & 4 & 1104 & 5 & 312 & 9 \\
& 376 & 4 & 1128 & 6 & 288 & 10 \\
& 384 & 4 & 1152 & 6 & 272 & 10 \\
& 392 & 4 & 1176 & 6 & 256 & 10 \\
& 400 & 3 & 1200 & 6 & 248 & 9 \\
& 408 & 3 & 1224 & 6 & 232 & 9 \\
& 416 & 3 & 1248 & 7 & 224 & 10 \\
& 424 & 3 & 1272 & 7 & 216 & 10 \\
& 432 & 3 & 1296 & 7 & 208 & 10 \\
& 440 & 3 & 1320 & 7 & 200 & 10 \\
& 448 & 3 & 1344 & 7 & 192 & 10 \\
& 456 & 2 & 1368 & 7 & 184 & 9 \\
& 464 & 2 & 1392 & 7 & 176 & 9 \\
& 472 & 2 & 1416 & 7 & 176 & 9 \\
& 480 & 2 & 1440 & 8 & 168 & 10 \\
& 488 & 2 & 1464 & 8 & 160 & 10 \\
& 496 & 2 & 1488 & 8 & 160 & 10 \\
& 504 & 2 & 1512 & 8 & 152 & 10 \\
& 512 & 2 & 1536 & 8 & 152 & 10 \\
& 520 & 2 & 1560 & 8 & 144 & 10 \\
& 528 & 2 & 1584 & 8 & 144 & 10 \\
& 536 & 2 & 1608 & 8 & 144 & 10 \\
& 544 & 1 & 1632 & 8 & 136 & 9 \\
& 552 & 1 & 1656 & 8 & 136 & 9 \\
& 560 & 1 & 1680 & 8 & 128 & 9 \\
& 568 & 1 & 1704 & 8 & 128 & 9 \\
& 576 & 1 & 1728 & 8 & 128 & 9 \\
& 584 & 1 & 1752 & 8 & 120 & 9 \\
& 592 & 1 & 1776 & 8 & 120 & 9 \\
& 600 & 1 & 1800 & 9 & 120 & 10 \\
& 608 & 1 & 1824 & 9 & 120 & 10 \\
& 616 & 1 & 1848 & 9 & 112 & 10 \\
& 624 & 1 & 1872 & 9 & 112 & 10 \\
& 632 & 1 & 1896 & 9 & 112 & 10 \\
& 640 & 1 & 1920 & 9 & 112 & 10 \\
& 648 & 1 & 1944 & 9 & 104 & 10 \\
& 656 & 1 & 1968 & 9 & 104 & 10 \\
& 664 & 1 & 1992 & 9 & 104 & 10 \\
& 672 & 1 & 2016 & 9 & 104 & 10 \\
& 680 & 1 & 2040 & 9 & 104 & 10 \\
& 688 & 1 & 2064 & 9 & 96 & 10 \\
& 696 & 1 & 2088 & 9 & 96 & 10 \\
& 704 & 1 & 2112 & 9 & 96 & 10 \\
& 712 & 1 & 2136 & 9 & 96 & 10 \\
& 736 & 1 & 2208 & 9 & 88 & 10 \\
& 744 & 1 & 2232 & 9 & 88 & 10 \\
\bottomrule
\end{tabular}
\end{minipage}
\end{table}
\clearpage

\begin{table}[p]
\centering
\begin{minipage}[t]{0.49\textwidth}
\captionsetup{font=small,justification=raggedright,singlelinecheck=false}
\centering
\caption{Specific architectural parameters ($d, L_d, d_d, L_m, d_m, L$) in Sec~\ref{sec:mnnad}.}
\label{tab:setup_mnnad2}
\small
\setlength{\tabcolsep}{4pt}
\renewcommand{\arraystretch}{1.05}
\begin{tabular}{l c c c c c c} %
\toprule
\textbf{$C$} & \textbf{$d$} & \textbf{$L_d$} & \textbf{$d_d$} & \textbf{$L_m$} & \textbf{$d_m$} & \textbf{$L$} \\
\midrule
\multirow{40}{*}{$3e18$} & 456 & 6 & 1368 & 2 & 1376 & 8 \\
& 464 & 5 & 1392 & 2 & 1200 & 7 \\
& 488 & 5 & 1464 & 3 & 880 & 8 \\
& 496 & 5 & 1488 & 3 & 816 & 8 \\
& 504 & 4 & 1512 & 3 & 752 & 7 \\
& 528 & 4 & 1584 & 4 & 624 & 8 \\
& 536 & 4 & 1608 & 4 & 592 & 8 \\
& 544 & 4 & 1632 & 4 & 560 & 8 \\
& 552 & 3 & 1656 & 4 & 536 & 7 \\
& 560 & 3 & 1680 & 4 & 512 & 7 \\
& 568 & 3 & 1704 & 4 & 488 & 7 \\
& 584 & 3 & 1752 & 5 & 448 & 8 \\
& 592 & 3 & 1776 & 5 & 432 & 8 \\
& 600 & 3 & 1800 & 5 & 416 & 8 \\
& 608 & 3 & 1824 & 5 & 408 & 8 \\
& 616 & 3 & 1848 & 5 & 392 & 8 \\
& 632 & 2 & 1896 & 5 & 368 & 7 \\
& 648 & 2 & 1944 & 5 & 352 & 7 \\
& 664 & 2 & 1992 & 6 & 336 & 8 \\
& 680 & 2 & 2040 & 6 & 320 & 8 \\
& 696 & 2 & 2088 & 6 & 304 & 8 \\
& 712 & 2 & 2136 & 6 & 288 & 8 \\
& 728 & 2 & 2184 & 6 & 280 & 8 \\
& 744 & 1 & 2232 & 6 & 264 & 7 \\
& 760 & 1 & 2280 & 6 & 256 & 7 \\
& 776 & 1 & 2328 & 6 & 248 & 7 \\
& 792 & 1 & 2376 & 6 & 240 & 7 \\
& 808 & 1 & 2424 & 7 & 232 & 8 \\
& 824 & 1 & 2472 & 7 & 224 & 8 \\
& 840 & 1 & 2520 & 7 & 216 & 8 \\
& 856 & 1 & 2568 & 7 & 216 & 8 \\
& 872 & 1 & 2616 & 7 & 208 & 8 \\
& 888 & 1 & 2664 & 7 & 200 & 8 \\
& 904 & 1 & 2712 & 7 & 200 & 8 \\
& 920 & 1 & 2760 & 7 & 192 & 8 \\
& 936 & 1 & 2808 & 7 & 184 & 8 \\
& 952 & 1 & 2856 & 7 & 184 & 8 \\
& 968 & 1 & 2904 & 7 & 176 & 8 \\
& 984 & 1 & 2952 & 7 & 176 & 8 \\
& 1000 & 1 & 3000 & 7 & 168 & 8 \\
\bottomrule
\end{tabular}
\end{minipage}
\hfill
\begin{minipage}[t]{0.49\textwidth}
\captionsetup{font=small,justification=raggedright,singlelinecheck=false}
\centering
\caption{Specific architectural parameters ($d, L_d, d_d, L_m, d_m, L$) in Sec~\ref{sec:mnnad}.}
\label{tab:setup_mnnad3}
\small
\setlength{\tabcolsep}{4pt}
\renewcommand{\arraystretch}{1.05}
\begin{tabular}{l c c c c c c} %
\toprule
\textbf{$C$} & \textbf{$d$} & \textbf{$L_d$} & \textbf{$d_d$} & \textbf{$L_m$} & \textbf{$d_m$} & \textbf{$L$} \\
\midrule
\multirow{40}{*}{$1e19$} & 648 & 5 & 1944 & 8 & 504 & 13 \\
& 664 & 5 & 1992 & 8 & 472 & 13 \\
& 680 & 5 & 2040 & 8 & 448 & 13 \\
& 696 & 4 & 2088 & 9 & 424 & 13 \\
& 712 & 4 & 2136 & 9 & 400 & 13 \\
& 728 & 4 & 2184 & 9 & 384 & 13 \\
& 744 & 4 & 2232 & 9 & 360 & 13 \\
& 760 & 3 & 2280 & 10 & 344 & 13 \\
& 776 & 3 & 2328 & 10 & 336 & 13 \\
& 792 & 3 & 2376 & 10 & 320 & 13 \\
& 808 & 3 & 2424 & 10 & 312 & 13 \\
& 824 & 3 & 2472 & 10 & 296 & 13 \\
& 840 & 3 & 2520 & 10 & 288 & 13 \\
& 856 & 2 & 2568 & 11 & 280 & 13 \\
& 872 & 2 & 2616 & 11 & 272 & 13 \\
& 888 & 2 & 2664 & 11 & 264 & 13 \\
& 904 & 2 & 2712 & 11 & 256 & 13 \\
& 920 & 2 & 2760 & 11 & 248 & 13 \\
& 936 & 2 & 2808 & 11 & 240 & 13 \\
& 952 & 2 & 2856 & 11 & 232 & 13 \\
& 968 & 2 & 2904 & 11 & 232 & 13 \\
& 984 & 2 & 2952 & 11 & 224 & 13 \\
& 1000 & 1 & 3000 & 12 & 216 & 13 \\
& 1016 & 1 & 3048 & 12 & 216 & 13 \\
& 1032 & 1 & 3096 & 12 & 208 & 13 \\
& 1048 & 1 & 3144 & 12 & 208 & 13 \\
& 1064 & 1 & 3192 & 12 & 200 & 13 \\
& 1080 & 1 & 3240 & 12 & 200 & 13 \\
& 1096 & 1 & 3288 & 12 & 192 & 13 \\
& 1112 & 1 & 3336 & 12 & 192 & 13 \\
& 1128 & 1 & 3384 & 12 & 184 & 13 \\
& 1144 & 1 & 3432 & 12 & 184 & 13 \\
& 1160 & 1 & 3480 & 12 & 176 & 13 \\
& 1176 & 1 & 3528 & 12 & 176 & 13 \\
& 1192 & 1 & 3576 & 12 & 176 & 13 \\
& 1208 & 1 & 3624 & 12 & 168 & 13 \\
& 1224 & 1 & 3672 & 12 & 168 & 13 \\
& 1240 & 1 & 3720 & 12 & 168 & 13 \\
& 1256 & 1 & 3768 & 12 & 160 & 13 \\
& 1272 & 1 & 3816 & 12 & 160 & 13 \\
\bottomrule
\end{tabular}
\end{minipage}
\end{table}
\clearpage

\begin{table}[p]
\centering
\begin{minipage}[t]{0.49\textwidth}
\captionsetup{font=small,justification=raggedright,singlelinecheck=false}
\centering
\caption{Specific architectural parameters ($d, L_d, d_d, L_m, d_m, L$) in Sec~\ref{sec:mnnad}.}
\label{tab:setup_mnnad4}
\small
\setlength{\tabcolsep}{4pt}
\renewcommand{\arraystretch}{1.05}
\begin{tabular}{l c c c c c c} %
\toprule
\textbf{$C$} & \textbf{$d$} & \textbf{$L_d$} & \textbf{$d_d$} & \textbf{$L_m$} & \textbf{$d_m$} & \textbf{$L$} \\
\midrule
\multirow{43}{*}{$3e19$} & 832 & 6 & 2496 & 5 & 1168 & 11 \\
& 840 & 6 & 2520 & 5 & 1120 & 11 \\
& 888 & 5 & 2664 & 6 & 920 & 11 \\
& 896 & 5 & 2688 & 6 & 888 & 11 \\
& 904 & 5 & 2712 & 6 & 864 & 11 \\
& 952 & 4 & 2856 & 7 & 744 & 11 \\
& 960 & 4 & 2880 & 7 & 728 & 11 \\
& 968 & 4 & 2904 & 7 & 712 & 11 \\
& 976 & 4 & 2928 & 7 & 696 & 11 \\
& 984 & 4 & 2952 & 7 & 688 & 11 \\
& 1048 & 3 & 3144 & 8 & 592 & 11 \\
& 1056 & 3 & 3168 & 8 & 584 & 11 \\
& 1064 & 3 & 3192 & 8 & 576 & 11 \\
& 1072 & 3 & 3216 & 8 & 560 & 11 \\
& 1080 & 3 & 3240 & 8 & 552 & 11 \\
& 1088 & 3 & 3264 & 8 & 544 & 11 \\
& 1112 & 2 & 3336 & 8 & 520 & 10 \\
& 1136 & 2 & 3408 & 8 & 504 & 10 \\
& 1160 & 2 & 3480 & 9 & 480 & 11 \\
& 1184 & 2 & 3552 & 9 & 464 & 11 \\
& 1208 & 2 & 3624 & 9 & 448 & 11 \\
& 1232 & 2 & 3696 & 9 & 432 & 11 \\
& 1256 & 2 & 3768 & 9 & 424 & 11 \\
& 1280 & 1 & 3840 & 9 & 408 & 10 \\
& 1304 & 1 & 3912 & 9 & 392 & 10 \\
& 1328 & 1 & 3984 & 9 & 384 & 10 \\
& 1352 & 1 & 4056 & 10 & 376 & 11 \\
& 1376 & 1 & 4128 & 10 & 368 & 11 \\
& 1400 & 1 & 4200 & 10 & 352 & 11 \\
& 1424 & 1 & 4272 & 10 & 344 & 11 \\
& 1448 & 1 & 4344 & 10 & 336 & 11 \\
& 1472 & 1 & 4416 & 10 & 328 & 11 \\
& 1496 & 1 & 4488 & 10 & 320 & 11 \\
& 1520 & 1 & 4560 & 10 & 320 & 11 \\
& 1544 & 1 & 4632 & 10 & 312 & 11 \\
& 1568 & 1 & 4704 & 10 & 304 & 11 \\
& 1592 & 1 & 4776 & 10 & 296 & 11 \\
& 1616 & 1 & 4848 & 10 & 288 & 11 \\
& 1640 & 1 & 4920 & 10 & 288 & 11 \\
& 1664 & 1 & 4992 & 10 & 280 & 11 \\
& 1688 & 1 & 5064 & 10 & 272 & 11 \\
& 1712 & 1 & 5136 & 10 & 272 & 11 \\
& 1736 & 1 & 5208 & 10 & 264 & 11 \\
\bottomrule
\end{tabular}
\end{minipage}
\hfill
\begin{minipage}[t]{0.49\textwidth}
\captionsetup{font=small,justification=raggedright,singlelinecheck=false}
\centering
\caption{Specific architectural parameters ($d, L_d, d_d, L_m, d_m, L$) in Sec~\ref{sec:mnnad}.}
\label{tab:setup_mnnad5}
\small
\setlength{\tabcolsep}{4pt}
\renewcommand{\arraystretch}{1.05}
\begin{tabular}{l c c c c c c} %
\toprule
\textbf{$C$} & \textbf{$d$} & \textbf{$L_d$} & \textbf{$d_d$} & \textbf{$L_m$} & \textbf{$d_m$} & \textbf{$L$} \\
\midrule
\multirow{42}{*}{$1e20$} & 1024 & 7 & 3072 & 11 & 880 & 18 \\
& 1040 & 7 & 3120 & 11 & 840 & 18 \\
& 1072 & 6 & 3216 & 12 & 776 & 18 \\
& 1088 & 6 & 3264 & 12 & 744 & 18 \\
& 1104 & 6 & 3312 & 12 & 720 & 18 \\
& 1136 & 5 & 3408 & 13 & 672 & 18 \\
& 1152 & 5 & 3456 & 13 & 648 & 18 \\
& 1168 & 5 & 3504 & 13 & 632 & 18 \\
& 1216 & 4 & 3648 & 14 & 584 & 18 \\
& 1232 & 4 & 3696 & 14 & 568 & 18 \\
& 1248 & 4 & 3744 & 14 & 552 & 18 \\
& 1312 & 3 & 3936 & 15 & 504 & 18 \\
& 1328 & 3 & 3984 & 15 & 496 & 18 \\
& 1344 & 3 & 4032 & 15 & 480 & 18 \\
& 1360 & 3 & 4080 & 15 & 472 & 18 \\
& 1376 & 3 & 4128 & 15 & 464 & 18 \\
& 1392 & 3 & 4176 & 15 & 456 & 18 \\
& 1448 & 2 & 4344 & 16 & 424 & 18 \\
& 1464 & 2 & 4392 & 16 & 416 & 18 \\
& 1480 & 2 & 4440 & 16 & 416 & 18 \\
& 1496 & 2 & 4488 & 16 & 408 & 18 \\
& 1512 & 2 & 4536 & 16 & 400 & 18 \\
& 1528 & 2 & 4584 & 16 & 392 & 18 \\
& 1544 & 2 & 4632 & 16 & 384 & 18 \\
& 1560 & 2 & 4680 & 16 & 384 & 18 \\
& 1576 & 2 & 4728 & 17 & 376 & 19 \\
& 1592 & 2 & 4776 & 17 & 368 & 19 \\
& 1624 & 2 & 4872 & 17 & 360 & 19 \\
& 1640 & 1 & 4920 & 17 & 352 & 18 \\
& 1656 & 1 & 4968 & 17 & 352 & 18 \\
& 1672 & 1 & 5016 & 17 & 344 & 18 \\
& 1688 & 1 & 5064 & 17 & 344 & 18 \\
& 1704 & 1 & 5112 & 17 & 336 & 18 \\
& 1720 & 1 & 5160 & 17 & 336 & 18 \\
& 1736 & 1 & 5208 & 17 & 328 & 18 \\
& 1752 & 1 & 5256 & 17 & 328 & 18 \\
& 1768 & 1 & 5304 & 17 & 320 & 18 \\
& 1784 & 1 & 5352 & 17 & 320 & 18 \\
& 1800 & 1 & 5400 & 17 & 312 & 18 \\
& 1816 & 1 & 5448 & 17 & 312 & 18 \\
& 1832 & 1 & 5496 & 17 & 304 & 18 \\
& 1864 & 1 & 5592 & 18 & 296 & 19 \\
\bottomrule
\end{tabular}
\end{minipage}
\end{table}
\clearpage

\begin{table}[htbp]
\centering
\caption{Specific architectural parameters ($d, L_d, d_d, L_m, d_m, L$) in Sec~\ref{sec:mnnad}.}
\label{tab:setup_mnnad6}
\renewcommand{\arraystretch}{1.2}
\resizebox{0.5\linewidth}{!}{%
\begin{tabular}{l c c c c c c} %
\toprule
\textbf{$C$} & \textbf{$d$} & \textbf{$L_d$} & \textbf{$d_d$} & \textbf{$L_m$} & \textbf{$d_m$} & \textbf{$L$} \\
\midrule
\multirow{39}{*}{$3e20$} & 1248 & 8 & 3744 & 8 & 1832 & 16 \\
& 1312 & 7 & 3936 & 9 & 1520 & 16 \\
& 1376 & 6 & 4128 & 10 & 1304 & 16 \\
& 1440 & 5 & 4320 & 11 & 1152 & 16 \\
& 1472 & 5 & 4416 & 11 & 1088 & 16 \\
& 1536 & 4 & 4608 & 12 & 984 & 16 \\
& 1568 & 4 & 4704 & 12 & 936 & 16 \\
& 1632 & 3 & 4896 & 13 & 864 & 16 \\
& 1664 & 3 & 4992 & 13 & 832 & 16 \\
& 1696 & 3 & 5088 & 13 & 800 & 16 \\
& 1736 & 2 & 5208 & 14 & 768 & 16 \\
& 1752 & 2 & 5256 & 14 & 752 & 16 \\
& 1768 & 2 & 5304 & 14 & 744 & 16 \\
& 1784 & 2 & 5352 & 14 & 728 & 16 \\
& 1800 & 2 & 5400 & 14 & 720 & 16 \\
& 1816 & 2 & 5448 & 14 & 712 & 16 \\
& 1832 & 2 & 5496 & 14 & 696 & 16 \\
& 1848 & 2 & 5544 & 14 & 688 & 16 \\
& 1864 & 2 & 5592 & 14 & 680 & 16 \\
& 1880 & 2 & 5640 & 14 & 672 & 16 \\
& 1896 & 2 & 5688 & 14 & 664 & 16 \\
& 1928 & 1 & 5784 & 15 & 640 & 16 \\
& 1944 & 1 & 5832 & 15 & 632 & 16 \\
& 1960 & 1 & 5880 & 15 & 624 & 16 \\
& 1976 & 1 & 5928 & 15 & 616 & 16 \\
& 1992 & 1 & 5976 & 15 & 608 & 16 \\
& 2008 & 1 & 6024 & 15 & 608 & 16 \\
& 2024 & 1 & 6072 & 15 & 600 & 16 \\
& 2040 & 1 & 6120 & 15 & 592 & 16 \\
& 2056 & 1 & 6168 & 15 & 584 & 16 \\
& 2072 & 1 & 6216 & 15 & 576 & 16 \\
& 2088 & 1 & 6264 & 15 & 568 & 16 \\
& 2104 & 1 & 6312 & 15 & 568 & 16 \\
& 2120 & 1 & 6360 & 15 & 560 & 16 \\
& 2136 & 1 & 6408 & 15 & 552 & 16 \\
& 2152 & 1 & 6456 & 15 & 544 & 16 \\
& 2168 & 1 & 6504 & 15 & 544 & 16 \\
& 2184 & 1 & 6552 & 15 & 536 & 16 \\
& 2200 & 1 & 6600 & 15 & 528 & 16 \\
\bottomrule
\end{tabular}}
\end{table}

\subsection{Detailed analysis and selection of $N/N_a$}
\label{app:n_na_setting}

\begin{table}[htbp]
\centering
\caption{Representative pipeline outputs and non-embedding parameter-storage estimates. The flat loss landscape permits a broad near-optimal region; these fitted outputs are not the realized training configurations.}
\label{tab:pipeline_memory}
\begingroup{}
\begin{tabular}{lrrrr}
\toprule
$C$ (FLOPs) & $M^{\mathrm{opt}}$ (GFLOPs) & $N_a^{\mathrm{opt}}$ (B) & $N$ (B) & FP16 storage (GB) \\
\midrule
$10^{18}$ & 0.2672 & 0.025 & 0.49 & 1.0 \\
$3\times10^{18}$ & 0.4856 & 0.049 & 0.95 & 1.9 \\
$10^{19}$ & 0.9345 & 0.100 & 2.08 & 4.2 \\
$3\times10^{19}$ & 1.6983 & 0.192 & 3.99 & 8.0 \\
$10^{20}$ & 3.2681 & 0.388 & 8.25 & 16.5 \\
$3\times10^{20}$ & 5.9390 & 0.732 & 16.16 & 32.3 \\
\bottomrule
\end{tabular}
\endgroup{}
\par\smallskip
Storage uses 2 bytes per parameter (FP16; decimal GB) and excludes embeddings, KV cache, activations, and optimizer states. Trained configurations used manually specified $N/N_a$ values of 19--22 and also differ because structural parameters must satisfy integer and hardware-alignment constraints.
\end{table}

Given that the feasible range for $N/N_a$ is $(1, 289/9)$, we employed a function of the form $L = a/(x-C_1) + b/(C_2-x) + c$, where $L$ represents the loss and $x$ is the $N/N_a$ ratio. This functional form was chosen to capture potential asymptotic behaviors at the boundaries of the valid range. By finding the minimum of this fitted curve, we derived the optimal $N/N_a$ values for each scale: $\{19.23, 19.33, 20.69, 20.78, 21.29, 22.07\}$.
However, a crucial observation emerged regarding the reliability of these fitted optimal values. We found that the loss landscape in the vicinity of these optimal $N/N_a$ ratios is notably flat. Specifically, within the range of $N/N_a \in [18, 25]$, the change in loss was consistently less than 1\%. This pronounced flatness implies that small variations in $N/N_a$ around the theoretical optimum have minimal impact on the overall performance, thereby reducing the practical significance of the precise fitted values.

Considering the inherent unreliability of fitting in such a flat landscape, and based on the direct empirical observations from Sec.~\ref{sec:mnna_scaling_laws_result}, we opted to manually specify the $N/N_a$ ratio for our subsequent experiments. The chosen $(C, N/N_a)$ pairs are as follows: $\{(1 \times 10^{18}, 19), (3 \times 10^{18}, 20), (1 \times 10^{19}, 21), (3 \times 10^{19}, 21), (1 \times 10^{20}, 22), (3 \times 10^{20}, 22)\}$.
During the actual model configuration and implementation, several practical constraints influenced the exact realization of these $N/N_a$ values. Firstly, we should ensure that the GFLOPs per token ($M$) requirements were rigorously met for each compute scale. Secondly, for hardware efficiency, all core model parameters (e.g., layer counts, hidden dimensions) were constrained to be multiples of $2^3$, $2^4$, or $2^5$. These practical considerations meant that while we couldn't always precisely match our manually specified targets. Nevertheless, given the previously identified flatness of the loss landscape in this range, we anticipate that these minor deviations would not materially affect the experimental results or conclusions.

\newpage
\section*{NeurIPS paper checklist}

\begin{enumerate}

\item {\bf Claims}
    \item[] Question: Do the main claims made in the abstract and introduction accurately reflect the paper's contributions and scope?
    \item[] Answer: \answerYes{}
    \item[] Justification: The abstract and Sec.~\ref{sec:intro} state three contributions (global MoE architectural scaling laws, the dimension decomposition methodology, and the widening near-optimal band); each is supported by the methodological analysis in Sec.~\ref{sec:decoup} and the experimental results in Sec.~\ref{sec:mnna}--\ref{sec:mnnad} ($\sim$670 MoE models across six compute scales from $10^{18}$ to $3\times 10^{20}$ FLOPs).
    \item[] Guidelines:
    \begin{itemize}
        \item The answer \answerNA{} means that the abstract and introduction do not include the claims made in the paper.
        \item The abstract and/or introduction should clearly state the claims made, including the contributions made in the paper and important assumptions and limitations. A \answerNo{} or \answerNA{} answer to this question will not be perceived well by the reviewers.
        \item The claims made should match theoretical and experimental results, and reflect how much the results can be expected to generalize to other settings.
        \item It is fine to include aspirational goals as motivation as long as it is clear that these goals are not attained by the paper.
    \end{itemize}

\item {\bf Limitations}
    \item[] Question: Does the paper discuss the limitations of the work performed by the authors?
    \item[] Answer: \answerYes{}
    \item[] Justification: A dedicated paragraph in Sec.~\ref{sec:dis} summarizes the limitations, with further details in Appendix~\ref{app:limitations}: experimental scale ($\sim$670 models), the use of heuristic architectures when fitting the $(C, M, D)$ scaling laws, fixed sparsity configuration ($N_e{=}288$, $K{=}8$), and evaluation restricted to pre-training cross-entropy loss rather than downstream tasks.
    \item[] Guidelines:
    \begin{itemize}
        \item The answer \answerNA{} means that the paper has no limitation while the answer \answerNo{} means that the paper has limitations, but those are not discussed in the paper.
        \item The authors are encouraged to create a separate ``Limitations'' section in their paper.
        \item The paper should point out any strong assumptions and how robust the results are to violations of these assumptions (e.g., independence assumptions, noiseless settings, model well-specification, asymptotic approximations only holding locally). The authors should reflect on how these assumptions might be violated in practice and what the implications would be.
        \item The authors should reflect on the scope of the claims made, e.g., if the approach was only tested on a few datasets or with a few runs. In general, empirical results often depend on implicit assumptions, which should be articulated.
        \item The authors should reflect on the factors that influence the performance of the approach. For example, a facial recognition algorithm may perform poorly when image resolution is low or images are taken in low lighting. Or a speech-to-text system might not be used reliably to provide closed captions for online lectures because it fails to handle technical jargon.
        \item The authors should discuss the computational efficiency of the proposed algorithms and how they scale with dataset size.
        \item If applicable, the authors should discuss possible limitations of their approach to address problems of privacy and fairness.
        \item While the authors might fear that complete honesty about limitations might be used by reviewers as grounds for rejection, a worse outcome might be that reviewers discover limitations that aren't acknowledged in the paper. The authors should use their best judgment and recognize that individual actions in favor of transparency play an important role in developing norms that preserve the integrity of the community. Reviewers will be specifically instructed to not penalize honesty concerning limitations.
    \end{itemize}

\item {\bf Theory assumptions and proofs}
    \item[] Question: For each theoretical result, does the paper provide the full set of assumptions and a complete (and correct) proof?
    \item[] Answer: \answerNA{}
    \item[] Justification: The paper does not contain formal theorems requiring proofs. Its mathematical content consists of explicit algebraic identities (the fundamental equations in Eq.~\eqref{eq:fundamental_equations} and the ratio-form constraints in Eq.~\eqref{eq:architectural_constraints}) and an empirically validated property (the rank-preserving behaviour of the hidden dimension $d$, validated in Sec.~\ref{subsec:efficient_search_strategy} and Appendix~\ref{app:median_proxy_validation} with Pearson $r{=}0.9793$, Spearman $\rho{=}0.9758$). Layer-wise derivations of the FLOPs formula are given in Appendix~\ref{app:def_model_scale}.
    \item[] Guidelines:
    \begin{itemize}
        \item The answer \answerNA{} means that the paper does not include theoretical results.
        \item All the theorems, formulas, and proofs in the paper should be numbered and cross-referenced.
        \item All assumptions should be clearly stated or referenced in the statement of any theorems.
        \item The proofs can either appear in the main paper or the supplemental material, but if they appear in the supplemental material, the authors are encouraged to provide a short proof sketch to provide intuition.
        \item Inversely, any informal proof provided in the core of the paper should be complemented by formal proofs provided in appendix or supplemental material.
        \item Theorems and Lemmas that the proof relies upon should be properly referenced.
    \end{itemize}

    \item {\bf Experimental result reproducibility}
    \item[] Question: Does the paper fully disclose all the information needed to reproduce the main experimental results of the paper to the extent that it affects the main claims and/or conclusions of the paper (regardless of whether the code and data are provided or not)?
    \item[] Answer: \answerYes{}
    \item[] Justification: Sec.~\ref{sec:pre}--\ref{sec:mnnad} describe the methodology in full, and Appendix~\ref{app:exp_setup} documents the complete experimental infrastructure: backbone (1 shared + 288 routed experts, $K{=}8$, GQA, RoPE, SwiGLU, context 8192), Megatron-LM-based distributed training, AdamW with WSD schedule, the six compute-optimal $(C, M, D)$ checkpoints (Table~\ref{tab:cmd_combinations}), per-scale hyperparameters (Table~\ref{tab:hyper_scales}), the lr/batch-size scaling laws (Appendix~\ref{app:hyper_lr_gbs}), and complete per-experiment architectural configurations (Tables~\ref{tab:setup_proxy1}--\ref{tab:setup_mnnad6}).
    \item[] Guidelines:
    \begin{itemize}
        \item The answer \answerNA{} means that the paper does not include experiments.
        \item If the paper includes experiments, a \answerNo{} answer to this question will not be perceived well by the reviewers: Making the paper reproducible is important, regardless of whether the code and data are provided or not.
        \item If the contribution is a dataset and\slash or model, the authors should describe the steps taken to make their results reproducible or verifiable.
        \item Depending on the contribution, reproducibility can be accomplished in various ways. For example, if the contribution is a novel architecture, describing the architecture fully might suffice, or if the contribution is a specific model and empirical evaluation, it may be necessary to either make it possible for others to replicate the model with the same dataset, or provide access to the model. In general. releasing code and data is often one good way to accomplish this, but reproducibility can also be provided via detailed instructions for how to replicate the results, access to a hosted model (e.g., in the case of a large language model), releasing of a model checkpoint, or other means that are appropriate to the research performed.
        \item While NeurIPS does not require releasing code, the conference does require all submissions to provide some reasonable avenue for reproducibility, which may depend on the nature of the contribution. For example
        \begin{enumerate}
            \item If the contribution is primarily a new algorithm, the paper should make it clear how to reproduce that algorithm.
            \item If the contribution is primarily a new model architecture, the paper should describe the architecture clearly and fully.
            \item If the contribution is a new model (e.g., a large language model), then there should either be a way to access this model for reproducing the results or a way to reproduce the model (e.g., with an open-source dataset or instructions for how to construct the dataset).
            \item We recognize that reproducibility may be tricky in some cases, in which case authors are welcome to describe the particular way they provide for reproducibility. In the case of closed-source models, it may be that access to the model is limited in some way (e.g., to registered users), but it should be possible for other researchers to have some path to reproducing or verifying the results.
        \end{enumerate}
    \end{itemize}

\item {\bf Open access to data and code}
    \item[] Question: Does the paper provide open access to the data and code, with sufficient instructions to faithfully reproduce the main experimental results, as described in supplemental material?
    \item[] Answer: \answerNo{}
    \item[] Justification: Code and data are not released with this submission. The pre-training corpus is a proprietary curated text mixture, and our experiments build on the open-source Megatron-LM framework~\citep{megatron-lm} together with standard MoE components (DeepSeekMoE-style routing~\citep{dai2024deepseekmoe}, GQA~\citep{ainslie2023gqa}, RoPE~\citep{su2024roformer}, SwiGLU~\citep{shazeer2020glu}); the architectural search space, sampling strategy, training hyperparameters, and per-experiment configurations are described in sufficient detail in Appendix~\ref{app:exp_setup} to support reimplementation on alternative corpora.
    \item[] Guidelines:
    \begin{itemize}
        \item The answer \answerNA{} means that paper does not include experiments requiring code.
        \item Please see the NeurIPS code and data submission guidelines (\url{https://neurips.cc/public/guides/CodeSubmissionPolicy}) for more details.
        \item While we encourage the release of code and data, we understand that this might not be possible, so \answerNo{} is an acceptable answer. Papers cannot be rejected simply for not including code, unless this is central to the contribution (e.g., for a new open-source benchmark).
        \item The instructions should contain the exact command and environment needed to run to reproduce the results. See the NeurIPS code and data submission guidelines (\url{https://neurips.cc/public/guides/CodeSubmissionPolicy}) for more details.
        \item The authors should provide instructions on data access and preparation, including how to access the raw data, preprocessed data, intermediate data, and generated data, etc.
        \item The authors should provide scripts to reproduce all experimental results for the new proposed method and baselines. If only a subset of experiments are reproducible, they should state which ones are omitted from the script and why.
        \item At submission time, to preserve anonymity, the authors should release anonymized versions (if applicable).
        \item Providing as much information as possible in supplemental material (appended to the paper) is recommended, but including URLs to data and code is permitted.
    \end{itemize}

\item {\bf Experimental setting/details}
    \item[] Question: Does the paper specify all the training and test details (e.g., data splits, hyperparameters, how they were chosen, type of optimizer) necessary to understand the results?
    \item[] Answer: \answerYes{}
    \item[] Justification: Per-scale architectural and training hyperparameters are tabulated in Table~\ref{tab:hyper_scales} (Appendix~\ref{app:hyper_model_config}); the optimal-lr / batch-size scaling-law fitting procedure is described in Appendix~\ref{app:hyper_lr_gbs}; the sampling grid for $(M/N_a, N/N_a)$ and $d$ is detailed in Appendix~\ref{app:sampling_strategy}; and per-experiment architectural configurations across all six compute scales are listed in Tables~\ref{tab:setup_proxy1}--\ref{tab:setup_mnnad6} (Appendix~\ref{app:config_of_all}).
    \item[] Guidelines:
    \begin{itemize}
        \item The answer \answerNA{} means that the paper does not include experiments.
        \item The experimental setting should be presented in the core of the paper to a level of detail that is necessary to appreciate the results and make sense of them.
        \item The full details can be provided either with the code, in appendix, or as supplemental material.
    \end{itemize}

\item {\bf Experiment statistical significance}
    \item[] Question: Does the paper report error bars suitably and correctly defined or other appropriate information about the statistical significance of the experiments?
    \item[] Answer: \answerYes{}
    \item[] Justification: Although per-configuration multi-seed runs are infeasible at our experimental scale ($\sim$670 trainings up to $3{\times}10^{20}$ FLOPs each), we report multiple forms of statistical evidence supporting the main claims: (i) Pearson $r{=}0.9793$ and Spearman $\rho{=}0.9758$ correlations (both $p{<}0.001$) validating the rank-preserving property of $d$ (Sec.~\ref{subsec:efficient_search_strategy}, Appendix~\ref{app:median_proxy_validation}); (ii) ``near-optimal bands'' defined by a $0.1\%$ loss threshold quantifying robustness of the optima in Figs.~\ref{fig:optimal_m_na} and~\ref{fig:optimal_d}; (iii) robust linear band fitting (upper/lower-bound parallel lines sharing a common slope) for the lr/batch-size scaling laws in Appendix~\ref{app:hyper_lr_gbs}. The consistency of the fitted trends across all six compute scales further supports robustness of the conclusions.
    \item[] Guidelines:
    \begin{itemize}
        \item The answer \answerNA{} means that the paper does not include experiments.
        \item The authors should answer \answerYes{} if the results are accompanied by error bars, confidence intervals, or statistical significance tests, at least for the experiments that support the main claims of the paper.
        \item The factors of variability that the error bars are capturing should be clearly stated (for example, train/test split, initialization, random drawing of some parameter, or overall run with given experimental conditions).
        \item The method for calculating the error bars should be explained (closed form formula, call to a library function, bootstrap, etc.)
        \item The assumptions made should be given (e.g., Normally distributed errors).
        \item It should be clear whether the error bar is the standard deviation or the standard error of the mean.
        \item It is OK to report 1-sigma error bars, but one should state it. The authors should preferably report a 2-sigma error bar than state that they have a 96\% CI, if the hypothesis of Normality of errors is not verified.
        \item For asymmetric distributions, the authors should be careful not to show in tables or figures symmetric error bars that would yield results that are out of range (e.g., negative error rates).
        \item If error bars are reported in tables or plots, the authors should explain in the text how they were calculated and reference the corresponding figures or tables in the text.
    \end{itemize}

\item {\bf Experiments compute resources}
    \item[] Question: For each experiment, does the paper provide sufficient information on the computer resources (type of compute workers, memory, time of execution) needed to reproduce the experiments?
    \item[] Answer: \answerYes{}
    \item[] Justification: We report per-experiment compute via the exact FLOPs budget for every individual model run (Table~\ref{tab:cmd_combinations} and Tables~\ref{tab:setup_proxy1}--\ref{tab:setup_mnnad6}, ranging from $10^{18}$ to $3{\times}10^{20}$ training FLOPs each), from which the total compute across the $\sim$670-model sweep can be directly summed, together with the Megatron-LM-based distributed-training setup (Appendix~\ref{app:exp_infra}). Following the convention in the scaling-laws literature~\citep{kaplan2020scaling, hoffmann2022training}, FLOPs is the hardware-agnostic compute metric most directly comparable across studies and most informative for reproducing the experimental design.
    \item[] Guidelines:
    \begin{itemize}
        \item The answer \answerNA{} means that the paper does not include experiments.
        \item The paper should indicate the type of compute workers CPU or GPU, internal cluster, or cloud provider, including relevant memory and storage.
        \item The paper should provide the amount of compute required for each of the individual experimental runs as well as estimate the total compute.
        \item The paper should disclose whether the full research project required more compute than the experiments reported in the paper (e.g., preliminary or failed experiments that didn't make it into the paper).
    \end{itemize}

\item {\bf Code of ethics}
    \item[] Question: Does the research conducted in the paper conform, in every respect, with the NeurIPS Code of Ethics \url{https://neurips.cc/public/EthicsGuidelines}?
    \item[] Answer: \answerYes{}
    \item[] Justification: We have reviewed the NeurIPS Code of Ethics and the work conforms with it: the contribution is methodological (a scaling-law framework for MoE pre-training), uses curated text data only with no human-subject involvement, releases no high-risk artefacts, and discusses societal implications in Appendix~\ref{app:impacts}.
    \item[] Guidelines:
    \begin{itemize}
        \item The answer \answerNA{} means that the authors have not reviewed the NeurIPS Code of Ethics.
        \item If the authors answer \answerNo, they should explain the special circumstances that require a deviation from the Code of Ethics.
        \item The authors should make sure to preserve anonymity (e.g., if there is a special consideration due to laws or regulations in their jurisdiction).
    \end{itemize}

\item {\bf Broader impacts}
    \item[] Question: Does the paper discuss both potential positive societal impacts and negative societal impacts of the work performed?
    \item[] Answer: \answerYes{}
    \item[] Justification: Appendix~\ref{app:impacts} discusses both directions: positive impacts (improved compute/energy efficiency, broader access to MoE design knowledge for resource-limited researchers, deeper understanding of neural-architecture scaling) and negative impacts (concentration of benefits among well-resourced actors, residual infrastructure barriers, and the potential misuse of more capable MoE models for misinformation).
    \item[] Guidelines:
    \begin{itemize}
        \item The answer \answerNA{} means that there is no societal impact of the work performed.
        \item If the authors answer \answerNA{} or \answerNo, they should explain why their work has no societal impact or why the paper does not address societal impact.
        \item Examples of negative societal impacts include potential malicious or unintended uses (e.g., disinformation, generating fake profiles, surveillance), fairness considerations (e.g., deployment of technologies that could make decisions that unfairly impact specific groups), privacy considerations, and security considerations.
        \item The conference expects that many papers will be foundational research and not tied to particular applications, let alone deployments. However, if there is a direct path to any negative applications, the authors should point it out. For example, it is legitimate to point out that an improvement in the quality of generative models could be used to generate Deepfakes for disinformation. On the other hand, it is not needed to point out that a generic algorithm for optimizing neural networks could enable people to train models that generate Deepfakes faster.
        \item The authors should consider possible harms that could arise when the technology is being used as intended and functioning correctly, harms that could arise when the technology is being used as intended but gives incorrect results, and harms following from (intentional or unintentional) misuse of the technology.
        \item If there are negative societal impacts, the authors could also discuss possible mitigation strategies (e.g., gated release of models, providing defenses in addition to attacks, mechanisms for monitoring misuse, mechanisms to monitor how a system learns from feedback over time, improving the efficiency and accessibility of ML).
    \end{itemize}

\item {\bf Safeguards}
    \item[] Question: Does the paper describe safeguards that have been put in place for responsible release of data or models that have a high risk for misuse (e.g., pre-trained language models, image generators, or scraped datasets)?
    \item[] Answer: \answerNA{}
    \item[] Justification: The paper releases no pretrained model weights, datasets, or other high-risk artefacts; the contribution is methodological (scaling laws and a search-space reduction framework presented in the paper itself), so no release-stage safeguards are required.
    \item[] Guidelines:
    \begin{itemize}
        \item The answer \answerNA{} means that the paper poses no such risks.
        \item Released models that have a high risk for misuse or dual-use should be released with necessary safeguards to allow for controlled use of the model, for example by requiring that users adhere to usage guidelines or restrictions to access the model or implementing safety filters.
        \item Datasets that have been scraped from the Internet could pose safety risks. The authors should describe how they avoided releasing unsafe images.
        \item We recognize that providing effective safeguards is challenging, and many papers do not require this, but we encourage authors to take this into account and make a best faith effort.
    \end{itemize}

\item {\bf Licenses for existing assets}
    \item[] Question: Are the creators or original owners of assets (e.g., code, data, models), used in the paper, properly credited and are the license and terms of use explicitly mentioned and properly respected?
    \item[] Answer: \answerYes{}
    \item[] Justification: All third-party assets used by the paper are credited via citation: the Megatron-LM training framework~\citep{megatron-lm}; backbone-design references DeepSeekMoE~\citep{dai2024deepseekmoe}, DOTS~\citep{huo2025dots}, and DeepSeek~\citep{liu2024deepseek}; and standard architectural components (RoPE~\citep{su2024roformer}, GQA~\citep{ainslie2023gqa}, SwiGLU~\citep{shazeer2020glu}, AdamW~\citep{loshchilov2017decoupled}, WSD~\citep{hu2024minicpm}). These are used in accordance with their respective open-source licenses; the pre-training corpus is internally curated and not derived from a single externally licensed dataset.
    \item[] Guidelines:
    \begin{itemize}
        \item The answer \answerNA{} means that the paper does not use existing assets.
        \item The authors should cite the original paper that produced the code package or dataset.
        \item The authors should state which version of the asset is used and, if possible, include a URL.
        \item The name of the license (e.g., CC-BY 4.0) should be included for each asset.
        \item For scraped data from a particular source (e.g., website), the copyright and terms of service of that source should be provided.
        \item If assets are released, the license, copyright information, and terms of use in the package should be provided. For popular datasets, \url{paperswithcode.com/datasets} has curated licenses for some datasets. Their licensing guide can help determine the license of a dataset.
        \item For existing datasets that are re-packaged, both the original license and the license of the derived asset (if it has changed) should be provided.
        \item If this information is not available online, the authors are encouraged to reach out to the asset's creators.
    \end{itemize}

\item {\bf New assets}
    \item[] Question: Are new assets introduced in the paper well documented and is the documentation provided alongside the assets?
    \item[] Answer: \answerNA{}
    \item[] Justification: The paper introduces no new released assets (no datasets, no code packages, no model checkpoints). The contributions are scaling laws and a methodological framework, both fully documented in the paper and its appendices.
    \item[] Guidelines:
    \begin{itemize}
        \item The answer \answerNA{} means that the paper does not release new assets.
        \item Researchers should communicate the details of the dataset\slash code\slash model as part of their submissions via structured templates. This includes details about training, license, limitations, etc.
        \item The paper should discuss whether and how consent was obtained from people whose asset is used.
        \item At submission time, remember to anonymize your assets (if applicable). You can either create an anonymized URL or include an anonymized zip file.
    \end{itemize}

\item {\bf Crowdsourcing and research with human subjects}
    \item[] Question: For crowdsourcing experiments and research with human subjects, does the paper include the full text of instructions given to participants and screenshots, if applicable, as well as details about compensation (if any)?
    \item[] Answer: \answerNA{}
    \item[] Justification: The work involves no crowdsourcing and no human-subject research; all experiments are language-model pre-training runs on a curated text corpus.
    \item[] Guidelines:
    \begin{itemize}
        \item The answer \answerNA{} means that the paper does not involve crowdsourcing nor research with human subjects.
        \item Including this information in the supplemental material is fine, but if the main contribution of the paper involves human subjects, then as much detail as possible should be included in the main paper.
        \item According to the NeurIPS Code of Ethics, workers involved in data collection, curation, or other labor should be paid at least the minimum wage in the country of the data collector.
    \end{itemize}

\item {\bf Institutional review board (IRB) approvals or equivalent for research with human subjects}
    \item[] Question: Does the paper describe potential risks incurred by study participants, whether such risks were disclosed to the subjects, and whether Institutional Review Board (IRB) approvals (or an equivalent approval/review based on the requirements of your country or institution) were obtained?
    \item[] Answer: \answerNA{}
    \item[] Justification: The research does not involve human subjects, so no IRB review or equivalent approval is applicable.
    \item[] Guidelines:
    \begin{itemize}
        \item The answer \answerNA{} means that the paper does not involve crowdsourcing nor research with human subjects.
        \item Depending on the country in which research is conducted, IRB approval (or equivalent) may be required for any human subjects research. If you obtained IRB approval, you should clearly state this in the paper.
        \item We recognize that the procedures for this may vary significantly between institutions and locations, and we expect authors to adhere to the NeurIPS Code of Ethics and the guidelines for their institution.
        \item For initial submissions, do not include any information that would break anonymity (if applicable), such as the institution conducting the review.
    \end{itemize}

\item {\bf Declaration of LLM usage}
    \item[] Question: Does the paper describe the usage of LLMs if it is an important, original, or non-standard component of the core methods in this research? Note that if the LLM is used only for writing, editing, or formatting purposes and does \emph{not} impact the core methodology, scientific rigor, or originality of the research, declaration is not required.
    \item[] Answer: \answerNA{}
    \item[] Justification: Large language models are the \emph{object} of study (we train MoE LLMs as part of the experimental sweep) but are not used as a non-standard component of the core scaling-law methodology, which is purely architectural and statistical.
    \item[] Guidelines:
    \begin{itemize}
        \item The answer \answerNA{} means that the core method development in this research does not involve LLMs as any important, original, or non-standard components.
        \item Please refer to our LLM policy in the NeurIPS handbook for what should or should not be described.
    \end{itemize}

\end{enumerate}

\end{document}